\documentclass{article}
\usepackage{graphicx} 
\usepackage[square,numbers]{natbib}

\usepackage[preprint]{neurips_2024}

\usepackage[utf8]{inputenc} 
\usepackage[T1]{fontenc}    
\usepackage[hidelinks]{hyperref}        
\usepackage{url}            
\usepackage{booktabs}     
\usepackage[dvipsnames]{xcolor}   
\usepackage{amsfonts}       
\usepackage{nicefrac}       
\usepackage{microtype}      
\usepackage{amsmath}        

\usepackage{caption}        
\usepackage{subcaption}     
\usepackage{enumitem}       
\usepackage{multirow}       

\usepackage[cjk]{kotex}
\usepackage{wrapfig}
\usepackage{xspace}

\hypersetup{
    colorlinks=true,
    citecolor=RoyalBlue,
    urlcolor=teal
}

\title{LMLT: Low-to-high Multi-Level Vision Transformer for Image Super-Resolution}
\author{%
  Jeongsoo Kim, Jongho Nang, Junsuk Choe\thanks{Corresponding author}\\
  Sogang University\\
  \texttt{\{jskim1, jhnang, jschoe\}@sogang.ac.kr}
  }
  
\date{May 2024}

\begin{document}

\maketitle

\begin{abstract}
Recent Vision Transformer (ViT)-based methods for Image Super-Resolution have demonstrated impressive performance. However, they suffer from significant complexity, resulting in high inference times and memory usage. Additionally, ViT models using Window Self-Attention (WSA) face challenges in processing regions outside their windows. To address these issues, we propose the Low-to-high Multi-Level Transformer (LMLT), which employs attention with varying feature sizes for each head. LMLT divides image features along the channel dimension, gradually reduces spatial size for lower heads, and applies self-attention to each head. This approach effectively captures both local and global information. By integrating the results from lower heads into higher heads, LMLT overcomes the window boundary issues in self-attention. Extensive experiments show that our model significantly reduces inference time and GPU memory usage while maintaining or even surpassing the performance of state-of-the-art ViT-based Image Super-Resolution methods. Our codes are availiable at \url{https://github.com/jwgdmkj/LMLT}.
\end{abstract}

\section{Introduction}

Single Image Super-Resolution (SISR) is a technique that converts low-resolution images into high-resolution ones and has been actively researched in the field of computer vision. Traditional methods, such as nearest neighbor interpolation and bilinear interpolation, were used in the past, but recent super-resolution research has seen significant performance improvements, particularly through CNN-based methods~\cite{SRCNN, ECBSR, VDSR} and Vision Transformer (ViT)-based methods~\cite{SwinIR, elan, ngswin}.

Since the introduction of SRCNN~\cite{SRCNN}, CNN-based image super-resolution architectures have advanced by utilizing multiple convolutional layers to understand contexts at various scales. These architectures deliver this understanding through residual and/or dense connections~\cite{res1, res2, res3, res4, res5}. 

However, super-resolution using CNNs faces several issues in terms of performance and efficiency. Firstly, CNN-based models can become excessively complex and deep to improve performance, leading to increased model size and memory usage~\cite{ShuffleMixer, ECBSR, boost}. To mitigate this, several models share parameters between modules~\cite{CARN, res4}, but this approach does not guarantee efficiency during inference~\cite{SAFMN}. SAFMN~\cite{SAFMN} addresses the balance between accuracy and complexity by partitioning image features using a multi-head approach~\cite{Attn} and implementing non-local feature relationships at various scales. However, it struggles to capture long-range dependencies due to limited kernel sizes.

ViT-based models have shown superior performance compared to CNN-based models by effectively modeling global context interactions~\cite{ngswin, elan}. For example, SwinIR~\cite{SwinIR} utilized the Swin Transformer~\cite{swin} for image super-resolution, demonstrating the effectiveness of the transformer architecture. Subsequently, hybrid models combining ViT and CNN have been proposed, achieving significant performance increases~\cite{hnct}. 

However, ViT models face quadratically increasing computational costs as input size grows~\cite{orthogonal, qvit, slidet}. To address this, Window Self-Attention (WSA) has been developed, which perform self-attention by dividing the image into windows~\cite{swin}. Despite this, WSA suffers from quality degradation at window boundaries and lacks interaction between windows~\cite{fastervit, slidet}. Additionally, conventional ViT-based models stack self-attention layers in series~\cite{SwinIR, srformer}, which significantly increases computational load and inference time.



\begin{figure*}[!t]\footnotesize
 \begin{center}
  \begin{tabular}{cc}
  
    \includegraphics[width=0.6\linewidth, height=0.4\linewidth]{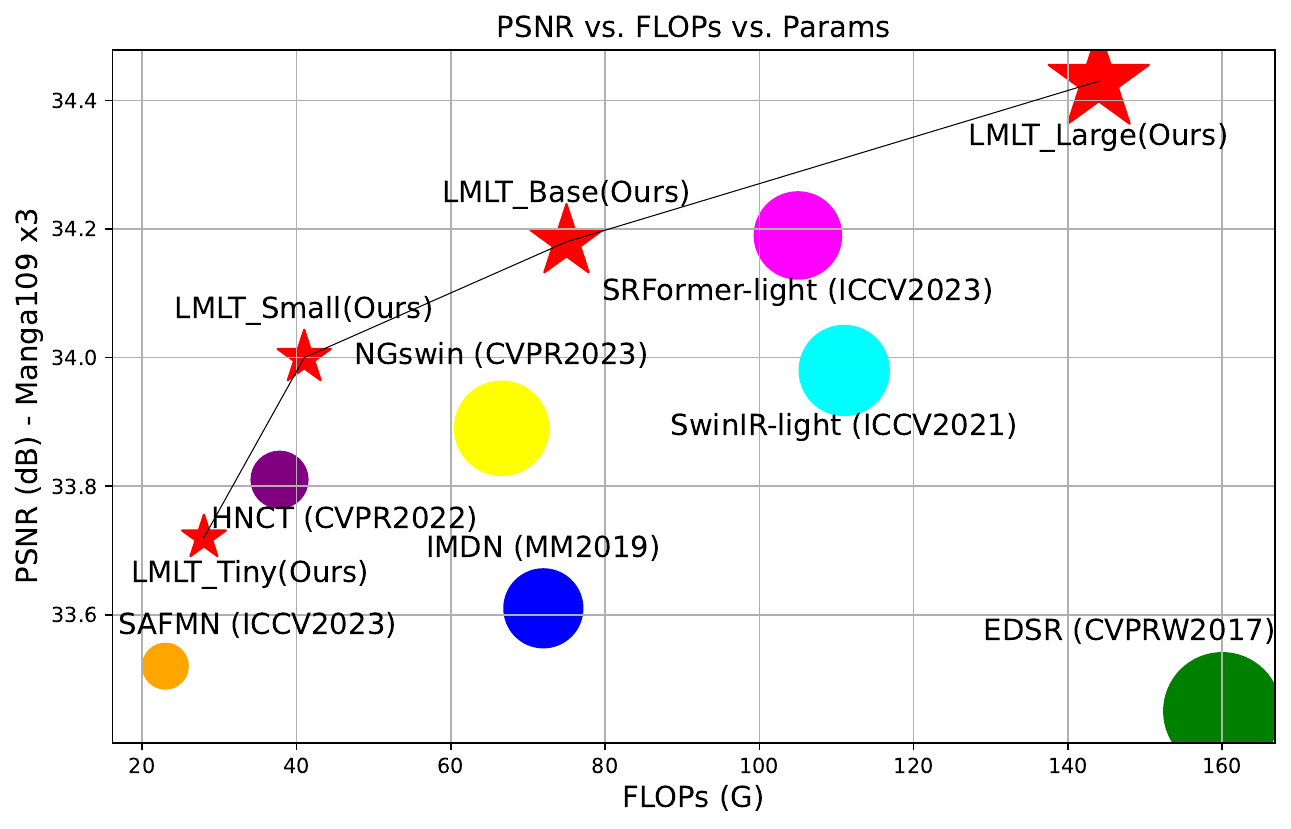} &\hspace{-3.5mm}
    \includegraphics[width=0.37\linewidth, height = 0.4\linewidth]{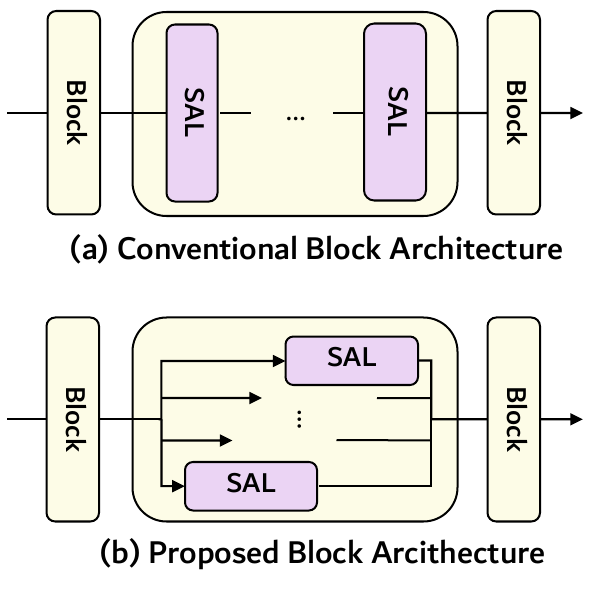} \\
\\
 \end{tabular}
 \end{center}
 \caption{\textbf{Left} PSNR comparison of our proposed LMLT and other state-of-the-art models when upscaling Manga109 by 3 times. The size of each circle represents the number of parameters. Our model achieves comparable performance in terms of FLOPs when the channels are set to 36, 36 with 12 blocks, 60, and 84. \textbf{Right} (a) The conventional Self-Attention block stacks multiple Self-Attention layers in series. (b) Our proposed Self-Attention block stacks the layers in parallel. Here, SAL stands for Self Attenion Layer.}
  \label{fig:teaser}
\end{figure*}

In this paper, we propose LMLT (Low-to-high Multi-Level Transformer) to improve efficiency during inference while maintaining performance. Similar to SAFMN, our approach uses a multi-head method~\cite{Attn} to split image features and apply pooling to each feature. Each head applies the self-attention mechanism. Unlike conventional self-attention blocks, which stack self-attention layers in series (Figure~\ref{fig:teaser}(a)), we stack them in parallel to reduce computation (Figure~\ref{fig:teaser}(b)). This means we integrate the number of heads and layers (depth) into a single mechanism. Note that we call the head with the most pooling the lower head, and the number of pooling applications decreases as we move to the upper heads. 

Since the window size is the same for all heads, the upper heads focus on smaller areas and effectively capture local context. In contrast, the lower heads focus on larger areas and learn more global information. This approach allows us to dynamically capture both local and global information. Additionally, we introduce a residual connection~\cite{resnet} to pass global information from the more pooled lower heads to the less pooled upper heads. This enables the windows of the upper heads to view a wider area, thereby resolving the cross-window communication problem.

Trained on DIV2K~\cite{DIV2K} and Flickr2K~\cite{Flickr2K-EDSR}, our extensive experiments demonstrate that ViT-based models can effectively achieve a balance between model complexity and accuracy. Compared to other state-of-the-art results, our approach significantly reduces memory usage and inference time while enhancing performance. Specifically, our base model with 60 channels and large model with 84 channels decrease memory usage to 38\% and 54\%, respectively, and inference time to 22\% and 19\% compared to ViT-based super-resolution models like NGswin~\cite{ngswin} and SwinIR-light\cite{SwinIR} at scale $\times 4$ scale. Moreover, our models achieve an average performance increase of 0.076db and 0.152db across all benchmark datasets.

\section{Related Works}

\textbf{CNN-Based Image Super-Resolution} is one of the most popular deep learning-based methods for enhancing image resolution. Since SRCNN~\cite{SRCNN} introduced a method to restore high-resolution (HR) images using three end-to-end layers, advancements like VDSR~\cite{VDSR} and DRCN~\cite{drcn} have leveraged deeper neural network structures. These methods introduced recursive neural network structures to produce higher quality results. ESPCN~\cite{espcn} significantly improved the speed of super-resolution by replacing bicubic-filter upsampling with sub-pixel convolution, a technique adopted in several subsequent works~\cite{SwinIR, srformer, hnct}. To address the limited receptive field of CNNs, some researchers incorporated attention mechanisms into super-resolution models to capture larger areas. RCAN~\cite{RCAN} applied channel attention to adaptively readjust the features of each channel, while SAN~\cite{SAN} used a second-order attention mechanism to capture more long-distance spatial contextual information. CSFM~\cite{csfm} dynamically modulated channel-wise and spatial attention, allowing the model to selectively emphasize various global and local features of the image. We use the same window size, similar to CNN kernels, but vary the spatial size of each feature. This allows our model to dynamically capture both local and global information by obtaining global information from smaller spatial sizes and local information from larger spatial sizes.

\textbf{ViT-Based Image Super-Resolution} has surpassed the performance of CNN-based models by efficiently modeling long-range dependencies and capturing global interactions between contexts~\cite{SwinIR, ShuffleMixer}. After the success of ViT~\cite{vit} in various fields such as classification~\cite{swin, cswin}, object detection~\cite{detr, camoformer}, and semantic segmentation~\cite{segformer, segmenter}, several models have aimed to use it for low-level vision tasks. IPT~\cite{ipt} constructed a Transformer-based large-scale pre-trained model for image processing. However, the complexity of ViT grows quadratically with input size. To mitigate this, many approaches have aimed to reduce computational load while capturing both local and global information. For example, SwinIR~\cite{SwinIR} used the Swin-Transformer~\cite{swin} model for image reconstruction. Restormer~\cite{restormer} organized self-attention in the channel direction to maintain global information and achieve high performance in image denoising. HAT~\cite{hat} combined self-attention, which captures representative information, with channel attention, which holds global information. To combine local and global information without adding extra complexity, we add features from lower heads, which contain global information, to upper heads, which contain local information. This enables the windows to see beyond their own area and cover a larger region.

\textbf{Efficient Image Super-Resolution} research focuses on making super-resolution models more efficient. The CNN-based model FSRCNN~\cite{fsrcnn} improved on SRCNN~\cite{SRCNN} by removing the bicubic interpolation pre-processing and increasing the scale through deconvolution, greatly speeding up computation. CARN~\cite{CARN} reused features at various stages through cascading residual blocks connected in a multi-stage manner. IMDN~\cite{IMDN} progressively refined features passing through the network. However, improving performance often requires stacking many convolution layers, leading to increased computational load and memory usage.

In contrast, the ViT-based model ELAN~\cite{elan} aimed to enhance spatial adaptability by using various window sizes in self-attention. HNCT~\cite{hnct} integrated CNN and Transformer structures to extract local features with global dependencies. NGswin~\cite{ngswin} addressed the cross-window communication problem of the original Swin Transformer~\cite{swin} by applying Ngram~\cite{ngram}. Despite these advances, the considerable computational load of overly deep stacked self-attention mechanisms still constrains the efficiency of ViT-based super-resolution models. To address computational load, we connect self-attention layers in parallel, integrating multi-head and depth (number of layers) to lighten the computation. This, along with reducing the spatial size of features, makes our model more efficient.

Additionally, efforts to lighten networks through methods such as knowledge distillation~\cite{FAKD}, model quantization~\cite{quantizationsr}, or pruning~\cite{srpn, prunesr} have been made. Some approaches differentiate between classical and lightweight image super-resolution models by using the same architecture but varying hyperparameters, such as the number of network blocks or feature channels~\cite{srformer, dat, dlgsanet}. 

\begin{figure}
	\centering
	\includegraphics[width=\textwidth]{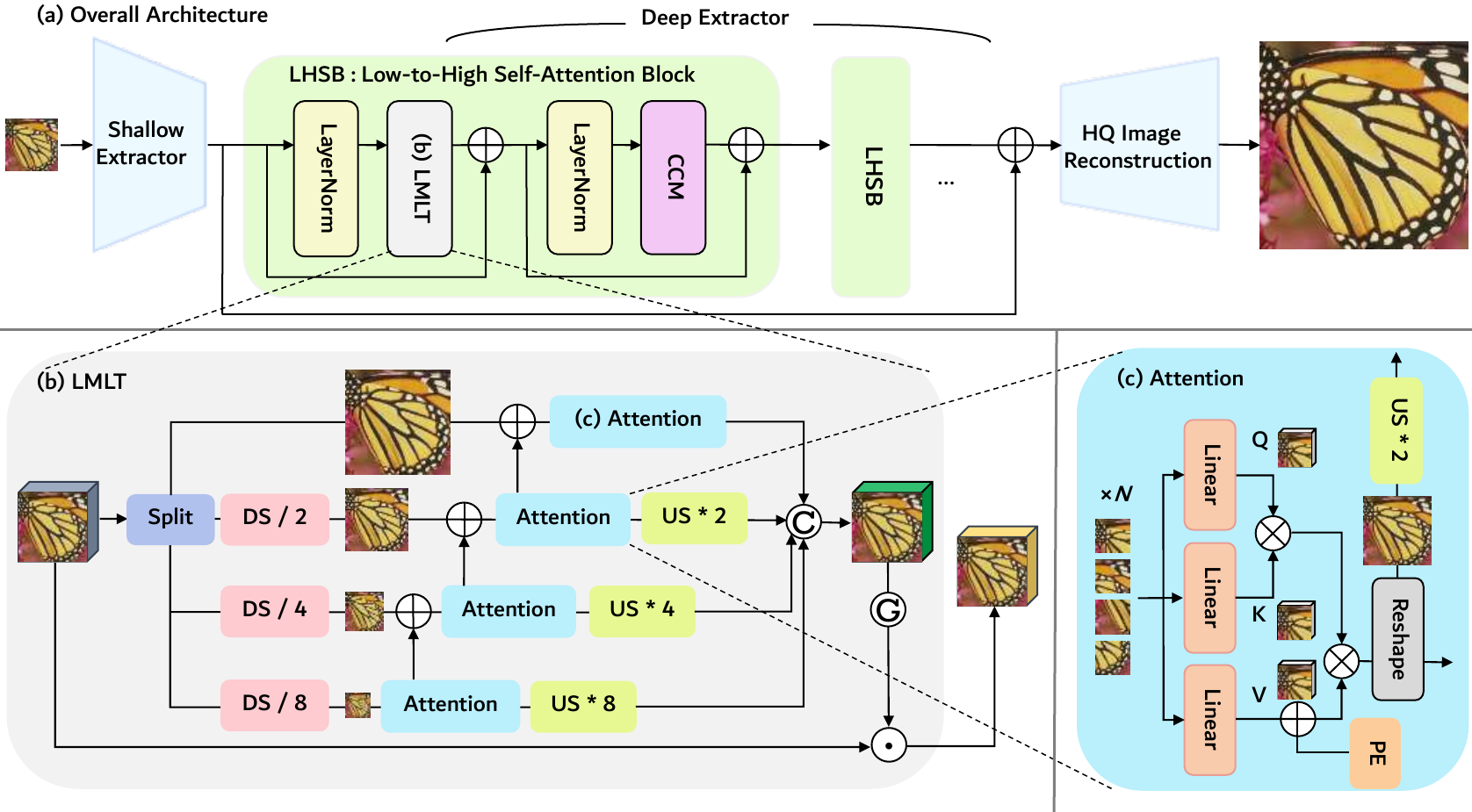}
	
	\caption{The architecture overview of the proposed method.}  
    \vspace{-1em}
	\label{fig:arch}
\end{figure}

\section{Proposed Method}

\textbf{Overall Architecture (Figure~\ref{fig:arch}(a)).} First, we use a $3 \times 3$ convolution to extract shallow-level features from the image. Next, we stack multiple LHS Blocks (Low-to-High Self-attention Blocks) to extract deep-level features. In each LHS Block, the features go through Layer Normalization (LN)~\cite{LN}, our proposed LMLT (Low-to-high Multi-Level Transformer), LN again, and the CCM~\cite{SAFMN}. Residual connections are also employed. Finally, we use a $3 \times 3$ convolution filter and a pixel-shuffle layer~\cite{espcn} to reconstruct high-quality images. For more details on CCM~\cite{SAFMN}, refer to Appendix~\ref{App:CCM}.

\textbf{Low-to-high Multi-Level Transformer (Figure~\ref{fig:arch}(b)).} LMLT operates within the LHS Block. After features pass through the first LN~\cite{LN}, we divide them into $H$ heads using a Multi-Head approach~\cite{Attn} and pool each split feature to a specified size. Specifically, the feature for the uppermost head is not pooled, and as we move to lower heads, the pooling becomes stronger, with the height and width halved for each subsequent head. Each feature then undergoes a self-attention mechanism. The output of the self-attention is interpolated to the size of the upper head's feature and added element-wise (called a low-to-high connection). The upper head then undergoes the self-attention process again, continuing up to the topmost head. Finally, the self-attention outputs of all heads are restored to their original size, concatenated, and merged through a $1 \times 1$ convolution before being multiplied with the original feature.

\textbf{Attention Layer (Figure~\ref{fig:arch}(c)).} In each attention layer, the feature is divided into $\mathit{N}$ non-overlapping windows. The dot product of the query and key is calculated, followed by the dot product with the value. LePE~\cite{cswin} is used as the Positional Encoding and added to the value. The output is upscaled by a factor of 2 and sequentially passed to the upper head until it reaches the topmost head.

\begin{figure*}[!t]\footnotesize
 \begin{center}
  \begin{tabular}{c}
        \includegraphics[width=\textwidth]{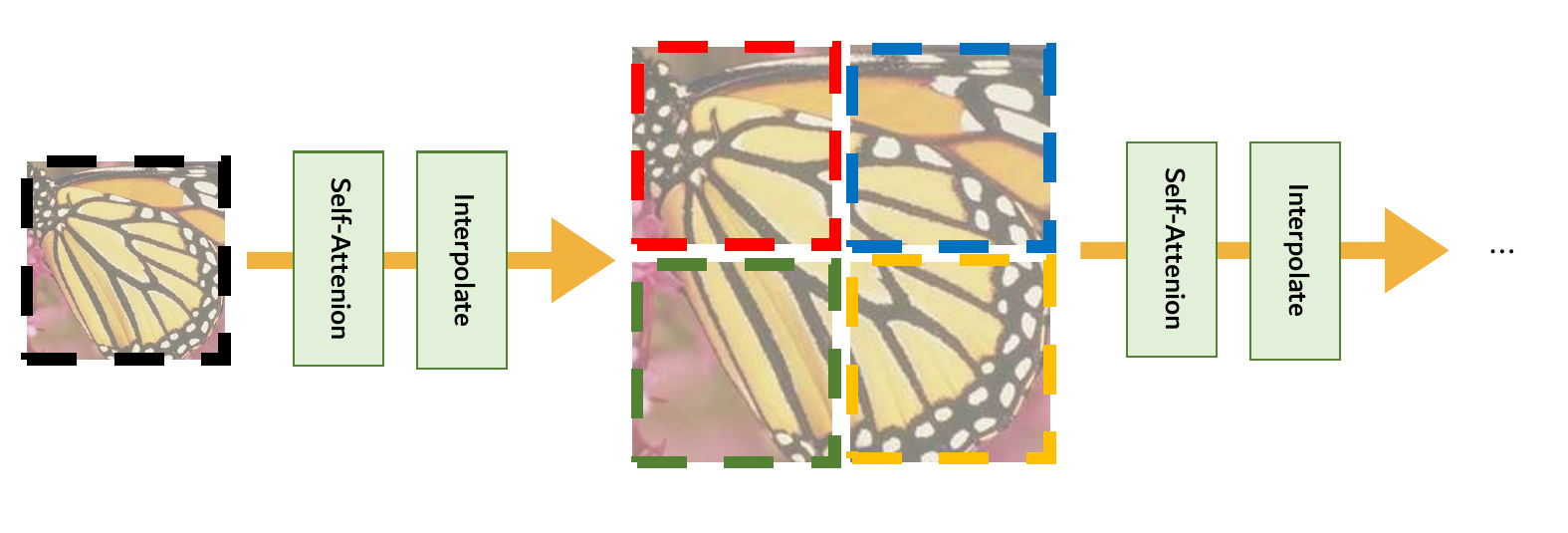}
\\
 \end{tabular}
 \end{center}
 \caption{Self-attention(SA) at different spatial resolutions of the image with the same window size.}
 \label{fig:window_mechanism}
\end{figure*}

\begin{wrapfigure}{R}{0.4\textwidth}
 \begin{center}
  \begin{tabular}{cccc}
      \includegraphics[width=0.19\linewidth, height = 0.19\linewidth]{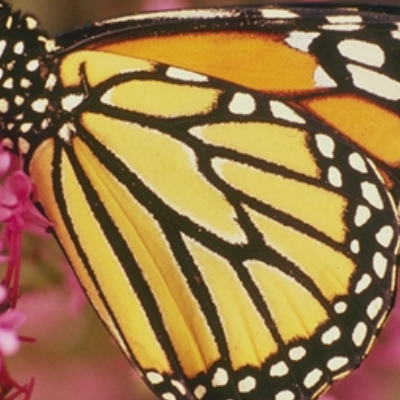} &\hspace{-2mm}
      \includegraphics[width=0.19\linewidth, height = 0.19\linewidth]{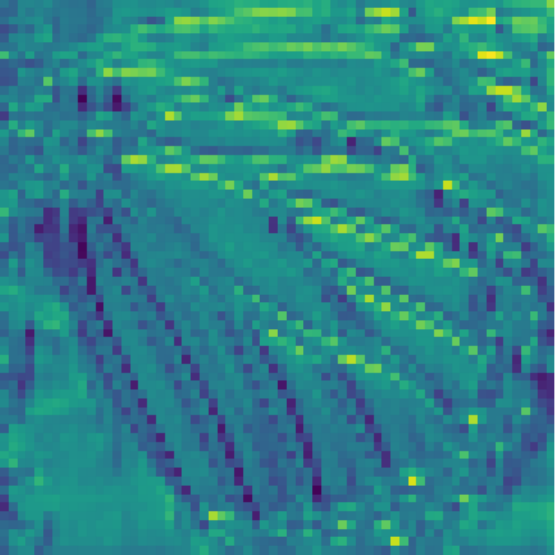} &\hspace{-2mm}
      \includegraphics[width=0.19\linewidth, height = 0.19\linewidth]{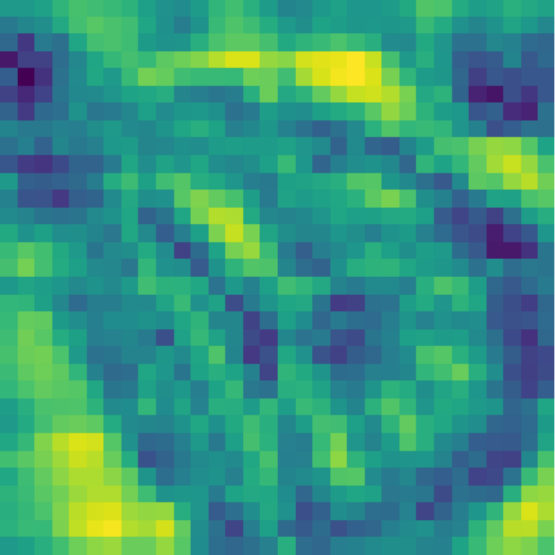} &\hspace{-2mm}
      \includegraphics[width=0.19\linewidth, height = 0.19\linewidth]{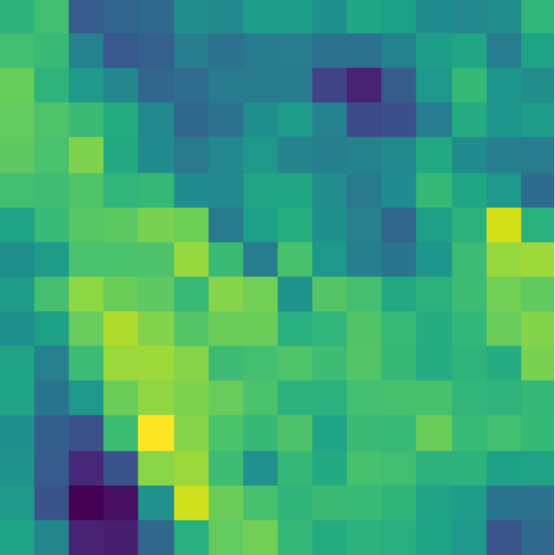} \\
      \hspace{-3.5mm}  &\hspace{-3.5mm}(a) &\hspace{-3.5mm}(b) &\hspace{-3.5mm}(c)  \\

      & \hspace{-2mm}
      \includegraphics[width=0.19\linewidth, height = 0.19\linewidth]{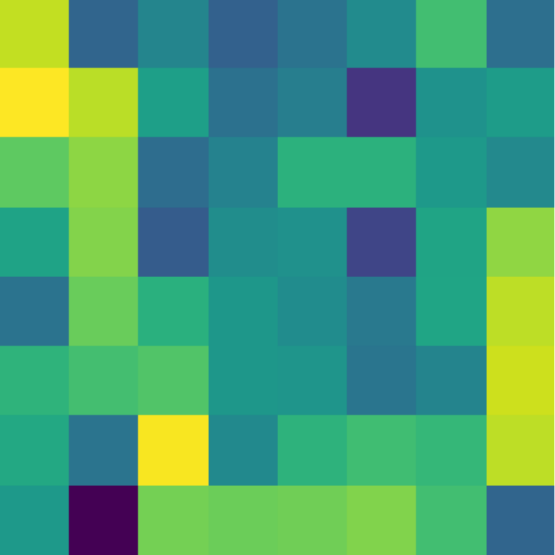} & \hspace{-2mm}
      \includegraphics[width=0.19\linewidth, height = 0.19\linewidth]{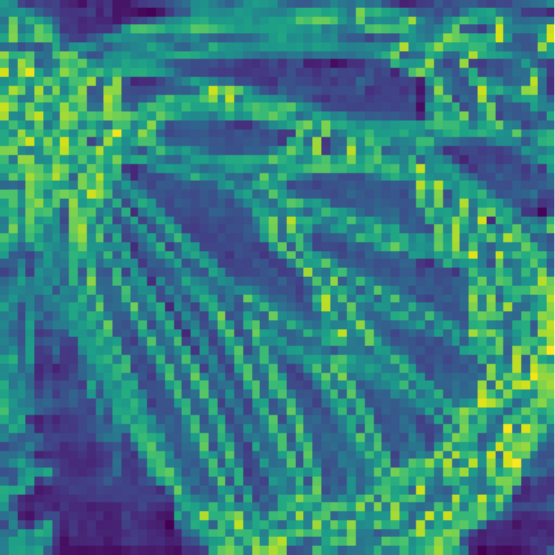} & \hspace{-2mm}
      \includegraphics[width=0.19\linewidth, height = 0.19\linewidth]{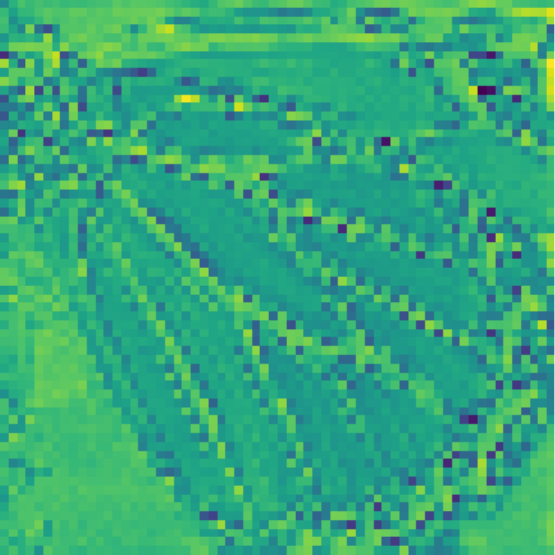} \\
       &\hspace{-3.5mm}(d)   &\hspace{-3.5mm}(e) &\hspace{-3.5mm}(f) 
      
\\
 \end{tabular}
 \end{center}
 \caption{Features from each head ((a) to (d)), aggregated feature (e), and feature multiplied with the original feature (f).}
 \label{fig:each_head_feat}
\end{wrapfigure}

\textbf{How the Proposed Method Works?} Our proposed LMLT effectively captures both local and global regions. As seen in Figure~\ref{fig:window_mechanism}, even if the window size is the same, the viewing area changes with different spatial sizes of the feature. Specifically, when self-attention is applied to smaller features, global information can be obtained. As the spatial size increases, the red window in the larger feature can utilize information beyond its own limits for self-attention calculation because it has already acquired information from other regions in the previous stage. This combination of lower heads capturing global context and upper heads capturing local context secures cross-window communication. Figure~\ref{fig:each_head_feat} visualizes the type of information each head captures. From Figure~\ref{fig:each_head_feat}(a) to ~\ref{fig:each_head_feat}(d), the features extracted from each head when $\mathit{H}$ is assumed to be 4 are visualized by averaging them along the channel dimension. The first head (\ref{fig:each_head_feat}(a)) captures relatively local patterns, while fourth head (\ref{fig:each_head_feat}(d)) captures global patterns. In Figure~\ref{fig:each_head_feat}(e), these local and global patterns are combined to provide a comprehensive representation. By merging this with the original feature (\ref{fig:each_head_feat}(f)), it emphasizes the parts that are important for super-resolution. 

\textbf{Computational Complexity.} We improve the model's efficiency by connecting self-attention layers in parallel and reducing spatial size. In the proposed model, given a feature \(\mathit{F} \in \mathbb{R}^{\mathit{H} \times \mathit{W} \times \mathit{D}}\) and a fixed window size of \(\mathit{M} \times \mathit{M}\), the number of windows in our LMLT is reduced by one-fourth as we move to lower heads, by halving the spatial size of the feature map. Additionally, since each head replaces depth, the channel count is also reduced to \(\frac{\mathit{D}}{\mathit{head}}\). Therefore, the total computation for each head is given by Equation~\ref{eq2}. Here, $\mathit{i}$ means $\mathit{i-1}$th head.

\begin{equation}
\label{eq2}
\Omega(\text{LMLT}) = 4\left[\frac{hw}{4^i}\left(\frac{D}{\mathit{head}}\right)^2\right] + 2 \left[ \frac{M^2 hw}{4^i} \frac{D}{\mathit{head}}\right].
\end{equation}

On the other hand, in WSA~\cite{swin, SwinIR}, the self-attention layers are stacked in series, and the spatial size and channel of the feature do not decrease, so the number of windows remains \(\frac{\mathit{HW}}{\mathit{M}^2}\), and the channel count stays at \(\mathit{D}\). Therefore, the total computation amount is shown in Equation~\ref{eq1}.

\begin{equation}
\label{eq1}
\Omega(\text{WSA}) = 4hw D^2 + 2M^2 hw D.
\end{equation}

Therefore, in our proposed model, if the number of heads is greater than 1, both the number of windows and channels decrease compared to WSA~\cite{swin, SwinIR}, resulting in reduced computational load. The more heads there are, the greater the reduction in computational load.

\section{Experiments}

\textbf{Datasets.} Following previous studies~\cite{SwinIR, ShuffleMixer, hnct}, we use DIV2K~\cite{DIV2K}, consisting of 800 images, and Flickr2K~\cite{Flickr2K-EDSR}, consisting of 2,650 images, as training datasets. For testing, we use the Set5~\cite{Set5}, Set14~\cite{Set14}, BSD100~\cite{BSD100}, Urban100~\cite{Urban100}, and Manga109~\cite{Manga109} datasets.

\textbf{Implementation Details.} We categorize our model into four types: a Tiny model with 36 channels, a Small model with 36 channels and 12 blocks, a Base model with 60 channels, and a Large model with 84 channels. First, the low-resolution (LR) images used as training inputs are cropped into $64 \times 64$ patches. Rotation and horizontal flip augmentations are applied to this training data. The number of blocks, heads, and growth ratio are set to 8 (except for the Small model), 4, and 2, respectively. We use the Adam Optimizer~\cite{Adam} with $\beta_1 = 0.9$ and $\beta_2 = 0.99$, running for 500,000 iterations. The initial learning rate is set to $1\times10^{-3}$ and is reduced to at least $1\times10^{-5}$ using the cosine annealing scheme~\cite{SGDR}. To accelerate the speed of our experiments, we set \texttt{backends.cudnn.benchmark} to \texttt{True} and \texttt{backends.cudnn.deterministic} to \texttt{False} for the 36-channel model. To account for potential variability in the results due to this setting, we conduct three separate experiments with the LMLT-Tiny model and reported the average of these results. All other experiments are conducted only once.


\textbf{Evaluation Metrics.} The quality of the recovered high-resolution images is evaluated using Peak Signal-to-Noise Ratio (PSNR) and Structural Similarity Index (SSIM)~\cite{SSIM}. These metrics are calculated on the Y channel of the YCbCr color space. To test the efficiency of our model, we follow the method of SAFMN~\cite{SAFMN}, measuring GPU memory consumption (\#GPU Mem) and inference time (\#AVG Time) for scaling a total of 50 images across various models. \#GPU Mem, obtained through PyTorch's \texttt{torch.cuda.max\_memory\_allocated()}, represents the maximum memory consumption during inference, and \#AVG Time is the average time per image for inferring a total of 50 LR images at $\times2$, $\times3$, and $\times4$ scales. The results for $\times2$, $\times3$, and $\times4$ scaling are based on upscaling random images of sizes $640\times360$, $427\times240$, and $320\times180$, respectively.

\subsection{Comparisons with State-of-the-Art Methods}
\label{comparisons}

\textbf{Image Reconstruction Comparisons.} To evaluate the performance of the proposed model, we compare our models with other state-of-the-art efficient and lightweight SR models at different scaling factors. PSNR, SSIM~\cite{SSIM}, the number of parameters, and FLOPs are used as the main performance evaluation metrics. Note that FLOPs refer to the computational amount required to create an image with a resolution of 1280$\times$720.

\begin{table}[t]
	\caption{Comparisons with our LMLT-Base, LMLT-Large and other Super-Resolution models on multiple benchmark datasets. Best and second-best performance are in \textcolor{red}{red} and \textcolor{blue}{blue} color.}
	\label{tab:baselarge}
	\renewcommand\arraystretch{1.1}
	\begin{center}
		\resizebox{\textwidth}{!}{
			\begin{tabular}{| c | c | c | c | c | c | c | c | c |}
				\hline
				Scale & Method & \#Params & \#FLOPs & Set5 & Set14 & B100 & Urban100 & Manga109 \\
				\hline
				\multirow{13}*{$\times 2$} 
                & IMDN ~\cite{IMDN}  &694K   &156G  & 38.00/0.9605 & 33.63/0.9177 & 32.19/0.8996 & 32.17/0.9283 & 38.88/0.9774 \\
				~ & LatticeNet~\cite{Lattice}    &756K   &170G   & 38.06/0.9607 & 33.70/0.9187 & 32.20/0.8999 & 32.25/0.9288  & 38.94/0.9774 \\
				~ & RFDN-L ~\cite{rfdn}  &626K   &146G  & \textcolor{blue}{38.08}/0.9606  & 33.67  /0.9190  & 32.18/0.8996  & 32.24/0.9290  & 38.95/0.9773 \\
				~ & SRPN-Lite~\cite{srpn}       &609K  &140G  &\textcolor{red}{38.10}/0.9608 & 33.70/0.9189 & 32.25/0.9005& 32.26/0.9294 & -  \\
				~ & HNCT ~\cite{hnct}       &357K     &82G & \textcolor{blue}{38.08}/0.9608  & 33.65/0.9182  & 32.22/0.9001  & 32.22/0.9294  & 38.87/0.9774 \\
				~ & FMEN~\cite{fmen}  &748K     &172G  & \textcolor{red}{38.10}/\textcolor{blue}{0.9609}  & 33.75/0.9192  & 32.26/0.9007  & 32.41/0.9311  & 38.95/\textcolor{blue}{0.9778} \\
				~ & NGswin~\cite{ngswin}    &990K     &140G &38.05/\textcolor{red}{0.9610} & \textcolor{red}{33.79}/\textcolor{blue}{0.9199} & \textcolor{blue}{32.27}/\textcolor{blue}{0.9008} & \textcolor{red}{32.53}/\textcolor{red}{0.9324} & \textcolor{blue}{38.97}/0.9777 \\
                ~ & \textbf{LMLT-Base(Ours)} & 652K & 158G  & \textcolor{red}{38.10}/\textcolor{red}{0.9610} & \textcolor{blue}{33.76}/\textcolor{red}{0.9201} & \textcolor{red}{32.28}/\textcolor{red}{0.9012} & \textcolor{blue}{32.52}/\textcolor{blue}{0.9316} & \textcolor{red}{39.24}/\textcolor{red}{0.9783} \\

                \cline{2-9}
                & ESRT~\cite{esrt}        & 751K & - &38.03/0.9600&33.75/0.9184&32.25/0.9001&32.58/0.9318&39.12/0.9774\\
				~ & SwinIR-light~\cite{SwinIR}     &910K & 244G  &38.14/0.9611&33.86/0.9206&32.31/0.9012&\textcolor{blue}{32.76}/\textcolor{blue}{0.9340}&39.12/0.9783 \\
                ~ & ELAN~\cite{elan} &621K &203G &38.17/0.9611&\textcolor{blue}{33.94}/0.9207&32.30/0.9012&\textcolor{blue}{32.76}/\textcolor{blue}{0.9340}&39.11/0.9782 \\ 
                ~ &SRformer-Light~\cite{srformer}       & 853K & 236G & \textcolor{red}{38.23}/\textcolor{red}{0.9613}&\textcolor{blue}{33.94}/\textcolor{blue}{0.9209}&\textcolor{red}{32.36}/\textcolor{red}{0.9019}&\textcolor{red}{32.91}/\textcolor{red}{0.9353}&\textcolor{blue}{39.28}/\textcolor{blue}{0.9785} \\ 
                ~ & \textbf{LMLT-Large(Ours)} & 1,270K & 306G& \textcolor{blue}{38.18}/\textcolor{blue}{0.9612} & \textcolor{red}{33.96}/\textcolor{red}{0.9212} & \textcolor{blue}{32.33}/\textcolor{blue}{0.9017} & 32.75/0.9336 & \textcolor{red}{39.41}/\textcolor{red}{0.9786} \\
				\hline
				\hline
    
				\multirow{13}*{$\times 3$} 
                & IMDN ~\cite{IMDN}         &703K     &72G  & 34.36/0.9270 & 30.32/0.8417 & 29.09/0.8046 & 28.17/0.8519 & 33.61/0.9445 \\
				~ & LatticeNet~\cite{Lattice}           &765K   &76G   &34.40/0.9272 & 30.32/0.8416 & 29.10/0.8049 & 28.19/0.8513 & 33.63/0.9442 \\
				~ & RFDN-L~\cite{rfdn}  &633K   &66G   &34.47/0.9280  &30.35/0.8421  &29.11/0.8053  &28.32/0.8547  &33.78/0.9458 \\
				~ & SRPN-Lite~\cite{srpn}       &615K    &63G  & 34.47/0.9280 & 30.38/0.8425 & 29.16/0.8061 & 28.22/0.8534 & - \\
				~ & HNCT ~\cite{hnct}       &363K     &38G & 34.47/0.9275  & \textcolor{blue}{30.44}/0.8439  & 29.15/0.8067  & 28.28/0.8557  & 33.81/0.9459 \\
				~ & FMEN~\cite{fmen}  &757K     &77G  &34.45/0.9275  & 30.40/0.8435  & 29.17/0.8063  & 28.33/0.8562  & 33.86/0.9462 \\
				~ & NGswin~\cite{ngswin}           &1,007K     &67G  & \textcolor{blue}{34.52}/\textcolor{blue}{0.9282} & \textcolor{red}{30.53}/\textcolor{blue}{0.8456} & \textcolor{blue}{29.19}/\textcolor{blue}{0.8078} & \textcolor{red}{28.52}/\textcolor{red}{0.8603} & \textcolor{blue}{33.89}/\textcolor{blue}{0.9470} \\

                ~ & \textbf{LMLT-Base(Ours)} & 660K & 75G  & \textcolor{red}{34.58}/\textcolor{red}{0.9285} & \textcolor{red}{30.53}/\textcolor{red}{0.8458} & \textcolor{red}{29.21}/\textcolor{red}{0.8084} & \textcolor{blue}{28.48}/\textcolor{blue}{0.8581} & \textcolor{red}{34.18}/\textcolor{red}{0.9477} \\

                \cline{2-9}
                & ESRT~\cite{esrt}    & 751K & - &34.42/0.9268&30.43/0.8433&29.15/0.8063&28.46/0.8574&33.95/0.9455\\
                ~ & SwinIR-light~\cite{SwinIR}   &918K & 111G &34.62/0.9289&30.54/0.8463&29.20/0.8082&28.66/0.8624&33.98/0.9478 \\
                ~ & ELAN~\cite{elan} &629K &90G &34.61/0.9288&30.55/0.8463&\textcolor{blue}{29.21}/0.8081&28.69/0.8624&34.00/0.9478 \\ 
                ~ &SRformer-Light~\cite{srformer}  & 861K & 105G & \textcolor{red}{34.67}/\textcolor{red}{0.9296}&\textcolor{blue}{30.57}/\textcolor{blue}{0.8469}&\textcolor{red}{29.26}/\textcolor{red}{0.8099}&\textcolor{red}{28.81}/\textcolor{red}{0.8655}&\textcolor{blue}{34.19}/\textcolor{blue}{0.9489} \\
                ~ & \textbf{LMLT-Large(Ours)} & 1,279K & 144G& \textcolor{blue}{34.64}/\textcolor{blue}{0.9293} & \textcolor{red}{30.60}/\textcolor{red}{0.8471} & \textcolor{red}{29.26}/\textcolor{blue}{0.8097} & \textcolor{blue}{28.72}/\textcolor{blue}{0.8626} & \textcolor{red}{34.43}/\textcolor{red}{0.9491} \\
    		\hline
				\hline
    
				\multirow{13}*{$\times 4$} 
                & IMDN ~\cite{IMDN}         &715K     &41G & 32.21/0.8948 & 28.58/0.7811 & 27.56/0.7353 & 26.04/0.7838 & 30.45/0.9075 \\
				~ & LatticeNet~\cite{Lattice}           &777K   &44G   & 32.18/0.8943 & 28.61/0.7812 & 27.57/0.7355 & 26.14/0.7844 & 30.54/0.9075 \\
				~ & RFDN-L ~\cite{rfdn}  &643K   &38G  & 32.28/0.8957 & 28.61/0.7818 & 27.58/0.7363 & 26.20/0.7883 & 30.61/0.9096 \\
                ~ & SRPN-Lite~\cite{srpn}       &623K    &36G  & 32.24/0.8958 & 28.69/0.7836 & 27.63/0.7373 & 26.16/0.7875 & - \\
				~ & HNCT ~\cite{hnct}       &373K     &22G & 32.31/0.8957  & 28.71/0.7834  & 27.63/0.7381  & 26.20/0.7896  & 30.70/0.9112 \\
				~ & FMEN~\cite{fmen}  &769K     &44G & 32.24/0.8955  & 28.70/\textcolor{blue}{0.7839}  & 27.63/0.7379  & 26.28/0.7908  & 30.70/0.9107 \\
				~ & NGswin~\cite{ngswin}           &1,019K     &36G & \textcolor{blue}{32.33}/\textcolor{blue}{0.8963} &\textcolor{blue}{28.78}/\textcolor{red}{0.7859} & \textcolor{blue}{27.66}/\textcolor{blue}{0.7396} & \textcolor{red}{26.45}/\textcolor{red}{0.7963} &  \textcolor{blue}{30.80}/\textcolor{blue}{0.9128} \\
                ~ & \textbf{LMLT-Base(Ours)} & 672K & 41G & \textcolor{red}{32.38}/\textcolor{red}{0.8971} & \textcolor{red}{28.79}/\textcolor{red}{0.7859} & \textcolor{red}{27.70}/\textcolor{red}{0.7403} & \textcolor{blue}{26.44}/\textcolor{blue}{0.7947} & \textcolor{red}{31.09}/\textcolor{red}{0.9139} \\
                \cline{2-9}


                & ESRT~\cite{esrt}  & 751K & - &32.19/0.8947&28.69/0.7833&27.69/0.7379&26.39/0.7962&30.75/0.9100\\
				~ & SwinIR-light~\cite{SwinIR}  &930K & 64G &32.44/0.8976&28.77/0.7858&27.69/0.7406&26.47/0.7980&30.92/0.9151 \\
				~ & ELAN~\cite{elan} &640K &54G &32.43/0.8975&28.78/0.7858&27.69/0.7406&26.54/0.7982&30.92/0.9150 \\ 
                ~ &SRformer-Light~\cite{srformer}  & 873K & 63G & \textcolor{red}{32.51}/\textcolor{red}{0.8988} &\textcolor{blue}{28.82}/\textcolor{blue}{0.7872} &\textcolor{blue}{27.73}/\textcolor{red}{0.7422}&\textcolor{red}{26.67}/\textcolor{red}{0.8032}&\textcolor{blue}{31.17}/\textcolor{red}{0.9165}\\
                ~ & \textbf{LMLT-Large(Ours)} & 1,295K & 78G& \textcolor{blue}{32.48}/\textcolor{blue}{0.8987} & \textcolor{red}{28.87}/\textcolor{red}{0.7879} & \textcolor{red}{27.75}/\textcolor{blue}{0.7421} & \textcolor{blue}{26.63}/\textcolor{blue}{0.8001} & \textcolor{red}{31.32}/\textcolor{blue}{0.9163} \\
				\hline
    
				\hline
		\end{tabular}}
	\end{center}
\end{table}

We first compare the LMLT-Base with IMDN~\cite{IMDN}, LatticeNet~\cite{Lattice}, RFDN-L~\cite{rfdn}, SRPN-Lite~\cite{srpn}, HNCT~\cite{hnct}, FMEN~\cite{fmen}, and NGswin~\cite{ngswin}. Table~\ref{tab:baselarge} shows that our LMLT-Base achieves the best or second-best performance on most benchmark datasets. Notably, we observe a significant performance increase on the Manga109 dataset, while our model uses up to 30\% fewer parameters compared to the next highest performing model, NGswin~\cite{ngswin}, while the PSNR increases by 0.27dB, 0.29dB, and 0.29dB, respectively, at all scales.

Next, we compare the LMLT-Large model with other SR models. The comparison group includes ESRT~\cite{esrt}, SwinIR-light~\cite{SwinIR}, ELAN~\cite{elan}, and SRFormer-light~\cite{srformer}. As shown in Table~\ref{tab:baselarge}, our large model ranks first or second in performance on most datasets. Among the five test datasets, LMLT-Large shows the best performance for all scales on the Manga109~\cite{Manga109} dataset compared to others, and also the best performance on the Set14~\cite{Set14} dataset. Specifically, compared to SRFormer-light~\cite{srformer}, which showed the highest performance on Urban100~\cite{Urban100} among the comparison group, our model shows performance gains of 0.13dB, 0.24dB, and 0.15dB at each scale on the Manga109~\cite{Manga109} dataset. In addition to this, we demonstrate that our model has a significant advantage in inference time and GPU memory occupancy at next paragraph. The comparison results of LMLT-Tiny and LMLT-Small with other state-of-the-art models can be found in Appendix~\ref{App:reconstruction}.

\textbf{Memory and Running Time Comparisons.} To test the efficiency of the proposed model, we compare the performance of our LMLT model against other ViT-based state-of-the-art super-resolution models at different scales. We evaluate LMLT-Base against NGswin~\cite{ngswin} and HNCT~\cite{hnct}, and LMLT-Large against SwinIR-light~\cite{SwinIR} and SRFormer-light~\cite{srformer}. The results are shown in Table~\ref{tab:memtime}.

\begin{table}[t]
	\caption{The memory consumption and inference times are reported. A single RTX 3090 GPU is used.}
	\label{tab:memtime}
	\renewcommand\arraystretch{1.1}
	\begin{center}
		\resizebox{\textwidth}{!}{
			\begin{tabular}{| c | c | c | c | c | c | c | c | c |}
				\hline
				Scale & Method & \#GPU Mem [M] & \#Avg Time [ms] & Set5 & Set14 & B100 & Urban100 & Manga109 \\
				\hline
    
				\multirow{8}*{$\times 2$} 
                &\textbf{LMLT-Tiny(Ours)}    & \textbf{324.01} & \textbf{57.37} & 38.01/0.9606 & 33.59/0.9183 & 32.19/0.8999 & 32.04/0.9273 & 38.90/0.9775 \\ 
                ~ &\textbf{LMLT-Small(Ours)}       & \textbf{324.5} & \textbf{84.22} & 38.05/0.9608 & 33.65/0.9187 & 32.24/0.9006 & 32.31/0.9298 & 39.10/0.9780 \\  
                \cline{2-9}

                & HNCT~\cite{hnct}     &1200.55 & 351.49 & 38.08/0.9608  & 33.65/0.9182  & 32.22/0.9001  & 32.22/0.9294  & 38.87/0.9774 \\
                ~ & NGswin~\cite{ngswin} &1440.40 &375.19 &38.05/0.9610 & 33.79/0.9199 & 32.27/0.9008 & 32.53/0.9324 & 38.97/0.9777 \\
                ~ &\textbf{LMLT-Base(Ours)}       &  \textbf{567.75} &  \textbf{81.64} & 38.10/0.9610 & 33.76/0.9201 & 32.28/0.9012 & 32.52/0.9316 & 39.24/0.9783 \\ 
                \cline{2-9} 

                ~ & SwinIR-light~\cite{SwinIR}  &1278.64 & 944.11 &38.14/0.9611&33.86/0.9206&32.31/0.9012&32.76/0.9340&39.12/0.9783 \\
                ~ &SRformer-Light~\cite{srformer}  & 1176.15 & 1006.48 & 38.23/0.9613&33.94/0.9209&32.36/0.9019&32.91/0.9353&39.28/0.9785 \\ 
                ~ & \textbf{LMLT-Large(Ours)} &  \textbf{717.31} &  \textbf{123.07} & 38.18/0.9612 & 33.96/0.9212 & 32.33/0.9017 & 32.75/0.9336 & 39.41/0.9786 \\
    		\hline    
                \hline
                
    		  \multirow{8}*{$\times 3$} 
                &\textbf{LMLT-Tiny(Ours)}    & \textbf{151.96} & \textbf{31.06} & 34.36/0.9271 & 30.37/0.8427 & 29.12/0.8057 & 28.10/0.8503 & 33.72/0.9448 \\
                ~ &\textbf{LMLT-Small(Ours)}       & \textbf{152.5} & \textbf{44.22} & 34.50/0.9280 & 30.47/0.8446 & 29.16/0.8070 & 28.29/0.8544 & 33.99/0.9464 \\ 
                \cline{2-9}

                & HNCT~\cite{hnct}     &545.64 & 117.20 & 34.47/0.9275  & 30.44/0.8439  & 29.15/0.8067  & 28.28/0.8557  & 33.81/0.9459 \\
                ~ & NGswin~\cite{ngswin} &696.97 &168.49 &34.52/0.9282& 30.53/0.8456 & 29.19/0.8078 &28.52/0.8603 & 33.89/0.9470 \\
                ~ &\textbf{LMLT-Base(Ours)}       & \textbf{266.31} & \textbf{41.43} & 34.58/0.9285 & 30.53/0.8458 & 29.21/0.8084 & 28.48/0.8581 & 34.18/0.9477 \\ 
                \cline{2-9}

                ~ & SwinIR-light~\cite{SwinIR}  &587.63 & 287.96 &34.62/0.9289&30.54/0.8463&29.20/0.8082&28.66/0.8624&33.98/0.9478 \\
                ~ &SRformer-Light~\cite{srformer}  & 529.28 & 312.37 & 34.67/0.9296&30.57/0.8469&29.26/0.8099&28.81/0.8655&34.19/0.9489 \\
                ~ & \textbf{LMLT-Large(Ours)} & \textbf{338.36} & \textbf{58.68} & 34.64/0.9293 & 30.60/0.8471 & 29.26/0.8097 & 28.72/0.8626 & 34.43/0.9491 \\
    		\hline    
                \hline
    
				\multirow{8}*{$\times 4$}
                &\textbf{LMLT-Tiny(Ours)}    & \textbf{81.44} & \textbf{23.54} & 32.19/0.8947 &  28.64/0.7823 & 27.60/0.7369 & 26.08/0.7838 & 30.60/0.9083 \\
                ~ &\textbf{LMLT-Small(Ours)}       & \textbf{81.92} & \textbf{31.01} & 32.31/0.8968 & 28.74/0.7846 & 27.66/0.7387 & 26.26/0.7894 & 30.87/0.9117 \\ 
                \cline{2-9} 

               & HNCT~\cite{hnct}     &312.72 & 69.61 & 32.31/0.8957  & 28.71/0.7834  & 27.63/0.7381  &26.20/0.7896  & 30.70/0.9112 \\
                ~ & NGswin~\cite{ngswin} &372.94 &118.13 & 32.33/0.8963 & 28.78/0.7859 & 27.66/0.7396& 26.45/0.7963 &  30.80/0.9128 \\
                ~ &\textbf{LMLT-Base(Ours)}       & \textbf{144.00} & \textbf{26.15} & 32.38/0.8971 & 28.79/0.7859 & 27.70/0.7403 & 26.44/0.7947 & 31.09/0.9139 \\ 
                \cline{2-9}

                ~ & SwinIR-light~\cite{SwinIR}  &342.46 & 176.76  &32.44/0.8976&28.77/0.7858&27.69/0.7406&26.47/0.7980&30.92/0.9151 \\
                ~ &SRformer-Light~\cite{srformer}  & 320.95 & 180.42 & 32.51/0.8988&28.82/0.7872&27.73/0.7422&26.67/0.8032&31.17/0.9165\\
                ~ & \textbf{LMLT-Large(Ours)} & \textbf{185.68} & \textbf{34.07} & 32.48/0.8987 & 28.87/0.7879 & 27.75/0.7421 & 26.63/0.8001 & 31.32/0.9163 \\
    		\hline    

		\end{tabular}}
	\end{center}
\end{table}

We observe that LMLT-Base and LMLT-Large is quite efficient in terms of inference speed and memory usage compared to other ViT-based SR models. Specifically, compared to NGswin~\cite{ngswin}, our LMLT-Base maintains similar performance while reducing memory usage by 61\%, 62\%, and 61\% for $\times2$, $\times3$, and $\times4$ scales, respectively, and decreasing inference time by an average of 78\%, 76\%, and 78\%. Similarly, when comparing SwinIR~\cite{SwinIR} and our LMLT-Large, despite maintaining similar performance, memory usage decreases by 44\%, 43\%, and 46\% for each scale, respectively, and inference time decreases by an average of 87\%, 80\%, and 81\%. This demonstrates both the efficiency and effectiveness of the proposed model.

\begin{wraptable}{R}{0.5\textwidth}
\centering
	\caption{Time consumption from LN to LHSB (Ours) and WSA (SwinIR~\cite{SwinIR}). A RTX 3090 GPU is used.}
	\label{tab:moduletime}
        \renewcommand\arraystretch{1.1}
        \resizebox{0.5\textwidth}{!}{
    	\begin{tabular}{ c | c c c }
    		\toprule
    		method        & $\times 2$         & $\times 3$   & $\times 4$  \\
    		\midrule
    		LMLT-Tiny(Ours) & 35.28$\mathit{ms}$  &  23.21$\mathit{ms}$  & 18.36$\mathit{ms}$   \\
    		LMLT-Base(Ours) & 49.44$\mathit{ms}$ &  25.51$\mathit{ms}$   & 21.18$\mathit{ms}$   \\
    		LMLT-Large(Ours) & 68.97$\mathit{ms}$ &  32.72$\mathit{ms}$  & 22.66$\mathit{ms}$   \\
            SwinIR-Light~\cite{SwinIR} & 1084.57$\mathit{ms}$  & 336.00$\mathit{ms}$   & 185.23$\mathit{ms}$  \\
    		\bottomrule
	       \end{tabular}}
\end{wraptable}

Table~\ref{tab:moduletime} shows the time consumption for module in the proposed method and SwinIR-light~\cite{SwinIR}, specifically detailing the time from the first Layer Normalization (LN)~\cite{LN} to the LHSB and WSA. Our LMLT significantly reduces the time required for the self-attention mechanism. Especially, LMLT-Large achieves time reductions of 94\% at the $\times2$ scale, 90\% at the $\times3$ scale, and 88\% at the $\times4$ scale compared to SwinIR. Given the similar performance between SwinIR and our method, this represents a significant increase in efficiency. Note that the inference time for a total of 50 random images is measured. The time measurements are conducted using \texttt{torch.cuda.Event}'s record and \texttt{elapsed\_time()}, and due to hardware access latency during log output, the time of the modules might be longer than the time report in Table~\ref{tab:memtime}. Tables comparing the memory usage and inference speed of our LMLT with other models can be found in Appendix~\ref{App:memtime}.

\begin{figure*}[!t]\footnotesize
 \begin{center}
  \begin{tabular}{ccccccc}
  
      \multicolumn{3}{c}{\multirow{5}*[53pt]{
        \includegraphics[width=0.3\linewidth, height=0.33\linewidth]{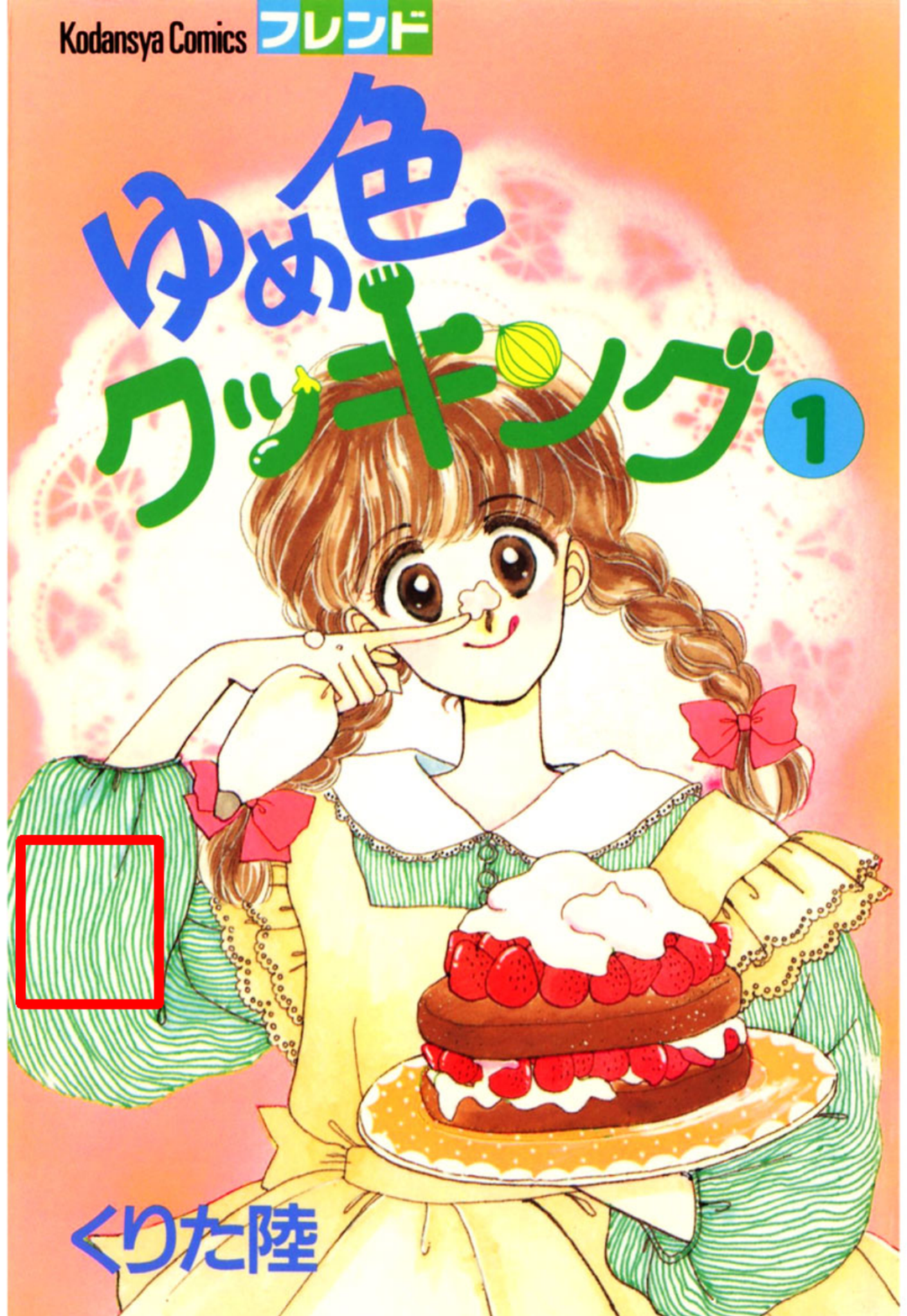}}} &\hspace{-3.5mm}
      \includegraphics[width=0.15\linewidth, height = 0.15\linewidth]{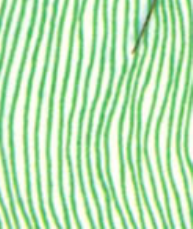} &\hspace{-2mm}
      \includegraphics[width=0.15\linewidth, height = 0.15\linewidth]{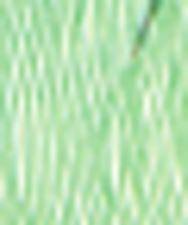} &\hspace{-2mm}
      \includegraphics[width=0.15\linewidth, height = 0.15\linewidth]{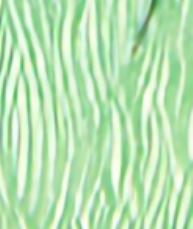} &\hspace{-2mm}
      \includegraphics[width=0.15\linewidth, height = 0.15\linewidth]{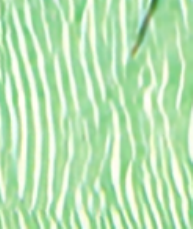} \\
      \multicolumn{3}{c}{~} &\hspace{-3.5mm}(a) GT  &\hspace{-3.5mm}(b) Bicubic &\hspace{-3.5mm}(c) IMDN &\hspace{-3.5mm}(d) NGswin \\

      \multicolumn{3}{c}{~} & \hspace{-3.5mm}
      \includegraphics[width=0.15\linewidth, height = 0.15\linewidth]{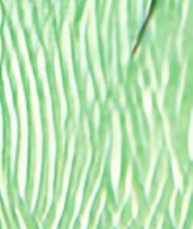} & \hspace{-2mm}
      \includegraphics[width=0.15\linewidth, height = 0.15\linewidth]{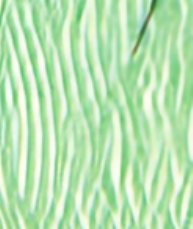} & \hspace{-2mm}
      \includegraphics[width=0.15\linewidth, height = 0.15\linewidth]{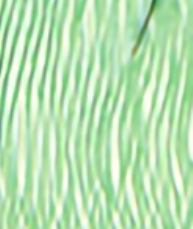} & \hspace{-2mm}
      \includegraphics[width=0.15\linewidth, height = 0.15\linewidth]{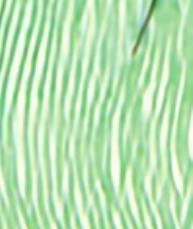} \\
      \multicolumn{3}{c}{\hspace{-3.5mm} YumeiroCooking from Manga109} &\hspace{-3.5mm}(e) SwinIR-light   &\hspace{-3.5mm}(f) SwinIR-NG  &\hspace{-3.5mm}(g) LMLT-Base  &\hspace{-3.5mm}(h) LMLT-Large
      \\
      
 \end{tabular}
 \end{center}
 \caption{Visual comparisons for $\times 4$ SR on Manga109 dataset. Compared with the results in (c) to (f), the Ours(LMLT-Base(g), LMLT-Large(h)) restore much more accurate and clear images. More results are in the Appendix~\ref{App:qualitative}.}
 \label{fig:sec4_compare}
\end{figure*}

\textbf{Qualitative Comparisons.} Figure~\ref{fig:sec4_compare} illustrates the differences on the Manga109~\cite{Manga109} dataset between our model and other models. As shown, our LMLT successfully reconstructs areas with continuous stripes better than other models. Additionally, we include more comparison images, and further compare our proposed models LMLT-Tiny with CARN~\cite{CARN}, EDSR~\cite{Flickr2K-EDSR}, PAN~\cite{PAN}, ShuffleMixer~\cite{ShuffleMixer} and SAFMN~\cite{SAFMN} on the Urban100~\cite{Urban100} dataset at $\times4$ scale. Detailed results can be seen in Appendix~\ref{App:qualitative}.

\subsection{Ablation Study}

\textbf{Effects of Low-to-high Connection.} We examine the effects of the low-to-high element-wise sum (low-to-high connection) and downsizing elements of our proposed model. As shown in Table~\ref{tab:nosum}, the low-to-high connection yields significant results. Specifically, on the Urban100~\cite{Urban100} dataset, PSNR increases by 0.04 dB to 0.05 dB across all scales, and SSIM~\cite{SSIM} increases by nearly 0.0011 at the $\times4$ scale, demonstrating the benefits of including the low-to-high connection. Appendix~\ref{App:low-to-high} visualizes the differences in features on Urban100~\cite{Urban100} with and without the low-to-high connection, showing that it significantly reduces the boundary lines between windows. Additionally, experiments adding the proposed low-to-high connection to SAFMN~\cite{SAFMN} for $\times2$, $\times3$, and $\times4$ scales are also provided in Appendix~\ref{App:low-to-high}.

\begin{table}[t]
	\caption{Performance results when the low-to-high element-wise sum removed. Better results are highlighted.}
	\label{tab:nosum}
	\renewcommand\arraystretch{1.1}
	\begin{center}
		\resizebox{\textwidth}{!}{
			\begin{tabular}{| c | c | c | c |  c | c | c |}
				\hline
				Scale & Method & Set5 & Set14 & B100 & Urban100 & Manga109 \\
				\hline
    
				\multirow{2}*{$\times 2$} 
                & LMLT-Tiny & \textbf{38.01}/0.9606 & \textbf{33.59}/\textbf{0.9183} & \textbf{32.19}/0.8999 & \textbf{32.04}/\textbf{0.9273} & \textbf{38.90}/0.9775 \\ 
                ~ & LMLT-Tiny $\mathit{w/o}$ sum & 38.00/0.9606 &33.58/0.9181 &32.18/0.8999 &31.99/0.9268&38.88/0.9775 \\ 
				\hline

    		  \multirow{2}*{$\times 3$} 
                & LMLT-Tiny & 34.36/0.9271 & \textbf{30.37}/\textbf{0.8427} & \textbf{29.12/0.8057} & \textbf{28.10/0.8503} & \textbf{33.72}/0.9448 \\
                ~ & LMLT-Tiny$\mathit{w/o}$ sum & \textbf{34.40}/\textbf{0.9272} &30.35/0.8425 &29.11/0.8056 &28.05/0.8496&33.71/0.9448 \\
    		\hline
    
				\multirow{2}*{$\times 4$}
                & LMLT-Tiny & \textbf{32.19/0.8947} &  28.64/0.7823 & 27.60/\textbf{0.7369} & \textbf{26.08/0.7838} & 30.60/\textbf{0.9083} \\
                ~ & LMLT-Tiny $\mathit{w/o}$ sum & 32.16/0.8944 &28.64/0.7823 &27.60/0.7368 &26.04/0.7827& \textbf{30.64}/0.9078 \\
				\hline
		\end{tabular}}
	\end{center}
\end{table}

\begin{table}[t]
	\caption{Performance with or without pooling and merging. Best results are highlighted in \textbf{bold}.}
	\label{tab:nopool}
	\renewcommand\arraystretch{1.1}
	\begin{center}
		\resizebox{\textwidth}{!}{
			\begin{tabular}{| c | c | c | c | c | c | c | c | c | c |}
				\hline
				Scale & Method & \#Params & \#FLOPs &\#Acts & Set5 & Set14 & B100 & Urban100 & Manga109 \\
				\hline
    
				\multirow{3}*{$\times 2$} 
                & LMLT-Tiny & 239K & 59G & 603M &  \textbf{38.01}/\textbf{0.9606} & \textbf{33.59}/\textbf{0.9183} & \textbf{32.19}/\textbf{0.8999} & \textbf{32.04}/\textbf{0.9273} & \textbf{38.90}/\textbf{0.9775} \\ 
                ~ & LMLT-Tiny $\mathit{w/o}$ pool & 239K & 67G & 1223M & 37.98/0.9605 &33.56/0.9178 &32.16/0.8996 &31.87/0.9255 &38.79/0.9773 \\ 
                ~ & LMLT-Tiny $\mathit{w/o}$ pool and merge & 229K & 64G & 1152M & 37.95/0.9604 &33.51/0.9173 &32.14/0.8993 &31.76/0.9245 &38.68/0.9771 \\ 
				\hline

    		  \multirow{3}*{$\times 3$} 
                & LMLT-Tiny  &244K &28G &283M & \textbf{34.36}/\textbf{0.9271} & \textbf{30.37}/\textbf{0.8427} & \textbf{29.12/0.8057} & \textbf{28.10/0.8503} & \textbf{33.72}/\textbf{0.9448} \\
                ~ & LMLT-Tiny $\mathit{w/o}$ pool & 244K & 32G & 572M & \textbf{34.36}/0.9270&30.34/0.8421 &29.10/0.8051 &28.02/0.8488 &33.66/0.9445 \\
                ~ & LMLT-Tiny $\mathit{w/o}$ pool and merge & 234K & 31G &539M & 34.28/0.9265 &30.31/0.8417 &29.07/0.8044 &27.94/0.8467&33.55/0.9438 \\ 
    		\hline
    
				\multirow{3}*{$\times 4$}
                & LMLT-Tiny & 251K & 15G & 152M & \textbf{32.19/0.8947} &  \textbf{28.64}/\textbf{0.7823} & \textbf{27.60}/\textbf{0.7369} & \textbf{26.08/0.7838} & \textbf{30.60}/\textbf{0.9083} \\
                ~ & LMLT-Tiny $\mathit{w/o}$ pool & 251K & 17G & 308M  & 32.12/0.8940 &28.61/0.7820 &27.58/0.7362 &26.01/0.7815& 30.51/0.9074 \\
                ~ & LMLT-Tiny $\mathit{w/o}$ pool and merge &240K & 17G & 290M & 32.07/0.8934 &28.60/0.7817 &27.56/0.7355 &25.95/0.7795 &30.45/0.9064 \\ 
				\hline
		\end{tabular}}
	\end{center}
\end{table}

\textbf{Effects of Multi-Scale Heads.} Table~\ref{tab:nopool} validates the effectiveness of using multiple scales by experimenting with cases where pooling is not applied to any head. Specifically, we compare our proposed LMLT with cases where pooling and the low-to-high connection are not applied, as well as cases where merging is also not performed. The results show that performance is lower in all cases compared to the proposed model. Appendix~\ref{App:pool} demonstrates that when pooling is not applied, the lack of information connection between windows hinders the proper capture of informative features, even though spatial information is retained.

\textbf{Importance of LHSB, CCM, and MLP.} We analyze the impact of LHSB and CCM~\cite{SAFMN} and their interplay. Following the approach in SAFMN~\cite{SAFMN}, we examine performance by individually removing LHSB and CCM~\cite{SAFMN}. Results are shown in the `Module' row of Table~\ref{tab:each_module}. Removing LHSB reduces the number of parameters by nearly 10\%, decreases memory usage to nearly 74\%, and drops PSNR by 0.59 dB on the Urban100~\cite{Urban100} dataset. Conversely, removing CCM reduces the number of parameters by nearly 90\% and PSNR by 1.83 dB. Adding an MLP after the self-attention module, as done in traditional Transformers~\cite{Attn, vit}, reduces parameters by about 69\% and PSNR by approximately 0.85 dB. To maintain the same number of layers as in the previous two experiments (8 layers), we conducted another experiment with only 4 blocks, resulting in a 49\% reduction in parameters and only a 0.55 dB drop in PSNR, indicating the least performance loss. This suggests that the combination of LHSB and CCM effectively extracts features. Additionally, incorporating FMBConv~\cite{effnetv2} reduces parameters by nearly 58\% and PSNR by 0.28 dB, while memory usage remains similar.

\textbf{Importance of Aggregation and Activation.} We analyze the impact of aggregating features from each head using a $1\times1$ convolution or applying activation before multiplying with the original input. Results are shown in the `Act / Aggr' row of Table~\ref{tab:each_module}. Without aggregation, PSNR decreases by 0.10 dB on the Urban100~\cite{Urban100} dataset. If features are directly output without applying activation and without multiplying with the original input, PSNR decreases by 0.12 dB. Omitting both steps leads to an even greater decrease of 0.22 dB, indicating that including both aggregation and activation is more efficient. Conversely, multiplying features directly to the original feature without the activation function improves performance by 0.1 dB. Detailed experimental results are discussed in Appendix~\ref{App:each_module}.

\textbf{Importance of Positional Encoding.} Lastly, we examine the role of Positional Encoding (PE) in performance improvement. Results are shown in the `PE' row of Table~\ref{tab:each_module}. Removing PE results in decreased performance across all benchmark datasets, notably with a PSNR drop of 0.06 dB and an SSIM decrease of 0.0006 on the Urban100~\cite{Urban100} dataset. Using RPE~\cite{swin} results in a maximum PSNR increase of 0.03 dB on the Set14~\cite{Set14} dataset, but has little effect on other datasets. Additionally, parameters and GPU memory increase by 5K and 45M, respectively.

\begin{table}[t]
	\caption{Ablation studies on each component of our method at scale $\times 2$. LMLT-Tiny is used. }
	\label{tab:each_module}
	\renewcommand\arraystretch{1.1}
	\begin{center}
		\resizebox{\textwidth}{!}{
			\begin{tabular}{| c | c | c | c |  c | c | c | c| c| c|}
				\hline
				Ablation & Variants & \#Param & \#Flops & \#GPU Mem & Set5 & Set14 & B100 & Urban100 & Manga109 \\
				\hline
                \multirow{1}*{\textbf{Baseline}} 
                & -  & \textbf{239K} & \textbf{59G} & \textbf{324M} & \textbf{38.01/0.9606} & \textbf{33.59/0.9183} &\textbf{32.19/0.8999} & \textbf{32.04/0.9273} & \textbf{38.90/0.9775} \\
				\hline
    
				\multirow{5}*{Module} 
                & LHSB $\rightarrow$ None & 214K & 52G & 241M & 37.88/0.9601 & 33.39/0.9160 & 32.05/0.8981 &  31.45/0.9210 & 38.40/0.9765 \\
                ~& CCM $\rightarrow$ None & 31K & 8G & 323M & 37.26/0.9573  & 32.85/0.9113  & 31.64/0.8927 & 30.21/0.9071 & 36.91/0.9720 \\
                ~& CCM $\rightarrow$ MLP  & 73K & 18G & 324M & 37.71/0.9593  & 33.25/0.9150  & 31.96/0.8968 & 31.19/0.9187 & 38.03/0.9754 \\
                ~& CCM $\rightarrow$ IVF  & 101K & 19G & 324M & 37.91/0.9603  & 33.46/0.9173  & 32.11/0.8990 & 31.71/0.9243 & 38.62/0.9770 \\
                ~& 4 Blocks & 122K & 30G & 307M &  37.88/0.9601 & 33.40/0.9166 & 32.06/0.8984 &31.49/0.9217 & 38.47/0.9765 \\ 
				\hline

    		  \multirow{4}*{Act / Aggr} 
                & No Aggregation  & 229K & 56G & 324M & 37.99/0.9606 & 33.55/0.9178 & 32.17/0.8997 &  31.94/0.9262 & 38.84/0.9774 \\
                ~& No Activation  & 239K & 59G & 324M & 37.99/0.9605  & 33.55/0.9180  & 32.16/0.8997 & 31.92/0.9260 & 38.83/0.9774 \\
                ~& No Aggr, No Act  & 229K & 56G & 324M & 37.95/0.9606  & 33.53/0.9175  & 32.15/0.8994 & 31.82/0.9250 & 38.73/0.9771 \\
                ~& GELU $\rightarrow$ None  & 239K & 59G & 324M & 38.03/0.9606  & 33.60/0.9184  & 32.19/0.9000 & 32.05/0.9272 & 38.91/0.9776 \\
    		\hline
    
				\multirow{2}*{PE}
                & No PE & 236K & 59G & 309M & 37.98/0.9606 & 33.55/0.9176 & 32.18/0.8998 &  31.98/0.9267 & 38.86/0.9774 \\
                ~& LePE $\rightarrow$ RPE\cite{swin}  & 244K & 59G & 369M & 38.02/0.9606  & 33.62/0.9182  & 32.20/0.9000 & 32.05/0.9275 & 38.90/0.9775 \\
				\hline
		\end{tabular}}
	\end{center}
\end{table}

\section{Conclusion}

In this paper, we introduced the Low-to-high Multi-Level Transformer (LMLT) for efficient image super-resolution. By combining multi-head and depth reduction, our model addresses the excessive computational load and memory usage of traditional ViT models. In addition to this, LMLT applies self-attention to features at various scales, aggregating lower head outputs to inform higher heads, thus solving the cross-window communication issue. Our extensive experiments demonstrate that LMLT achieves a favorable balance between model complexity and performance, significantly reducing memory usage and inference time while maintaining or improving image reconstruction quality. This makes the proposed LMLT a highly efficient solution for image super resolution tasks, suitable for deployment on resource-constrained devices.

\small
{
\bibliographystyle{plainnat}
\bibliography{main_ref}    
}

\appendix
\newpage
\setcounter{section}{0}
\setcounter{table}{0}
\setcounter{figure}{0}
\renewcommand\thesection{\Alph{section}}
\renewcommand\thetable{\Alph{table}}
\renewcommand\thefigure{\Alph{figure}}


\section{Impact of number of Blocks, Channels, Heads and Depths}

In this section, we analyze how the performance of our proposed model changes based on the number of blocks, heads, channels and depths. 

\begin{table}[h]
	\caption{Performance difference of LMLT based on the number of blocks.}
	\label{tab:perblock}
	\renewcommand\arraystretch{1.1}
	\begin{center}
		\resizebox{\textwidth}{!}{
			\begin{tabular}{| c | c | c | c | c | c | c | c | c | c|}
				\hline
				\#Block & \#Params & \#FLOPs & \#GPU Mem & \#AVG Time & Set5 & Set14 & B100 & Urban100 & Manga109 \\
				\hline
    
                4 & 122K & 30G & 323.52M & 29.75$\mathit{ms}$ &  37.88/0.9601 & 33.40/0.9166 & 32.06/0.8984 &31.49/0.9217 & 38.47/0.9765 \\ 
                6 & 181K & 44G & 323.77M & 43.51$\mathit{ms}$ &  37.94/0.9601 & 33.52/0.9175 & 32.15/0.8995 &31.82/0.9253 & 38.74/0.9771 \\ 
                8 & 239K & 59G & 324.01M & 57.37$\mathit{ms}$ & 38.01/0.9606 & 33.59/0.9183 & 32.19/0.8999 & 32.04/0.9273 & 38.90/0.9775 \\ 
                10 & 298K & 73G & 324.26M & 70.55$\mathit{ms}$  &  38.05/0.9608 & 33.66/0.9188 & 32.22/0.9003 &32.17/0.9286 & 39.00/0.9778 \\ 
                12 & 357K & 88G & 324.5M & 84.22$\mathit{ms}$ & 38.05/0.9608 & 33.65/0.9187 & 32.24/0.9006 & 32.31/0.9298 & 39.10/0.9780 \\  

				\hline

		\end{tabular}}
	\end{center}
\end{table}
\textbf{Impact of Number of Blocks.} 
First, We evaluate the performance by varying the number of blocks to 4, 6, 8, 10, and 12. Experiments are conducted on $\times2$ scale and the performance is evaluated using benchmark datasets, and analyzed in terms of the number of parameters, FLOPs, GPU memory usage, and average inference time. 

As shown in Table~\ref{tab:perblock}, the increase in the number of parameters, FLOPs and inference time tends to be proportional to the number of blocks, and performance also gradually improves. For the Manga109~\cite{Manga109} dataset, as the number of blocks increases from 4 to 12 in increments of 2, PSNR increases by 0.27 db, 0.16 db, 0.10 db, and 0.10 db, respectively. Interestingly, despite the increase in the number of blocks from 4 to 12, the GPU memory usage remains almost unchanged. While the number of parameters nearly triples, the GPU memory usage remains stable, 323.5M to 324.5M. We observe the overall increase in PSNR with the increase in the number of blocks and designate the model with 8 blocks as LMLT-Tiny and the model with 12 blocks as LMLT-Small.

\begin{table}[h]
	\caption{Performance difference of LMLT based on the number of channels.}
	\label{tab:perchan}
	\renewcommand\arraystretch{1.1}
	\begin{center}
		\resizebox{\textwidth}{!}{
			\begin{tabular}{| c | c | c | c | c | c | c | c | c | c |}
				\hline
				\#Channel & \#Params & \#FLOPs & \#GPU MEM & \#AVG Time & Set5 & Set14 & B100 & Urban100 & Manga109 \\
				\hline
    
                24 & 109K & 27G & 255.77M & 38.34$\mathit{ms}$ & 37.91/0.9602 & 33.44/0.9169 & 32.09/0.8988 &31.62/0.9231 & 38.58/0.9768 \\ 
                
                36 & 239K & 59G & 324.01M & 57.37$\mathit{ms}$ & 38.01/0.9606 & 33.59/0.9183 & 32.19/0.8999 & 32.04/0.9273 & 38.90/0.9775 \\ 
                
                48 & 420K & 103G &460.32M & 65.66$\mathit{ms}$ & 38.06/0.9609 & 33.67/0.9189 & 32.25/0.9007 & 32.33/0.9299 & 39.14/0.9780 \\ 
                
                60 & 652K & 158G & 567.75M & 81.64$\mathit{ms}$ & 38.10/0.9610 & 33.76/0.9201 & 32.28/0.9012 &32.52/0.9316 & 39.24/0.9783 \\ 
                
                72 & 935K & 226G & 684.82M & 108.74$\mathit{ms}$ & 38.17/0.9612 & 33.83/0.9205 & 32.32/0.9016 & 32.65/0.9329 & 39.36/0.9786 \\  
                
                84 & 1,270K & 306G & 717.31M & 123.07$\mathit{ms}$ & 38.18/0.9612 & 33.96/0.9212 & 32.33/0.9017 & 32.75/0.9336 & 39.41/0.9786 \\  

				\hline

		\end{tabular}}
	\end{center}
\end{table}
\textbf{Impact of Number of Channels.}
Next, we evaluate how performance changes with the number of channels. Similar to the performance evaluation based on the number of blocks, this experiment evaluates performance using benchmark datasets, the number of parameters, FLOPs, GPU memory usage, and average inference time as performance metrics. 

As shown in Table~\ref{tab:perchan}, LMLT's performance increases with channels, along with parameters and FLOPs. However, unlike the variations in the number of blocks, increasing the number of channels results in a more significant increase in the number of parameters, FLOPs, and memory usage. Inference time, however, increases proportionally with the number of channels. For instance, with 36 channels, the average inference time is 57.16$\mathit{ms}$, and when doubled, it requires approximately 108.74$\mathit{ms}$, nearly twice the time. As the number of channels increases from 24 to 84 in increments of 12, the PSNR on the Urban100~\cite{Urban100} dataset increases by 0.42 db, 0.29 db, 0.19 db, 0.13 db, and 0.10 db, respectively. Based on the overall performance increase, we designate the model with 60 channels as the Base model and the model with 84 channels as the Large model. In this context, the Small model has an inference time about 3ms longer than the Base model, but it has fewer parameters, lower memory usage, and fewer FLOPs, thus justifying its designation.

\textbf{Impact of Number of Heads.}
\label{App:head}
\begin{table}[t]
	\caption{Performance difference of LMLT based on the number of heads. Chan is the number of channels in each heads. Best results are highlighted.}
	\label{tab:perhead}
	\renewcommand\arraystretch{1.1}
	\begin{center}
		\resizebox{\textwidth}{!}{
			\begin{tabular}{| c | c | c | c | c | c | c | c | c | c | c| c| c|}
				\hline
				Scale & \#Heads & \#Chan & \#Params & \#FLOPs & \#Acts & \#GPU & Set5 & Set14 & B100 & Urban100 & Manga109 \\
				\hline
    
				\multirow{4}*{$\times 2$} 
                & 1 &36 & 270K & 75G & 845M & 437.21M & 38.00/0.9606 & 33.58/0.9179 & 32.17/0.8997 & 31.93/0.9260 & 38.83 /0.9774 \\ 
                ~ & 2 & 18 & 250K & 64G & 717M  & 385.96M & \textbf{38.01}/0.9606 &\textbf{33.59}/0.9180 &32.18/\textbf{0.8999} &32.02/0.9270 &38.87/\textbf{0.9776} \\ 
                ~ & 3 & 12 & 243K & 60G & 646M  & 346.71M & 38.00/0.9606 &\textbf{33.59}/0.9182 &\textbf{32.19/0.8999} &32.02/\textbf{0.9273} &38.88/0.9775 \\ 
                ~ & 4 & 9 & 239K & 59G & 603M & 324.01M & \textbf{38.01}/0.9606 & \textbf{33.59}/\textbf{0.9183} & \textbf{32.19}/\textbf{0.8999} & \textbf{32.04/0.9273} & \textbf{38.90}/0.9775 \\ 
				\hline

    		  \multirow{4}*{$\times 3$} 
                & 1 &36 & 275K & 35G & 396M & 205.52M & 34.37/0.9271 & \textbf{30.39/0.8431} & 29.11/0.8054 & 28.05/0.8495 & 33.71/0.9449 \\ 
                ~ & 2 & 18 & 255K & 30G & 336M  & 181.14M & 34.37/0.9271 &30.37/0.8427 &29.11/0.8056 &28.05/0.8496 &\textbf{33.73/0.9450} \\ 
                ~ & 3 & 12 & 248K & 29G & 303M  & 161.90M & \textbf{34.41/0.9273} &30.37/0.8426 &\textbf{29.12/0.8059} &28.09/0.8502 &\textbf{33.73}/0.9449 \\ 
                ~ & 4 & 9 & 244K & 28G & 283M & 151.96M & 34.36/0.9271 & 30.37/0.8427 & \textbf{29.12}/0.8057 & \textbf{28.10/0.8503} & 33.72/0.9448 \\ 
				\hline
    
				\multirow{4}*{$\times 4$}
                & 1 &36 & 282K & 19G & 213M & 111.28M & 32.14/0.8943& \textbf{28.65}/\textbf{0.7826}& 27.60/0.7366& 26.04/0.7825 & 30.57/0.9080 \\ 
                ~ & 2 & 18 & 261K & 17G & 181M  & 98.69M & 32.18/\textbf{0.8948} &28.63/\textbf{0.7826} &27.60/\textbf{0.7370} &26.07/0.7839 & 30.59/0.9085 \\ 
                ~ & 3 & 12 & 254K & 16G & 163M  & 87.04M & \textbf{32.19}/0.8947 &28.63/0.7821 &27.60/0.7367 &\textbf{26.08}/0.7834 &30.58/0.9080 \\ 
                ~ & 4 & 9 & 251K & 15G & 152M & 81.44M & \textbf{32.19}/0.8947 &  28.64/0.7823 & 27.60/0.7369 & \textbf{26.08/0.7838} & \textbf{30.60/0.9083}\\
				\hline

		\end{tabular}}
	\end{center}
\end{table}
In this paragraph, we compare the performance differences based on the number of heads. In our model, as the number of heads decreases, the channel and the number of downsizing operations for each head decrease. For example, in our baseline with 4 heads and 36 channels, the lowest head has a total of 9 channels and is pooled 3 times. However, if there are 2 heads, the lowest head has 18 channels and is pooled once. Additionally, the maximum pooling times and the number of heads are related to the number of windows and the amount of self-attention computation. According to equation~\ref{eq2}, as the number of heads decreases, the computation increases. As a result, as the number of heads decreases, the number of parameters, FLOPs, and GPU memory usage increase.

As shown in Table~\ref{tab:perhead}, the performance with 4 heads and 3 heads is similar across all scales and test datasets. However, when the number of heads is reduced to 1, the performance drops significantly. This difference is particularly noticeable in the Urban100~\cite{Urban100} dataset, where at scale $\times2$, the performance with 4 heads is 32.04 db, whereas with 1 head, it drops to 31.93 db, a decrease of 0.11 db. Additionally, when the scale is $\times3$ and $\times4$, the PSNR decreases by 0.05 dB and 0.04 dB, respectively. This indicates that even if the spatial size of all features is maintained with a single head, even though the number of parameters, FLOPs, and channels per head increase, the inability to capture information from other windows can lead to a decline in performance.

\textbf{Impact of Number of Depths.}
\label{App:depth}
\begin{table}[t]
	\caption{Performance difference of LMLT based on the number of depths. Best results are highlighted.}
	\label{tab:perdepth}
	\renewcommand\arraystretch{1.1}
	\begin{center}
		\resizebox{\textwidth}{!}{
			\begin{tabular}{| c | c | c | c | c | c | c | c | c | c | c |}
				\hline
				Scale & \#Depths & \#Params & \#FLOPs & \#Acts & \#AVG Time & Set5 & Set14 & B100 & Urban100 & Manga109 \\
				\hline
    
				\multirow{3}*{$\times 2$} 
                & 1 & 239K & 59G & 603M & 57.37$\mathit{ms}$ & 38.01/0.9606 & 33.59/0.9183  & 32.19/0.8999 & 32.04/0.9273 & \textbf{38.90}/0.9775 \\ 
                ~ & 2 & 254K & 63G & 911M & 70.93$\mathit{ms}$  & 38.01/0.9606 &\textbf{33.61/0.9185} &\textbf{32.20/0.9000} &32.06/0.9274 &38.89/\textbf{0.9776} \\ 
                ~ & 3 & 268K & 67G & 1219M & 84.50$\mathit{ms}$ & 38.01/\textbf{0.9607} &33.59/0.9180 &32.19/\textbf{0.9000} &\textbf{32.08/0.9276} &38.89/\textbf{0.9776} \\ 
				\hline

    		  \multirow{3}*{$\times 3$} 
                & 1 & 244K & 28G & 283M & 31.06$\mathit{ms}$  & 34.36/0.9271 & 30.37/0.8427 & \textbf{29.12/0.8057} & 28.10/0.8503 & 33.72/0.9448 \\ 
                ~ & 2  & 259K & 30G & 427M  & 38.20$\mathit{ms}$ & 34.39/\textbf{0.9275} &30.38/\textbf{0.8429} &29.11/0.8056 &\textbf{28.11/0.8507} &\textbf{33.75/0.9451} \\ 
                ~ & 3  & 273K & 32G & 570M & 48.75$\mathit{ms}$ & \textbf{34.40}/0.9274 &\textbf{30.39}/0.8428 &29.11/0.8056 &28.08/0.8501 &33.73/0.9449 \\ 

				\hline
    
				\multirow{3}*{$\times 4$}
                & 1  & 251K & 15G & 152M & 23.54$\mathit{ms}$ & 32.19/0.8947& 28.64/0.7823& 27.60/0.7369& 26.08/0.7838 & 30.60/0.9083 \\ 
                ~ & 2  & 265K & 16G & 230M  & 33.27$\mathit{ms}$ & 32.20/0.8949 &28.65/0.7823 &27.60/0.7369 &26.08/0.7839 & 30.58/0.9083 \\ 
                ~ & 3 & 279K & 17G & 307M & 44.07$\mathit{ms}$ & \textbf{32.23/0.8954} &\textbf{28.66/0.7825} &\textbf{27.61}/0.7369 &\textbf{26.10/0.7845} &30.59/\textbf{0.9084} \\ 

				\hline

		\end{tabular}}
	\end{center}
\end{table}
Additionally, we examine how the performance changes when we add more attention modules to our model. The proposed LMLT connects self-attention layers in parallel, where self-attention is calculated in lower heads and then connected to the upper layers. However, a different approach, like other models~\cite{SwinIR, srformer}, could be to calculate self-attention multiple times(i.e., in series) before sending it to the upper heads, thus mixing serial and parallel connections.

However, our experimental results indicate that calculating self-attention multiple times in a single head before sending it to the upper heads is not an effective choice. As shown in Table~\ref{tab:perdepth}, increasing the self-attention calculations from once to three times increases the inference time by 87.7\% on the Urban100~\cite{Urban100} dataset at $\times 4$ scale, but the PSNR only improves by 0.02 dB. Similar trends are observed in other datasets and scales, showing minimal differences in PSNR and SSIM performance. This demonstrates that having one head composed of serial self-attention layers and connecting heads in parallel does not yield good performance relative to inference time.

\section{Effects of Low-to-high connection and Pooling}
\begin{table}[t]
	\caption{The comparison table between SAFMN and its variant with low-to-high element-wise sum added. For each model, the better results are highlighted in \textbf{bold}.}
	\label{tab:safmn_sum}
	\renewcommand\arraystretch{1.1}
	\begin{center}
		\resizebox{\textwidth}{!}{
			\begin{tabular}{| c | c | c | c |  c | c | c |}
				\hline
				Scale & Method & Set5 & Set14 & B100 & Urban100 & Manga109 \\
				\hline
    
				\multirow{2}*{$\times 2$} 
                & SAFMN~\cite{SAFMN}       &\textbf{38.00}/\textbf{0.9605}&\textbf{33.54}/\textbf{0.9177}&32.16/0.8995&31.84/0.9256&38.71/0.9771\\
				~ & SAFMN~\cite{SAFMN} $\mathit{w/}$ sum    &37.99/0.9604&33.52/0.9174&\textbf{32.17}/\textbf{0.8996} &\textbf{31.86}/\textbf{0.9257} &\textbf{38.77}/\textbf{0.9773} \\
				\hline

    		  \multirow{2}*{$\times 3$} 
                & SAFMN~\cite{SAFMN}       &34.34/0.9267 &\textbf{30.33}/0.8418 &29.08/0.8048 &27.95/0.8474 &33.52/0.9437 \\
				~ & SAFMN~\cite{SAFMN} $\mathit{w/}$ sum  &34.34/\textbf{0.9269} &30.32/0.8418 & \textbf{29.09}/\textbf{0.8049} & \textbf{27.96}/\textbf{0.8476}&\textbf{33.55}/\textbf{0.9438} \\
    		\hline
    
				\multirow{2}*{$\times 4$}
                & SAFMN~\cite{SAFMN}       & \textbf{32.18}/\textbf{0.8948} & \textbf{28.60}/\textbf{0.7813} & 27.58/\textbf{0.7359} & \textbf{25.97}/\textbf{0.7809} & 30.43/0.9063 \\
				~ & SAFMN~\cite{SAFMN} $\mathit{w/}$ sum   &32.10/0.8937&28.59/0.7812&27.58/0.7358&25.96/0.7799&\textbf{30.44}/0.9063 \\
				\hline
		\end{tabular}}
	\end{center}
\end{table}

\textbf{Difference between with and without Low-to-high connection}
\label{App:low-to-high}
\begin{figure*}[!t]\footnotesize
 \begin{center}
  \begin{tabular}{ccc}
      \includegraphics[width=0.30\linewidth, height = 0.30\linewidth]{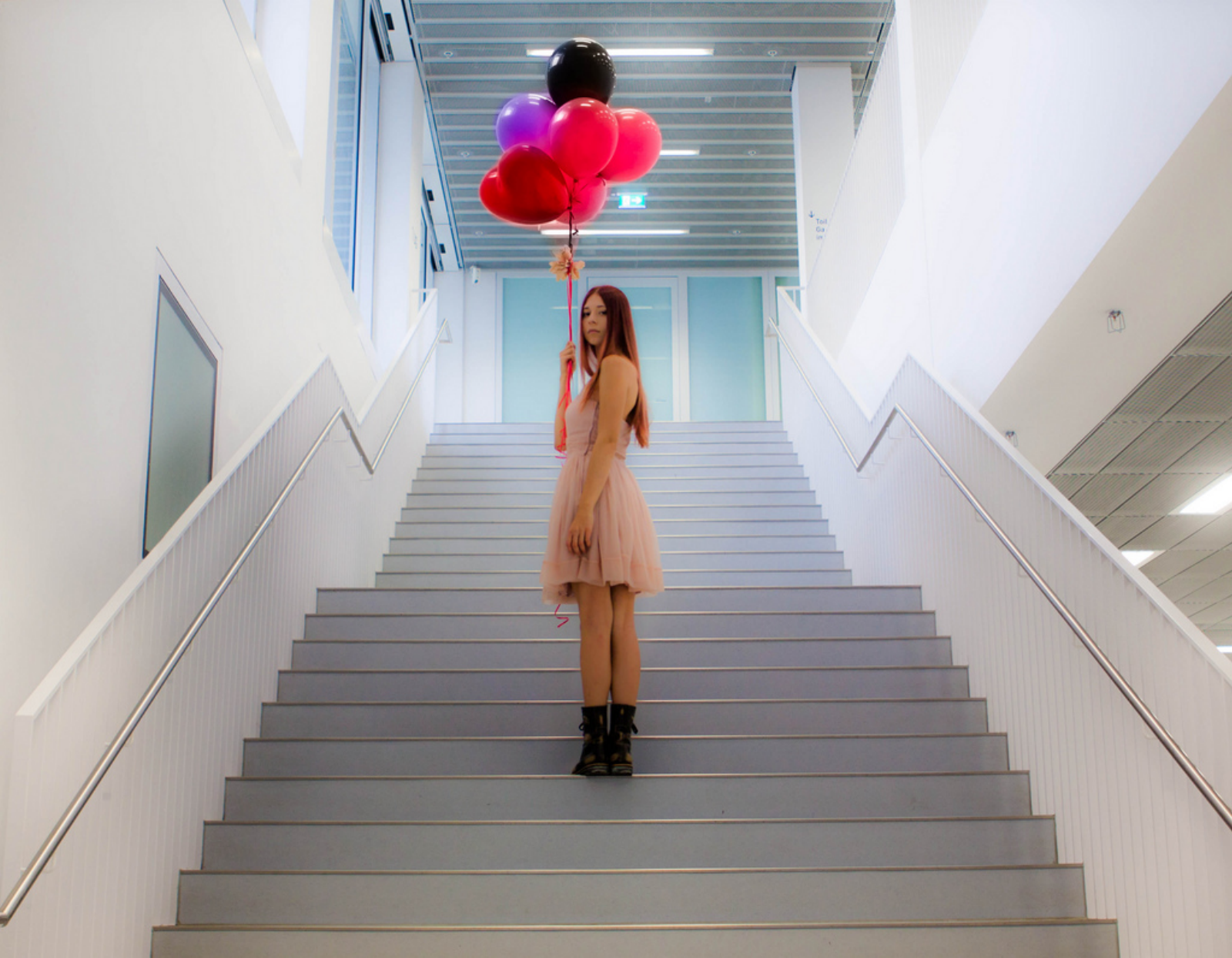} &\hspace{-2mm}
      \includegraphics[width=0.30\linewidth, height = 0.30\linewidth]{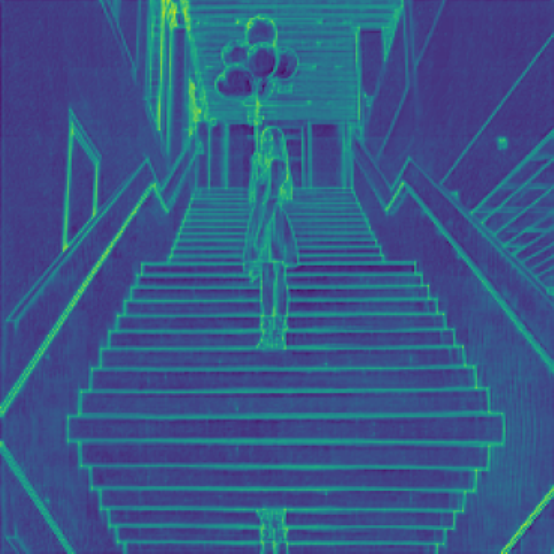} &\hspace{-2mm}
      \includegraphics[width=0.30\linewidth, height = 0.30\linewidth]{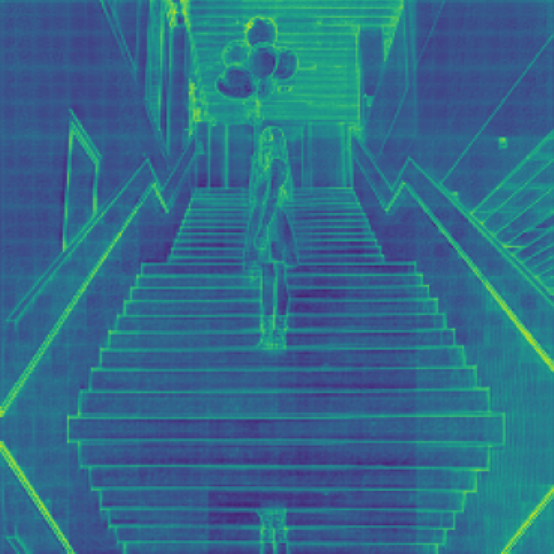} \\
      \hspace{-3.5mm}img009($\times4$) from Urban100  &\hspace{-3.5mm} &\hspace{-3.5mm}  \\

      \includegraphics[width=0.30\linewidth, height = 0.30\linewidth]{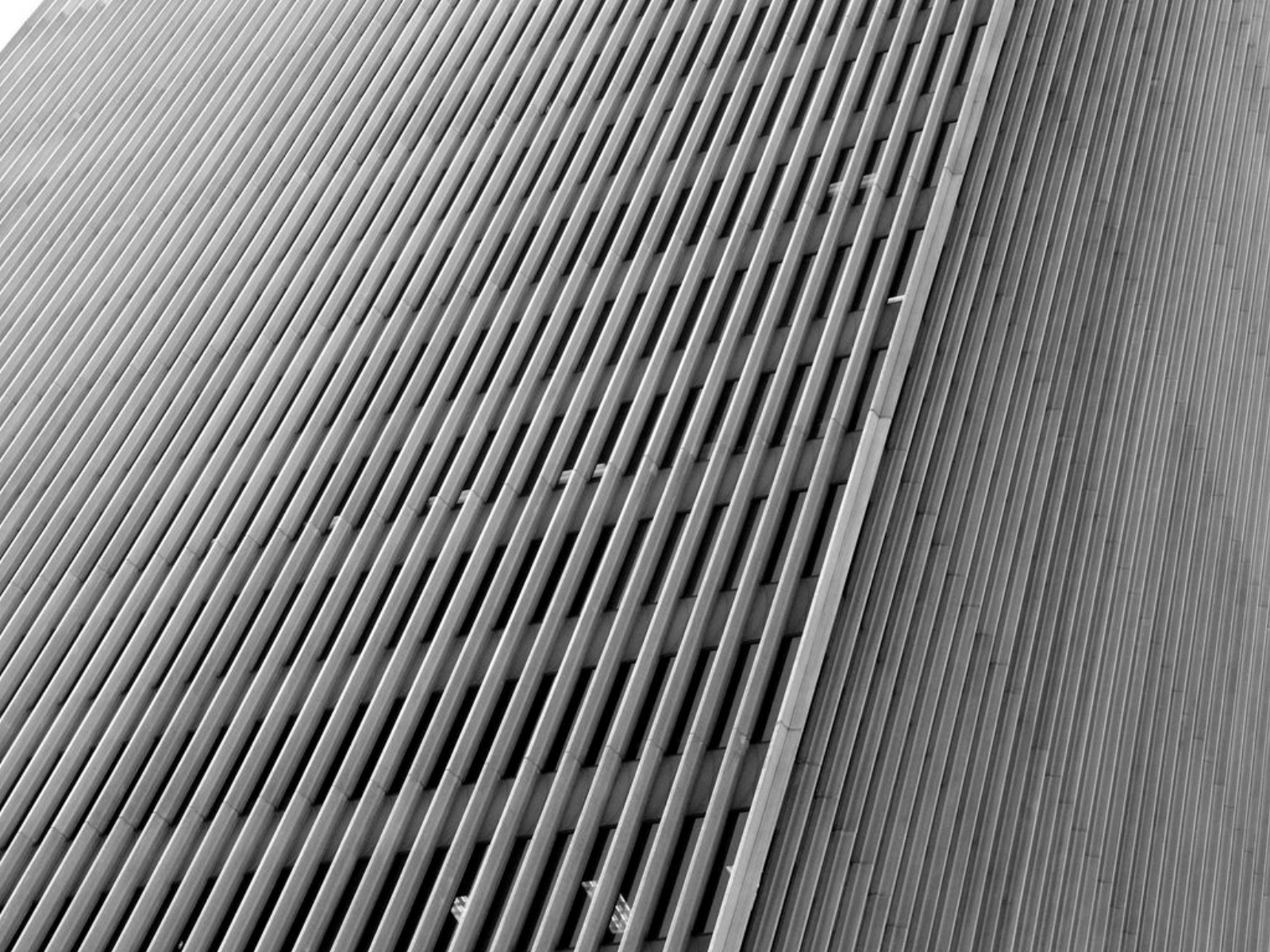} &\hspace{-2mm}
      \includegraphics[width=0.30\linewidth, height = 0.30\linewidth]{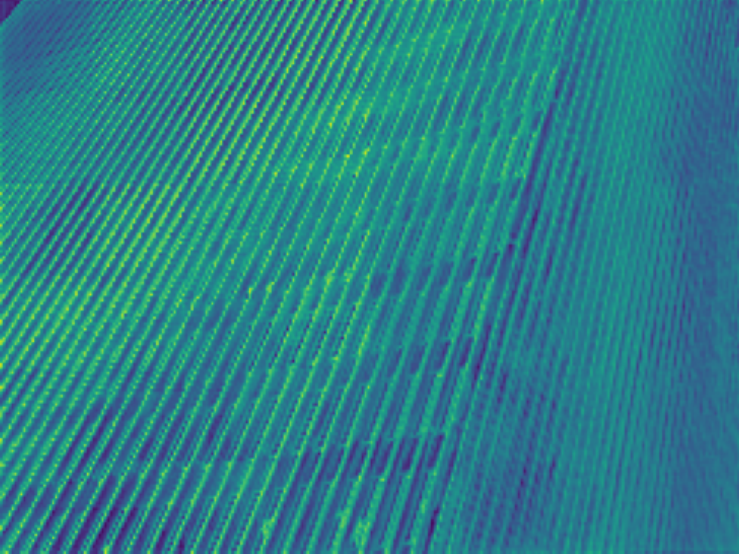} &\hspace{-2mm}
      \includegraphics[width=0.30\linewidth, height = 0.30\linewidth]{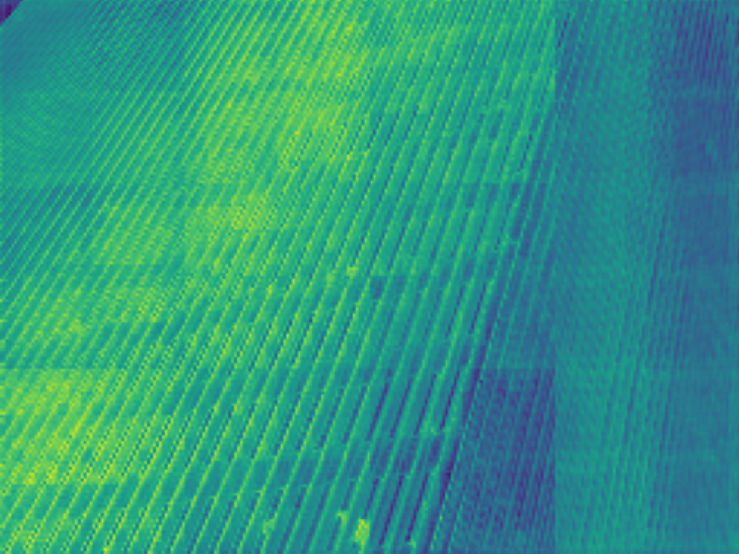} \\
      \hspace{-3.5mm}img045($\times4$) from Urban100  &\hspace{-3.5mm} &\hspace{-3.5mm}  \\

    \includegraphics[width=0.30\linewidth, height = 0.30\linewidth]{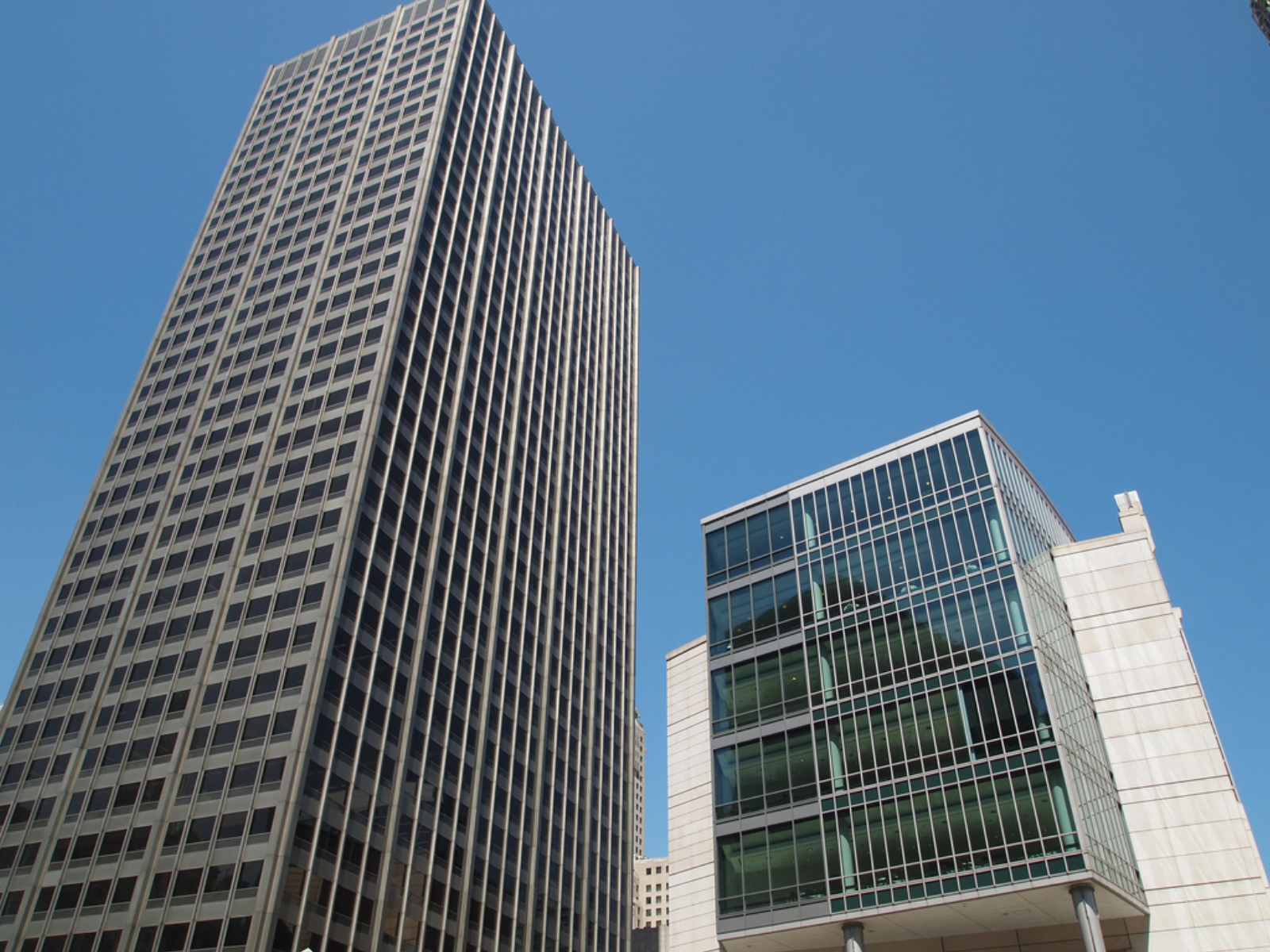} &\hspace{-2mm}
      \includegraphics[width=0.30\linewidth, height = 0.30\linewidth]{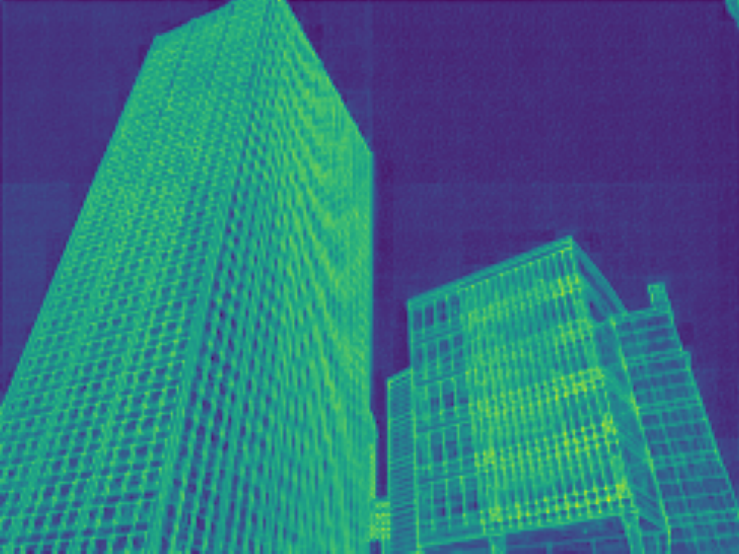} &\hspace{-2mm}
      \includegraphics[width=0.30\linewidth, height = 0.30\linewidth]{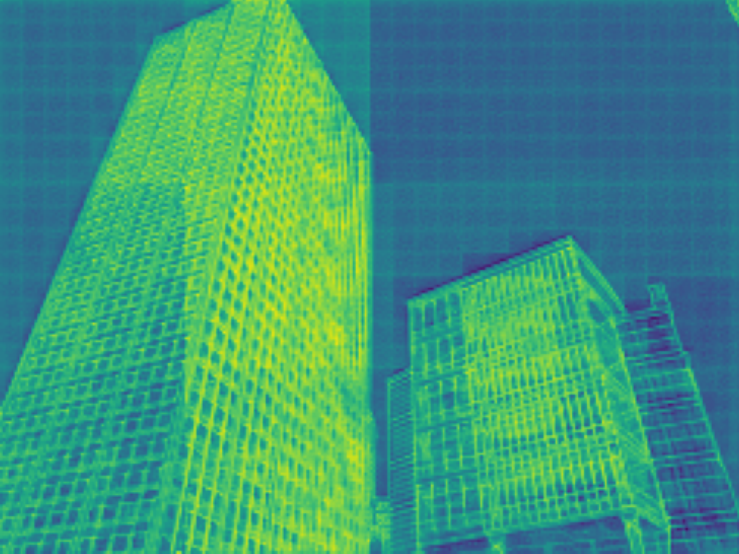} \\
      \hspace{-3.5mm}img096($\times4$) from Urban100  &\hspace{-3.5mm}(a) &\hspace{-3.5mm}(b)  \\

\\
 \end{tabular}
 \end{center}
 \caption{Visualization of features with low-to-high connection(a) and without connection(b) on Urban100$\times4$. As shown in the images, without the low-to-high connection, the boundaries between windows are clearly visible.}
 \label{fig:nosum}
\end{figure*}
In the Table~\ref{tab:nosum}, we confirm performance differences when the low-to-high connection is not applied to LMLT. Inspired by this, we also apply low-to-high connections between heads in SAFMN~\cite{SAFMN} and verify the experimental results. Table~\ref{tab:safmn_sum} shows that adding low-to-high connection to the upper head in SAFMN~\cite{SAFMN} does not yield significant performance differences. Moreover, at the $\times4$ scale, the SSIM for the Urban100~\cite{Urban100} and Set5~\cite{Set5} datasets decreases by 0.0010 and 0.0011, respectively, indicating a reduction in performance.

We then visualize the features of LMLT-Tiny to understand the effect of the low-to-high connection. Each column of Figure~\ref{fig:nosum} illustrates the original image, the aggregated feature visualization of LMLT-Tiny combining all heads~\ref{fig:nosum}(a), and the aggregated feature visualization of LMLT-Tiny without low-to-high connection~\ref{fig:nosum}(b). In ~\ref{fig:nosum}(b), the images show pronounced boundaries in areas such as stairs, buildings, and the sky. In contrast, ~\ref{fig:nosum}(a) shows these boundaries as less pronounced. This demonstrates that the low-to-high connection can address the border communication issues inherent in WSA.

\textbf{Difference between with and without Pooling}
\label{App:pool}
\begin{figure*}[!t]\footnotesize
 \begin{center}
  \begin{tabular}{cccc}
      \includegraphics[width=0.22\linewidth, height = 0.22\linewidth]{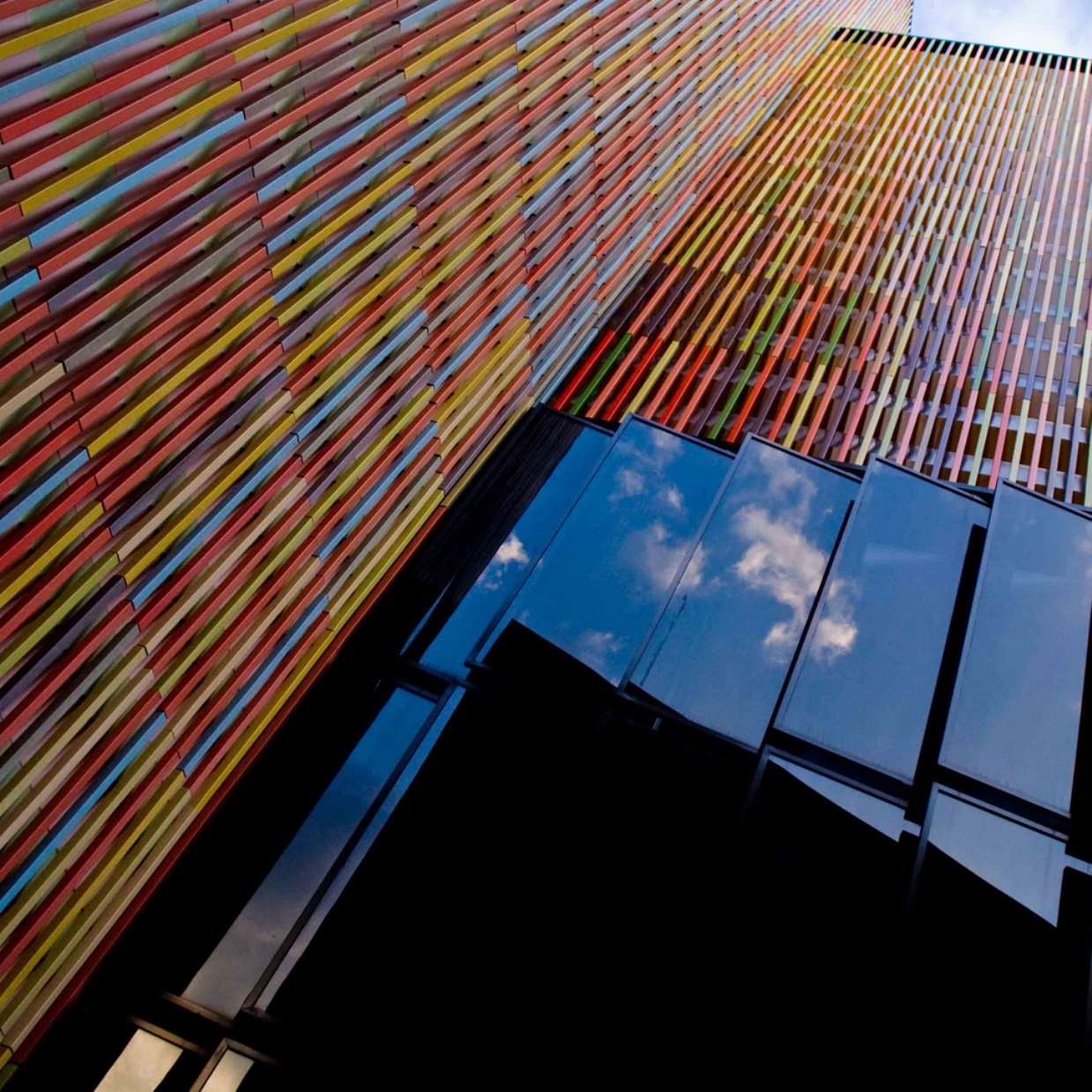} &\hspace{-2mm}
      \includegraphics[width=0.22\linewidth, height = 0.22\linewidth]{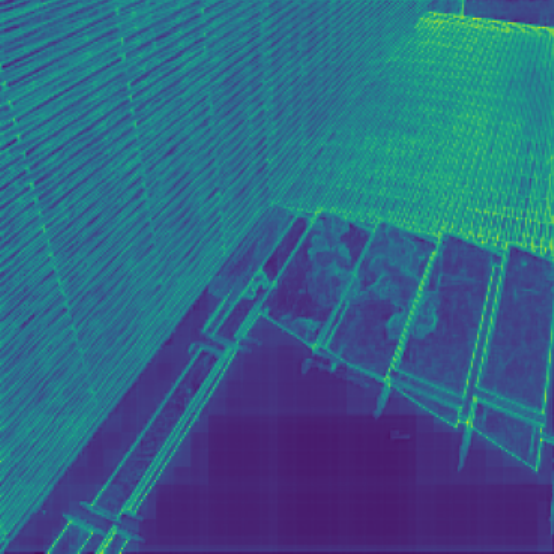} &\hspace{-2mm}
      \includegraphics[width=0.22\linewidth, height = 0.22\linewidth]{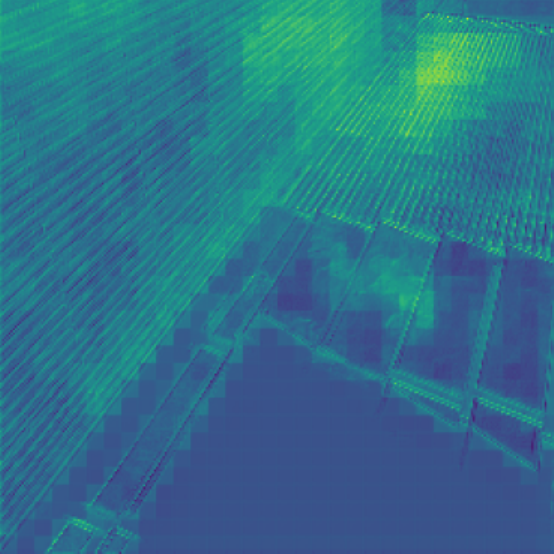} &\hspace{-2mm}
      \includegraphics[width=0.22\linewidth, height = 0.22\linewidth]{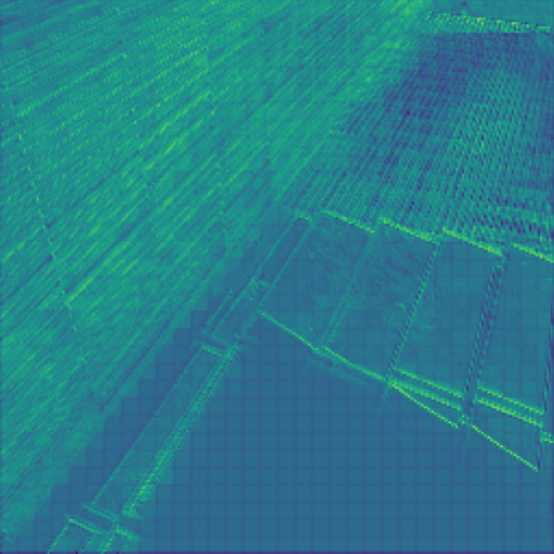} \\
      \hspace{-2mm} img023($\times4$) from Urban100  &\hspace{-2mm} &\hspace{-2mm} &\hspace{-2mm} \\

      \includegraphics[width=0.22\linewidth, height = 0.22\linewidth]{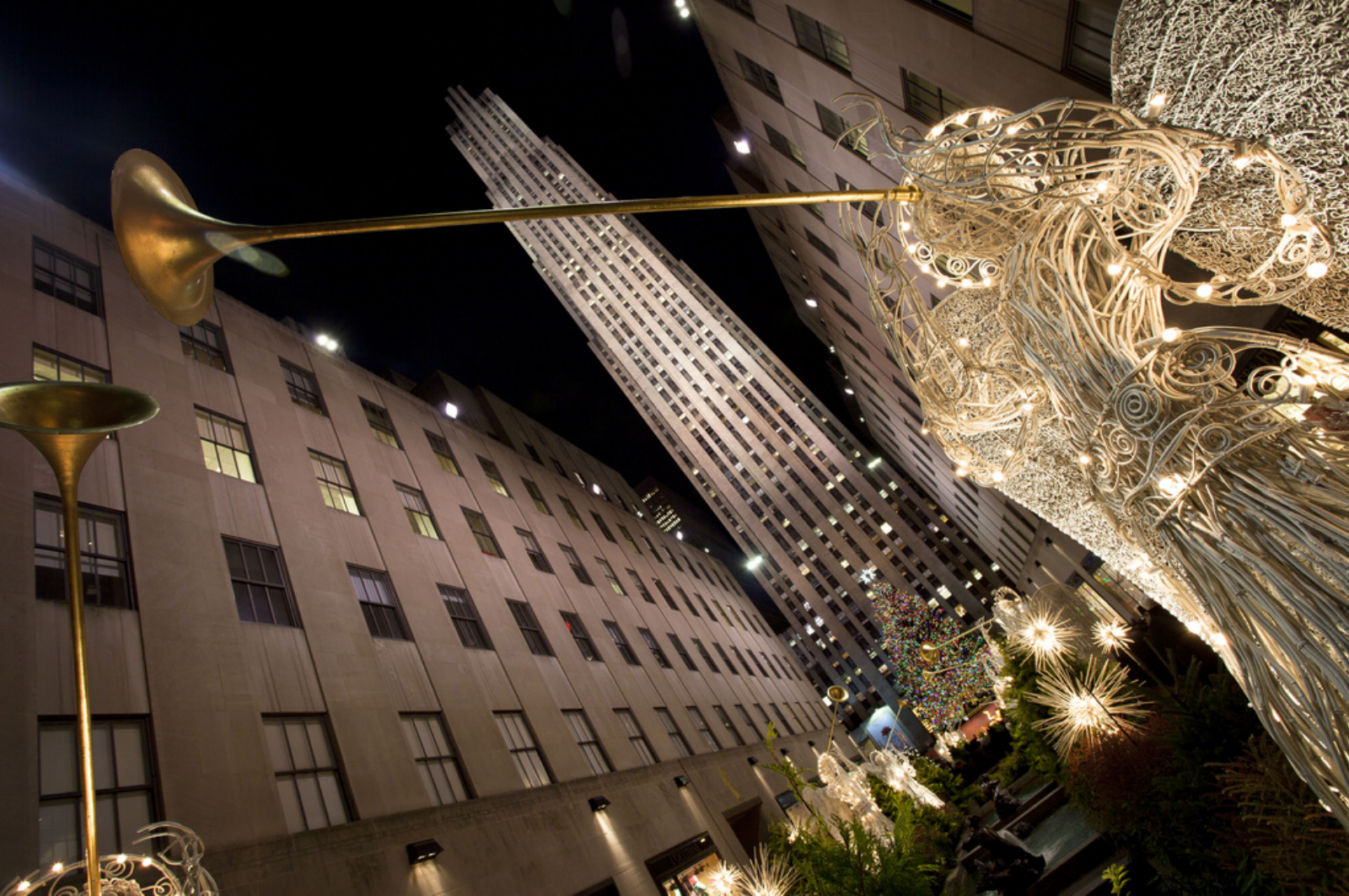} &\hspace{-2mm}
      \includegraphics[width=0.22\linewidth, height = 0.22\linewidth]{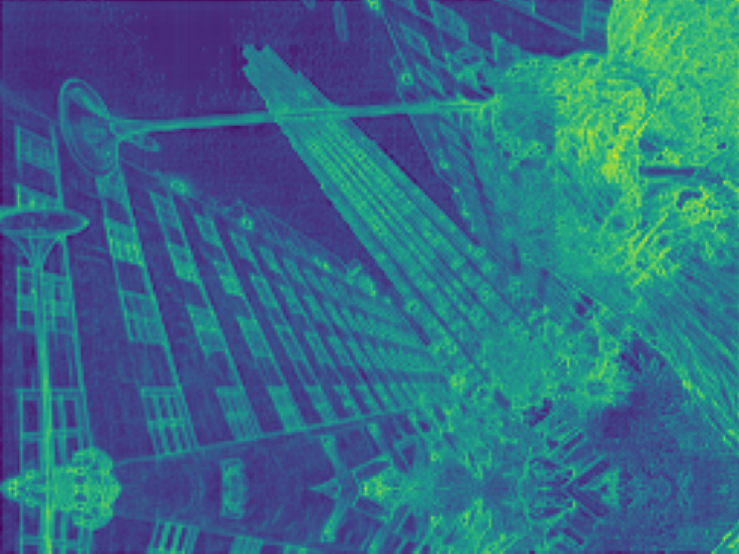} &\hspace{-2mm}
      \includegraphics[width=0.22\linewidth, height = 0.22\linewidth]{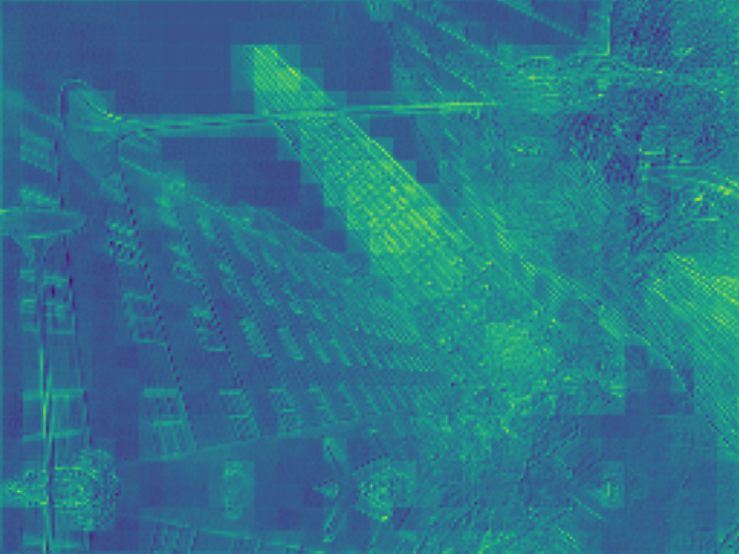}
      &\hspace{-2mm}
      \includegraphics[width=0.22\linewidth, height = 0.22\linewidth]{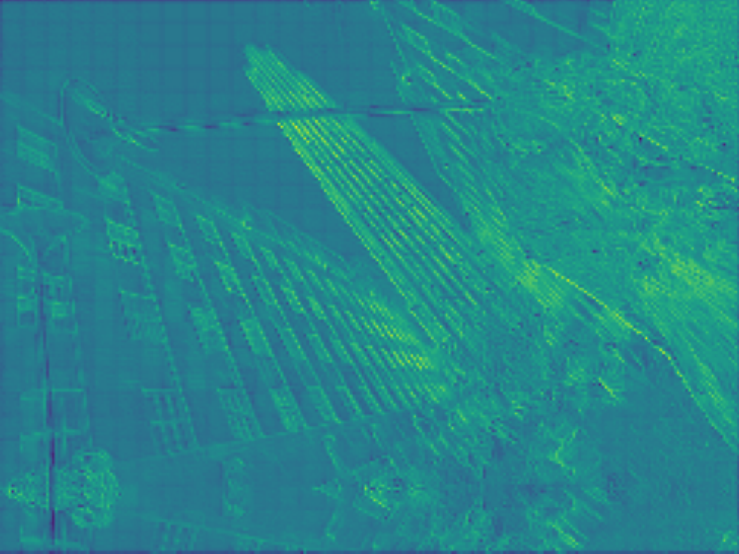} \\
      \hspace{-2mm}img031($\times4$) from Urban100  &\hspace{-2mm} &\hspace{-2mm} &\hspace{-2mm}  \\

    \includegraphics[width=0.22\linewidth, height = 0.22\linewidth]{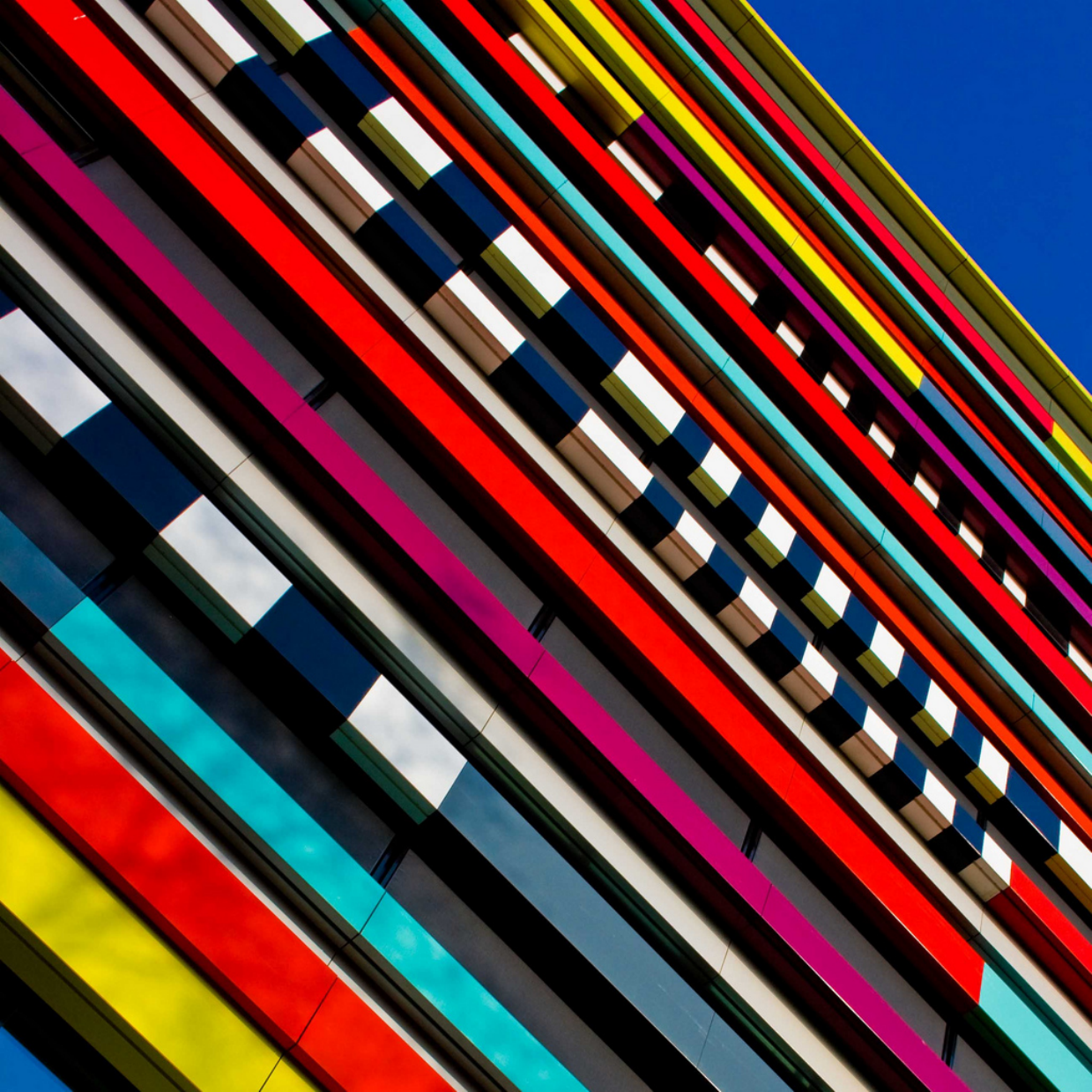} &\hspace{-2mm}
      \includegraphics[width=0.22\linewidth, height = 0.22\linewidth]{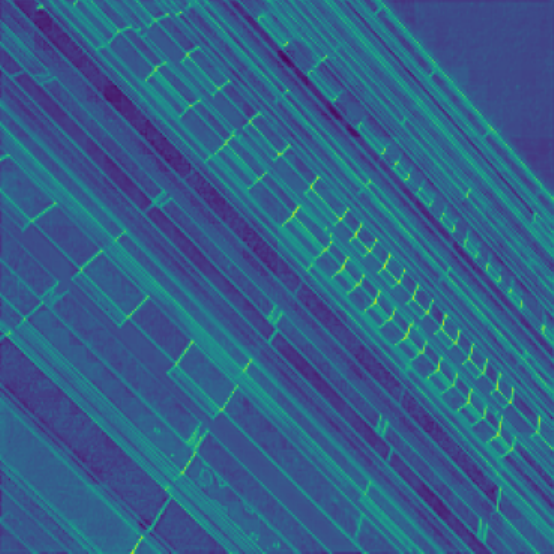} &\hspace{-2mm}
      \includegraphics[width=0.22\linewidth, height = 0.22\linewidth]{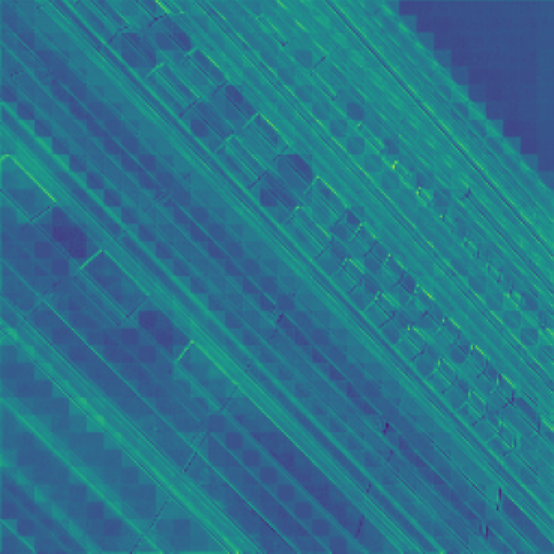}
      &\hspace{-2mm}
      \includegraphics[width=0.22\linewidth, height = 0.22\linewidth]{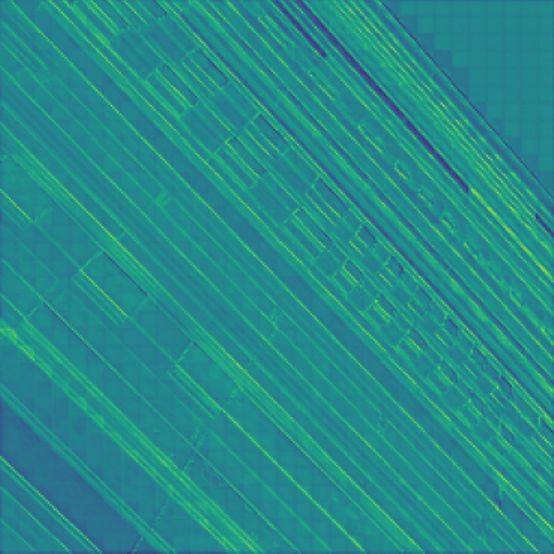} \\
      \hspace{-2mm}img081($\times4$) from Urban100  &\hspace{-2mm}(a) &\hspace{-2mm}(b) &\hspace{-2mm}(c)  \\

\\
 \end{tabular}
 \end{center}
 \caption{Comparison without pooling on Urban100 $\times4$ scale. From left to right: (a) LMLT with pooling applied, (b) without any pooling, (c) without pooling and without multiplication by activation. In (b) and (c), the boundaries between windows are visible.}
 \label{fig:nopool}
\end{figure*}
We analyze the impact of pooling on the performance of super-resolution. As observed in Table~\ref{tab:nopool}, even though pooling preserves spatial information, the overall performance decreases when it is not applied. We investigate the reason behind this through feature visualization. Figure~\ref{fig:nopool} visualizes the features when no pooling is applied to any head in LMLT. The leftmost image is the original Urban100~\cite{Urban100} image. Figure~\ref{fig:nopool}(a) shows the aggregated features of all heads in LMLT-Tiny. Column Figure~\ref{fig:nopool}(b) visualizes the features without pooling, and Figure~\ref{fig:nopool}(c) visualizes the features without both pooling and merging, all at the $\times4$ scale. In ~\ref{fig:nopool}(b) and ~\ref{fig:nopool}(c), grid patterns are evident across the images, indicating that the disadvantages of being limited to local windows outweigh the benefits of maintaining the original spatial size.

\section{Impact of Activation function}
\begin{table}[t]
	\caption{Performance difference of LMLT with GELU and without GELU. The better results are highlighted in \textbf{bold}.}
	\label{tab:gelu}
	\renewcommand\arraystretch{1.1}
	\begin{center}
		\resizebox{\textwidth}{!}{
			\begin{tabular}{| c | c | c | c | c | c | c | c |}
				\hline
				Scale & Ablation & \#Channel & Set5 & Set14 & B100 & Urban100 & Manga109 \\
				\hline
    
				\multirow{4}*{$\times 2$} 
                & LMLT & 36  & 38.01/0.9606 & 33.59/0.9183 & 32.19/0.8999 & 32.04/\textbf{0.9273} & 38.90 /0.9775 \\ 
                ~ & LMLT $\mathit{w/o}$ GELU & 36  & \textbf{38.03}/0.9606 &\textbf{33.60}/\textbf{0.9184} &32.19/\textbf{0.9000} &\textbf{32.05}/0.9272 &\textbf{38.91}/\textbf{0.9776} \\ 
                \cline{2-8}
                
               ~ & LMLT & 60  & 38.10/0.9610 &33.76/\textbf{0.9201} &32.28/\textbf{0.9012} &\textbf{32.52}/\textbf{0.9316} &39.24/0.9783 \\ 
               ~ & LMLT $\mathit{w/o}$ GELU & 60  & 38.10/0.9610 &\textbf{33.80}/0.9200 &32.28/0.9011 &32.51/0.9315 &\textbf{39.26}/0.9783 \\ 
				\hline
    \hline

    		  \multirow{4}*{$\times 3$} 
                & LMLT & 36  & 34.36/0.9271 & \textbf{30.37/0.8427} & \textbf{29.12}/0.8057 & \textbf{28.10/0.8503} & \textbf{33.72/0.9448} \\ 
                ~ & LMLT $\mathit{w/o}$ GELU & 36 & \textbf{34.37/0.9272} &30.36/0.8425 &29.11/0.8057 &28.08/0.8502 &33.71/0.9447 \\ 
                \cline{2-8}
                
                ~ & LMLT & 60   & \textbf{34.58/0.9285} &\textbf{30.53/0.8458} &\textbf{29.21/0.8084} &\textbf{28.48/0.8581} &\textbf{34.18/0.9477} \\ 
                ~ & LMLT $\mathit{w/o}$ GELU & 60  & 34.53/0.9283 &30.51/0.8457 &29.20/0.8080 &28.45/0.8576 &34.17/0.9476 \\

				\hline
                \hline
    
				\multirow{4}*{$\times 4$}
                &LMLT & 36   & 32.19/0.8947& \textbf{28.64/0.7823}& 27.60/0.7369& 26.08/\textbf{0.7838} & \textbf{30.60/0.9083} \\ 
                ~ & LMLT $\mathit{w/o}$ GELU & 36 & \textbf{32.23/0.8949} &28.62/0.7820 &27.60/0.7369 &26.08/0.7836 & 30.59/0.9082 \\ 
                \cline{2-8}
                
                ~ & LMLT & 60   & 32.38/0.8971 &\textbf{28.79/0.7859} &\textbf{27.70/0.7403} &\textbf{26.44/0.7947} &\textbf{31.09/0.9139} \\ 
                ~ & LMLT $\mathit{w/o}$ GELU & 60 & \textbf{32.39/0.8973} &28.78/0.7858 &27.69/0.7399 &26.39/0.7934 &31.04/0.9132 \\ 

				\hline

		\end{tabular}}
	\end{center}
\end{table}

\label{App:each_module}
In Table~\ref{tab:each_module}, we discuss that not applying the activation function GeLU~\cite{GELU} might improve performance. Therefore, we experiment with LMLT and LMLT without GeLU~\cite{GELU} across various scales and channels to confirm the results.

Table~\ref{tab:gelu} shows the results for our LMLT and the model without the activation function across different scales and channels. As shown, with 36 channels, there is minimal performance difference across all scales, with the largest being a 0.04 higher PSNR on the Set5~\cite{Set5} $\times4$ scale when GeLU~\cite{GELU} is removed. However, when expanded to 60 channels, our LMLT performs better on most benchmark datasets for both $\times3$ and $\times4$ scales. Specifically, on the $\times4$ scale of the Urban100~\cite{Urban100} dataset, PSNR and SSIM are higher by 0.05 dB and 0.0013, respectively. This demonstrates that adding GeLU~\cite{GELU} after aggregating features is more beneficial for performance improvement.

\section{LAM and ERF Comparisons}
\label{APP:LAM}
\begin{figure*}[!t]\footnotesize
 \begin{center}
  \begin{tabular}{c}
        \includegraphics[width=\textwidth]{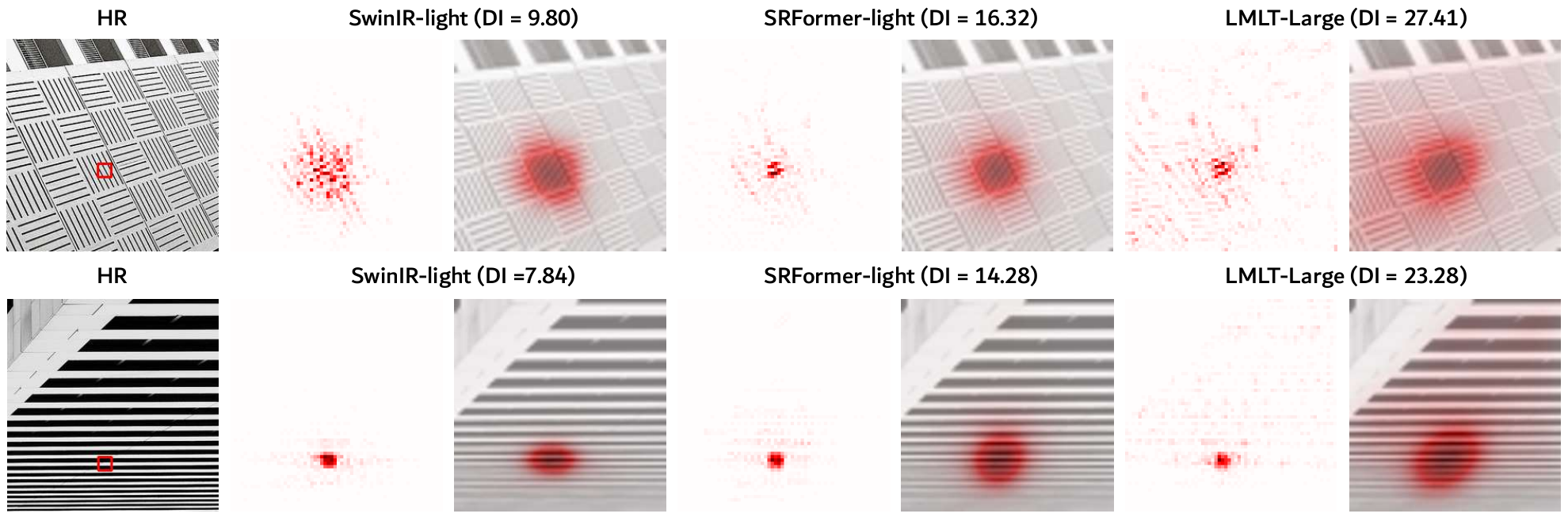}
\\
 \end{tabular}
 \end{center}
 \caption{The LAM results of SwinIR-Light~\cite{SwinIR}, SRFormer-Light~\cite{srformer}, and our proposed model(LMLT-Large). As shown in the figure, our proposed model references a broader range of pixels when reconstructing the image.}
 \label{fig:LAM}
\end{figure*}
\begin{figure*}[!t]\footnotesize
 \begin{center}
  \begin{tabular}{c}
        \includegraphics[width=\textwidth]{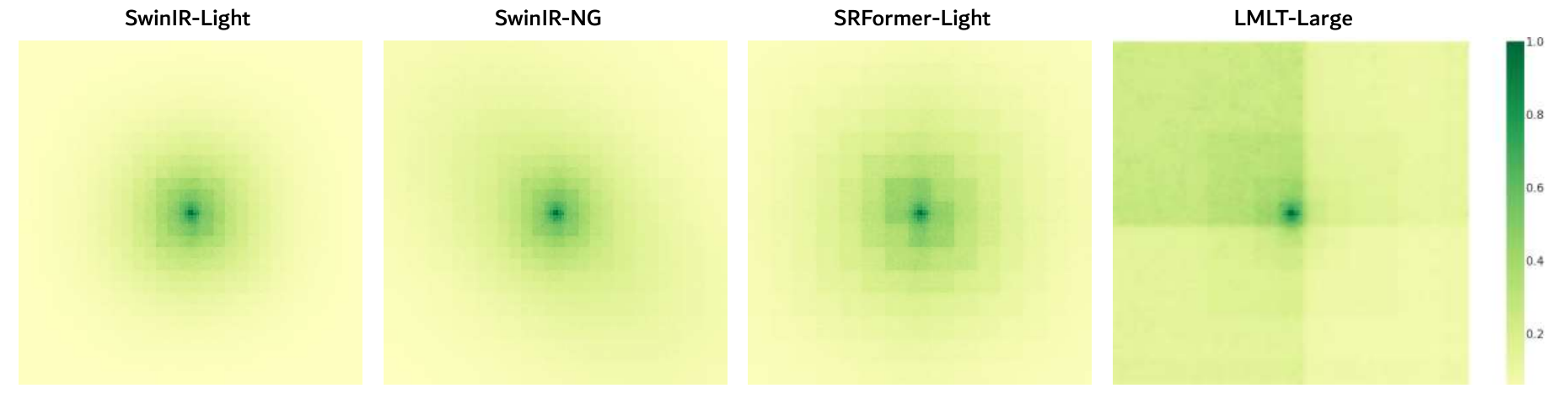}
\\
 \end{tabular}
 \end{center}
 \caption{The ERF visualizations of SwinIR-Light~\cite{SwinIR}, SwinIR-NG~\cite{ngswin}, SRFormer-Light~\cite{srformer}, and the proposed model (LMLT-Large). The darker areas are more widely distributed, indicating a larger ERF, and the figure shows that the proposed model effectively utilizes global information.}
 \label{fig:erf}
\end{figure*}

To verify whether the proposed LMLT exhibits a wider receptive field, we utilize local attribution map (LAM)~\cite{lam} and effective receptive field (ERF)~\cite{erf}. Specifically, we use LAM to show that our proposed LMLT-Large has a wider receptive field compared to SwinIR-Light~\cite{SwinIR} and SRFormer-Light~\cite{srformer}. Detailed visualizations are presented in Figure~\ref{fig:LAM}. Additionally, SwinIR-NG~\cite{ngswin} is included for comparison, and we visualize the ERF. The detailed results are shown in Figure~\ref{fig:erf}. Through these two analyses, we demonstrate that our proposed model exhibits a wider receptive field than existing ViT-based SR models.

\begin{figure*}[!t]\footnotesize
 \begin{center}
  \begin{tabular}{c}
        \includegraphics[width=\textwidth]{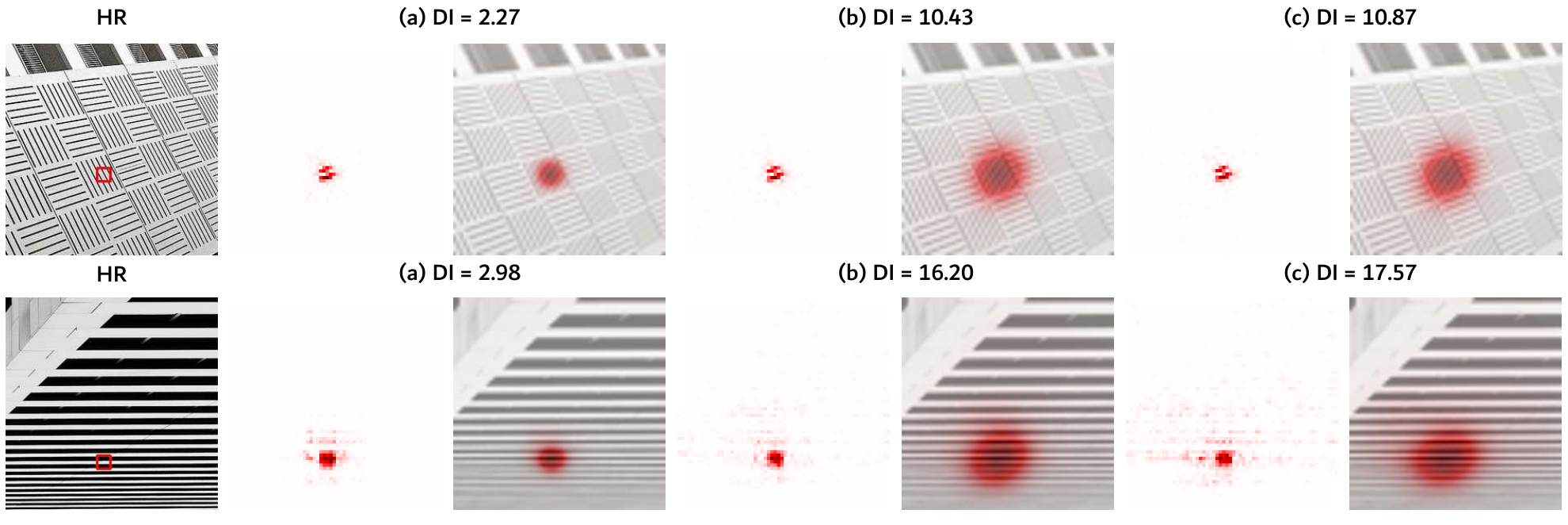}
\\
 \end{tabular}
 \end{center}
 \caption{Visualization of LAM-Tiny. From left to right: (a) LMLT-Tiny without pooling, (b) LMLT-Tiny without low-to-high connection and (c) LMLT-Tiny.}
 \label{fig:LAM_Tiny}
\end{figure*}
Additionally, we compare the LAM of our proposed model, LMLT-Tiny (Figure~\ref{fig:LAM_Tiny}(c)), with a version of the model that does not include pooling for each head (Figure~\ref{fig:LAM_Tiny}(a)), and a version where the low-to-high connection is removed (Figure~\ref{fig:LAM_Tiny}(b)), demonstrating that our proposed model effectively references a broader region. The results show that, even when the spatial size of the model is maintained without pooling, it fails to process information from a wider area. Moreover, the low-to-high connection proves to be effective in enabling the model to capture information from a larger region.

\section{CCM : Convolutional Channel Mixer}
\begin{wrapfigure}{R}{0.5\textwidth}
	\centering
	\includegraphics[width=0.5\textwidth]{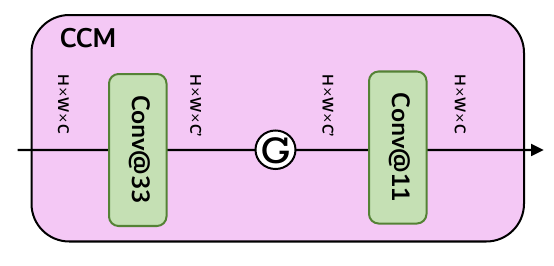}
	\caption{\textbf{CCM(Convolutional Channel Mixer)} proposed in SAFMN~\cite{SAFMN}.}  
	\label{fig:ccm}
\end{wrapfigure}

\label{App:CCM}
\textbf{CCM instead of MLP.} Since the feed-forward network (FFN) in the original transformer~\cite{Attn} is a fully connected layer, we assume that using it in ViT might disrupt the spatial information of the features. Therefore, we apply the convolutional channel mixer (CCM)~\cite{SAFMN} instead, an FFN based on FMBConv~\cite{effnetv2}, to preserve spatial information. CCM is a module that mixes each convolution channel. Specifically, the features pass through two convolution layers. The first layer has a $3 \times 3$ kernel and expands the channels. Then, GELU~\cite{GELU} is applied for non-linear mapping. Finally, a convolution layer with a $1 \times 1$ kernel restores the channels to their original state. In our method, the features pass through Layer Normalization~\cite{LN}, LMLT, and another Layer Normalization before being input to CCM~\cite{SAFMN}. Detailed structure can be seen in Figure~\ref{fig:ccm}.

\section{Comparisons on LMLT with Other Methods}
\label{App:Base_large}

\begin{table}[t]
	\caption{Comparisons with existing methods. Best and second-best performance are in \textcolor{red}{red} and \textcolor{blue}{blue}, and third-best is \underline{underlined}. Unreported results are left blank.}
	\label{tab:tiny_small}
	\renewcommand\arraystretch{1.1}
	\begin{center}
		\resizebox{\textwidth}{!}{
			\begin{tabular}{| c | c | c | c | c | c | c | c | c | c |}
				\hline
				Scale & Method & \#Params & \#FLOPs & \#Acts & Set5 & Set14 & B100 & Urban100 & Manga109 \\
				\hline
				\multirow{11}*{$\times 2$} 
                & CARN-M~\cite{CARN}         &412K     &91G &655M  &37.53/0.9583  &33.26/0.9141  &31.92/0.8960  &31.23/0.9193  & - \\
				~ & CARN~\cite{CARN}           &1,592K   &223G &\underline{522M}  &37.76/0.9590  &33.52/0.9166  &32.09/0.8978  &31.92/0.9256  &- \\
				~ & EDSR-baseline~\cite{Flickr2K-EDSR}  &1,370K   &316G  &563M &37.99/0.9604  &33.57/0.9175  &32.16/0.8994  &31.98/0.9272  &38.54/0.9769 \\
				~ & PAN~\cite{PAN}       &\underline{261K}    &\underline{71G}  &677M &\underline{38.00}/0.9605  &33.59/\underline{0.9181}  &\underline{32.18}/0.8997  &32.01/0.9273  &38.70/0.9773 \\
				~ & LAPAR-A~\cite{LAPAR}       &548K     &171G &656M &\textcolor{blue}{38.01}/0.9605  &33.62/\textcolor{blue}{0.9183}  &\textcolor{blue}{32.19}/\underline{0.8999}  &32.10/\underline{0.9283}  &38.67/0.9772 \\
				~ & ECBSR-M16C64~\cite{ECBSR}  &596K     &137G &\textcolor{red}{252M} &37.90/\textcolor{red}{0.9615}  &33.34/0.9178  &32.10/\textcolor{red}{0.9018}  &31.71/0.9250  &- \\
				~ & SMSR~\cite{SMSR}           &985K     &132G &- &\underline{38.00}/0.9601  &\textcolor{blue}{33.64}/0.9179  &32.17/0.8990  &\textcolor{blue}{32.19}/\textcolor{blue}{0.9284}  &38.76/0.9771 \\
				~ & ShuffleMixer~\cite{ShuffleMixer}     &394K     &91G  &832M &\textcolor{blue}{38.01}/\underline{0.9606}  &\underline{33.63}/0.9180 &32.17/0.8995  &31.89/0.9257  &\underline{38.83}/\underline{0.9774} \\
                ~ & SAMFN~\cite{SAFMN} & \textcolor{red}{228K} & \textcolor{red}{52G}  &\textcolor{blue}{299M} & \underline{38.00}/0.9605 & 33.54/0.9177 &  32.16/0.8995 &  31.84/0.9256 &  38.71/0.9771 \\
                ~ & \textbf{LMLT-Tiny(Ours)} & \textcolor{blue}{239K} & \textcolor{blue}{59G} &603M & \textcolor{blue}{38.01}/\underline{0.9606} & 33.59/\textcolor{blue}{0.9183} & \textcolor{blue}{32.19}/\underline{0.8999} & 32.04/0.9273 & \textcolor{blue}{38.90}/\textcolor{blue}{0.9775} \\
                ~ & \textbf{LMLT-Small(Ours)} & 357K & 88G &898M & \textcolor{red}{38.05}/\textcolor{blue}{0.9608} & \textcolor{red}{33.65}/\textcolor{red}{0.9187} & \textcolor{red}{32.24}/\textcolor{blue}{0.9006} & \textcolor{red}{32.31}/\textcolor{red}{0.9298} & \textcolor{red}{39.10}/\textcolor{red}{0.9780} \\
				\hline
				\multirow{10}*{$\times 3$} 
                & CARN-M~\cite{CARN}        &415K      &46G  &327M &33.99/0.9236  &30.08/0.8367  &28.91/0.8000  &27.55/0.8385  & - \\
				~ & CARN~\cite{CARN}          &1,592K    &119G &\underline{268M} &34.29/0.9255  &30.29/0.8407  &29.06/0.8034  &28.06/0.8493  & - \\
				~ & EDSR-baseline~\cite{Flickr2K-EDSR} &1,555K    &160G  &285M &\underline{34.37}/0.9270  &30.28/0.8417  &29.09/0.8052  &28.15/0.8527  &33.45/0.9439 \\
                ~ & PAN~\cite{PAN}       &\underline{261K}    &\underline{39G} &340M &\textcolor{blue}{34.40}/\underline{0.9271}  & 30.36/0.8423  &29.11/0.8050  & 28.11/0.8511  &33.61/\textcolor{blue}{0.9448} \\
				~ & LAPAR-A~\cite{LAPAR}      &594K  &114G &505M &34.36/0.9267  &\underline{30.34}/0.8421  &\underline{29.11}/\underline{0.8054}  &\underline{28.15}/\underline{0.8523}  &33.51/0.9441 \\
				~ & SMSR~\cite{SMSR}          &993K      &68G  &- &\textcolor{blue}{34.40}/0.9270  &30.33/0.8412  &29.10/0.8050  &\textcolor{blue}{28.25}/\textcolor{blue}{0.8536}  &33.68/\underline{0.9445} \\
				~ & ShuffleMixer~\cite{ShuffleMixer}    &415K &43G  &404M   &\textcolor{blue}{34.40}/\textcolor{blue}{0.9272}  &\textcolor{blue}{30.37}/\underline{0.8423}  &\textcolor{blue}{29.12}/\underline{0.8051}  &28.08/0.8498  &\underline{33.69}/\textcolor{blue}{0.9448} \\
                ~ & SAFMN~\cite{SAFMN} &\textcolor{red}{233K} &\textcolor{red}{23G} &\textcolor{red}{134M} &34.34/0.9267 &30.33/0.8418 &29.08/0.8048 &27.95/0.8474 &33.52/0.9437 \\
                ~ & \textbf{LMLT-Tiny(Ours)} & \textcolor{blue}{244K} &\textcolor{blue}{28G} &\textcolor{blue}{283M}& 34.36/\underline{0.9271} & \textcolor{blue}{30.37}/\textcolor{blue}{0.8427} & \textcolor{blue}{29.12}/\textcolor{blue}{0.8057} & 28.10/0.8503 & \textcolor{blue}{33.72}/\textcolor{blue}{0.9448} \\
                ~ & \textbf{LMLT-Small(Ours)} & 361K & 41G &421M & \textcolor{red}{34.50}/\textcolor{red}{0.9280} & \textcolor{red}{30.47}/\textcolor{red}{0.8446} & \textcolor{red}{29.16}/\textcolor{red}{0.8070} & \textcolor{red}{28.29}/\textcolor{red}{0.8544} & \textcolor{red}{33.99}/\textcolor{red}{0.9464} \\
				\hline
    
				\multirow{11}*{$\times 4$} & CARN-M~\cite{CARN}        &412K     &33G  &227M   &31.92/0.8903  &28.42/0.7762  &27.44/0.7304   &25.62/0.7694  & - \\
				~ & CARN~\cite{CARN}          &1,592K   &91G  &194M   &32.13/0.8937  &28.60/0.7806  &\underline{27.58}/0.7349   &26.07/0.7837  & - \\
				~ & EDSR-baseline~\cite{Flickr2K-EDSR} &1,518K   &114G  &202M  &32.09/0.8938  &28.58/0.7813  &27.57/0.7357   &26.04/0.7849  &30.35/0.9067 \\
				~ & PAN~\cite{PAN}          &\underline{272K}    &\underline{28G}  &238M   &32.13/\underline{0.8948}  &28.61/0.7822  &27.59/0.7363   &26.11/0.7854  &30.51/\textcolor{blue}{0.9095} \\ 
				~ & LAPAR-A~\cite{LAPAR}      &659K     &94G  &452M   &32.15/0.8944  &28.61/0.7818  &\textcolor{blue}{27.61}/0.7366   &\textcolor{blue}{26.14}/\textcolor{blue}{0.7871}  &30.42/0.9074 \\
				~ & ECBSR-M16C64~\cite{ECBSR} &603K     &35G  &\textcolor{red}{64M}   &31.92/0.8946  &28.34/0.7817  &27.48/\textcolor{red}{0.7393}   &25.81/0.7773  &- \\
				~ & SMSR~\cite{SMSR}          &1006K    &42G  &-   &32.12/0.8932  &28.55/0.7808  &27.55/0.7351   &\underline{26.11}/\underline{0.7868}  &30.54/0.9085 \\
				~ & ShuffleMixer~\cite{ShuffleMixer}     &411K      &28G  &269M   &\textcolor{blue}{32.21}/\textcolor{blue}{0.8953}  &\textcolor{blue}{28.66}/\textcolor{blue}{0.7827}  &\textcolor{blue}{27.61}/0.7366   &26.08/0.7835  &\textcolor{blue}{30.65}/\underline{0.9093} \\
                ~& SAFMN~\cite{SAFMN} & \textcolor{red}{240K} & \textcolor{red}{14G} &\textcolor{blue}{77M} & 32.18/0.8948 &  28.60/0.7813 & 27.58/0.7359 & 25.97/0.7809 & 30.43/0.9063 \\
                ~ & \textbf{LMLT-Tiny(Ours)} & \textcolor{blue}{251K} & \textcolor{blue}{15G} &\underline{152M} &  \underline{32.19}/0.8947 &  \underline{28.64}/\underline{0.7823} & \underline{27.60}/\underline{0.7369} & 26.08/0.7838 & \underline{30.60}/0.9083 \\
                ~ & \textbf{LMLT-Small(Ours)} & 368K & 23G &227M & \textcolor{red}{32.31}/\textcolor{red}{0.8968} & \textcolor{red}{28.74}/\textcolor{red}{0.7846} & \textcolor{red}{27.66}/\textcolor{blue}{0.7387} & \textcolor{red}{26.26}/\textcolor{red}{0.7894} & \textcolor{red}{30.87}/\textcolor{red}{0.9117} \\
		
				\hline
		\end{tabular}}
	\end{center}
\end{table}

\textbf{Image Reconstruction comparisons}
\label{App:reconstruction}
Here, We first compare the LMLT-Tiny and LMLT-Small with CARN-m, CARN~\cite{CARN}, EDSR-baseline~\cite{Flickr2K-EDSR}, PAN~\cite{PAN}, LAPAR-A~\cite{LAPAR}, ECBSR-M16C64~\cite{ECBSR}, SMSR~\cite{SMSR}, Shuffle-Mixer~\cite{ShuffleMixer}, and SAFMN~\cite{SAFMN}. Table~\ref{tab:tiny_small} shows that our LMLT significantly reduces number of parameters and computation overheads while achieving substantial performance gains on various datasets. LMLT-Small performs well on most datasets, and the LMLT-Tiny also performs second and third best on the BSD100~\cite{BSD100} and Manga109~\cite{Manga109} datasets, except for the Manga109 $\times4$ SSIM~\cite{SSIM}. In particular, the number of parameters and FLOPs are the second smallest after SAFMN~\cite{SAFMN}.  

\textbf{Memory and Running time Comparisons}
\label{App:memtime}
\begin{table}[t]
	\caption{The memory consumption and inference times are reported. All experiments were conducted on a single RTX 3090 GPU.}
	\label{tab:memtime_all}
	\renewcommand\arraystretch{1.1}
	\begin{center}
		\resizebox{\textwidth}{!}{
			\begin{tabular}{| c | c | c | c | c | c | c | c | c |}
				\hline
				Scale & Method & \#GPU Mem [M] & \#Avg Time [ms] & Set5 & Set14 & B100 & Urban100 & Manga109 \\
				\hline
    
				\multirow{14}*{$\times 2$} 
                & CARN-M~\cite{CARN}    & 2707.82 & 67.56 &37.53/0.9583  &33.26/0.9141  &31.92/0.8960  &31.23/0.9193  & - \\
				~ & CARN~\cite{CARN}     &2716.80 & 73.55  &37.76/0.9590  &33.52/0.9166  &32.09/0.8978  &31.92/0.9256  &- \\
                ~ & EDSR-baseline~\cite{Flickr2K-EDSR} &577.61 &43.58 &37.99/0.9604  &33.57/0.9175  &32.16/0.8994  &31.98/0.9272  &38.54/0.9769 \\
                ~ &LAPAR-A~\cite{LAPAR}       & 1812.60 & 43.50 &38.01/0.9605  &33.62/0.9183  &32.190.8999 &32.10/0.9283  &38.67/0.9772 \\
                ~ &SAFMN~\cite{SAFMN}       & 259.56 & 33.61 & 38.00/0.9605 & 33.54/0.9177 &  32.16/0.8995 &  31.84/0.9256 &  38.71/0.9771 \\ 
                ~ &\textbf{LMLT-Tiny(Ours)}    & \textbf{324.01} & \textbf{57.37} & 38.01/0.9606 & 33.59/0.9183 & 32.19/0.8999 & 32.04/0.9273 & 38.90/0.9775 \\ 
                ~ &\textbf{LMLT-Small(Ours)}       & \textbf{324.5} & \textbf{84.22} & 38.05/0.9608 & 33.65/0.9187 & 32.24/0.9006 & 32.31/0.9298 & 39.10/0.9780 \\  
                \cline{2-9}

                & IMDN~\cite{IMDN}        & 795.96 & 31.87 & 38.00/0.9605 & 33.63/0.9177 & 32.19/0.8996 & 32.17/0.9283 & 38.88/0.9774 \\
				~ & HNCT~\cite{hnct}     &1200.55 & 351.49 & 38.08/0.9608  & 33.65/0.9182  & 32.22/0.9001  & 32.22/0.9294  & 38.87/0.9774 \\
                ~ & NGswin~\cite{ngswin} &1440.40 &375.19 &38.05/0.9610 & 33.79/0.9199 & 32.27/0.9008 & 32.53/0.9324 & 38.97/0.9777 \\
                ~ &\textbf{LMLT-Base(Ours)}       &  \textbf{567.75} &  \textbf{81.64} & 38.10/0.9610 & 33.76/0.9201 & 32.28/0.9012 & 32.52/0.9316 & 39.24/0.9783 \\ 
                \cline{2-9} 

                ~ & SwinIR-light~\cite{SwinIR}  &1278.64 & 944.11 &38.14/0.9611&33.86/0.9206&32.31/0.9012&32.76/0.9340&39.12/0.9783 \\
                ~ &SRformer-Light~\cite{srformer}  & 1176.15 & 1006.48 & 38.23/0.9613&33.94/0.9209&32.36/0.9019&32.91/0.9353&39.28/0.9785 \\ 
                ~ & \textbf{LMLT-Large(Ours)} &  \textbf{717.31} &  \textbf{123.07} & 38.18/0.9612 & 33.96/0.9212 & 32.33/0.9017 & 32.75/0.9336 & 39.41/0.9786 \\
    		\hline    
                \hline
                
    		  \multirow{14}*{$\times 3$} 
                & CARN-M~\cite{CARN}    & 1213.10 & 37.56 &33.99/0.9236  &30.08/0.8367  &28.91/0.8000  &27.55/0.8385  & - \\
				~ & CARN~\cite{CARN}     &1222.08 & 41.08  &34.29/0.9255  &30.29/0.8407  &29.06/0.8034  &28.06/0.8493  & - \\
                ~ & EDSR-baseline~\cite{Flickr2K-EDSR} &541.61 &26.14 &34.37/0.9270  &30.28/0.8417  &29.09/0.8052  &28.15/0.8527  &33.45/0.9439 \\
                ~ &LAPAR-A~\cite{LAPAR}       & 1813.84 & 35.95 &34.36/0.9267 & 30.34/0.8421  &29.11/0.8054  &28.15/0.8523  &33.51/0.9441 \\
                ~ &SAFMN~\cite{SAFMN}       & 114.70 & 17.38 &34.34/0.9267 &30.33/0.8418 &29.08/0.8048 &27.95/0.8474 &33.52/0.9437 \\
                ~ &\textbf{LMLT-Tiny(Ours)}    & \textbf{151.96} & \textbf{31.06} & 34.36/0.9271 & 30.37/0.8427 & 29.12/0.8057 & 28.10/0.8503 & 33.72/0.9448 \\
                ~ &\textbf{LMLT-Small(Ours)}       & \textbf{152.5} & \textbf{44.22} & 34.50/0.9280 & 30.47/0.8446 & 29.16/0.8070 & 28.29/0.8544 & 33.99/0.9464 \\ 
                \cline{2-9}

                & IMDN~\cite{IMDN}        & 364.68 & 14.01 & 34.36/0.9270 & 30.32/0.8417 & 29.09/0.8046 & 28.17/0.8519 & 33.61/0.9445 \\
				~ & HNCT~\cite{hnct}     &545.64 & 117.20 & 34.47/0.9275  & 30.44/0.8439  & 29.15/0.8067  & 28.28/0.8557  & 33.81/0.9459 \\
                ~ & NGswin~\cite{ngswin} &696.97 &168.49 &34.52/0.9282& 30.53/0.8456 & 29.19/ 0.8078 &28.52/0.8603 & 33.89/0.9470 \\
                ~ &\textbf{LMLT-Base(Ours)}       & \textbf{266.31} & \textbf{41.43} & 34.58/0.9285 & 30.53/0.8458 & 29.21/0.8084 & 28.48/0.8581 & 34.18/0.9477 \\ 
                \cline{2-9}

                ~ & SwinIR-light~\cite{SwinIR}  &587.63 & 287.96 &34.62/0.9289&30.54/0.8463&29.20/0.8082&28.66/0.8624&33.98/0.9478 \\
                ~ &SRformer-Light~\cite{srformer}  & 529.28 & 312.37 & 34.67/0.9296&30.57/0.8469&29.26/0.8099&28.81/0.8655&34.19/0.9489 \\
                ~ & \textbf{LMLT-Large(Ours)} & \textbf{338.36} & \textbf{58.68} & 34.64/0.9293 & 30.60/0.8471 & 29.26/0.8097 & 28.72/0.8626 & 34.43/0.9491 \\
    		\hline    
                \hline
    
				\multirow{14}*{$\times 4$}
                & CARN-M~\cite{CARN}    & 680.84 & 21.39  &31.92/0.8903  &28.42/0.7762  &27.44/0.7304   &25.62/0.7694  & - \\
				~ & CARN~\cite{CARN}     &689.83 & 20.50  &32.13/0.8937  &28.60/0.7806  &27.58/0.7349   &26.07/0.7837  & - \\
                ~ & EDSR-baseline~\cite{Flickr2K-EDSR} &492.39 &19.86  &32.09/0.8938  &28.58/0.7813  &27.57/0.7357   &26.04/0.7849  &30.35/0.9067 \\
                ~ &LAPAR-A~\cite{LAPAR}       & 1811.47 & 32.24  &32.15/0.8944  &28.61/0.7818  &27.61/0.7366   &26.14/0.7871  &30.42/0.9074 \\
                ~ &SAFMN~\cite{SAFMN}       & 65.26 & 11.28 & 32.18/0.8948 &  28.60/0.7813 & 27.58/0.7359 & 25.97/0.7809 & 30.43/0.9063 \\
                ~ &\textbf{LMLT-Tiny(Ours)}    & \textbf{81.44} & \textbf{23.54} & 32.19/0.8947 &  28.64/0.7823 & 27.60/0.7369 & 26.08/0.7838 & 30.60/0.9083 \\
                ~ &\textbf{LMLT-Small(Ours)}       & \textbf{81.92} & \textbf{31.01} & 32.31/0.8968 & 28.74/0.7846 & 27.66/0.7387 & 26.26/0.7894 & 30.87/0.9117 \\ 
                \cline{2-9} 

                & IMDN~\cite{IMDN}        & 203.02 & 9.71 & 32.21/0.8948 & 28.58/0.7811 & 27.56/0.7353 & 26.04/0.7838 & 30.45/0.9075 \\
				~ & HNCT~\cite{hnct}     &312.72 & 69.61 & 32.31/0.8957  & 28.71/0.7834  & 27.63/0.7381  &26.20/0.7896  & 30.70/0.9112 \\
                ~ & NGswin~\cite{ngswin} &372.94 &118.13 & 32.33/0.8963 & 28.78/0.7859 & 27.66/0.7396& 26.45/0.7963 &  30.80/0.9128 \\
                ~ &\textbf{LMLT-Base(Ours)}       & \textbf{144.00} & \textbf{26.15} & 32.38/0.8971 & 28.79/0.7859 & 27.70/0.7403 & 26.44/0.7947 & 31.09/0.9139 \\ 
                \cline{2-9}

                ~ & SwinIR-light~\cite{SwinIR}  &342.46 & 176.76  &32.44/0.8976&28.77/0.7858&27.69/0.7406&26.47/0.7980&30.92/0.9151 \\
                ~ &SRformer-Light~\cite{srformer}  & 320.95 & 180.42 & 32.51/0.8988&28.82/0.7872&27.73/0.7422&26.67/0.8032&31.17/0.9165\\
                ~ & \textbf{LMLT-Large(Ours)} & \textbf{185.68} & \textbf{34.07} & 32.48/0.8987 & 28.87/0.7879 & 27.75/0.7421 & 26.63/0.8001 & 31.32/0.9163 \\
    		\hline    

		\end{tabular}}
	\end{center}
\end{table}
In this paragraph, we present the memory usage and average inference time of our proposed LMLT compared to other super-resolution methods. Similar to the experimental setup in Table~\ref{tab:memtime}, \#GPU Mem represents the maximum memory usage during inference, measured using PyTorch's \texttt{torch.cuda.max\_memory\_allocated()}. \#AVG Time indicates the average time taken to upscale a total of 50 random images by $\times2$, $\times3$, and $\times4$ scales. The experiments were conducted three times, and the average inference time is reported. Each random image has sizes of 640$\times$360 for $\times2$ scale, 427$\times$240 for $\times3$ scale, and 320$\times$180 for $\times4$ scale.

As shown in Table~\ref{tab:memtime_all}, our proposed LMLT-Tiny uses less memory at all scales compared to all models except SAFMN~\cite{SAFMN}. Although LMLT-Small requires more inference time compared to other models, its GPU usage is almost similar to LMLT-Tiny, and its performance is significantly superior as demonstrated in Table~\ref{tab:tiny_small}.

\textbf{Qualitative Comparisons}
\label{App:qualitative}
\begin{figure*}[!t]\footnotesize
 \begin{center}
  \begin{tabular}{ccccccc}
    \multicolumn{3}{c}{\multirow{5}*[53pt]{
        \includegraphics[width=0.3\linewidth, height=0.33\linewidth]{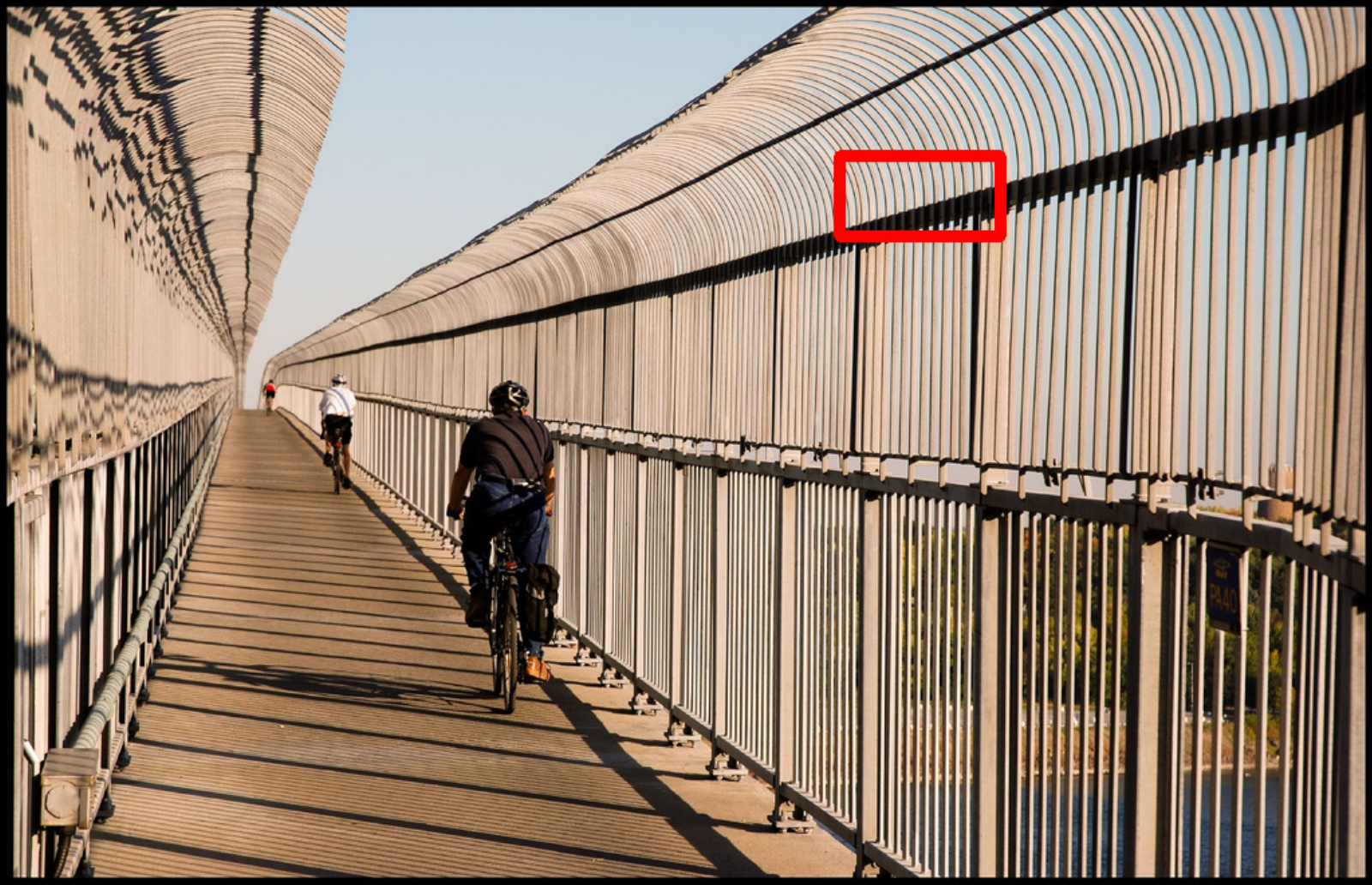}}} &\hspace{-3.5mm}
      \includegraphics[width=0.15\linewidth, height = 0.15\linewidth]{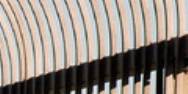} &\hspace{-2mm}
      \includegraphics[width=0.15\linewidth, height = 0.15\linewidth]{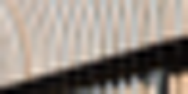} &\hspace{-2mm}
      \includegraphics[width=0.15\linewidth, height = 0.15\linewidth]{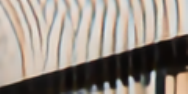} &\hspace{-2mm}
      \includegraphics[width=0.15\linewidth, height = 0.15\linewidth]{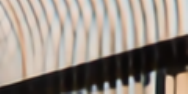} \\
      \multicolumn{3}{c}{~}  &\hspace{-3.5mm}(a) GT  &\hspace{-3.5mm}(b) Bicubic &\hspace{-3.5mm}(c) CARN &\hspace{-3.5mm}(d) EDSR \\

      \multicolumn{3}{c}{~} & \hspace{-3.5mm}
      \includegraphics[width=0.15\linewidth, height = 0.15\linewidth]{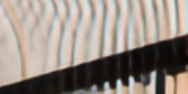} & \hspace{-2mm}
      \includegraphics[width=0.15\linewidth, height = 0.15\linewidth]{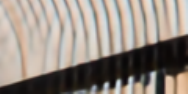} & \hspace{-2mm}
      \includegraphics[width=0.15\linewidth, height = 0.15\linewidth]{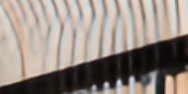} & \hspace{-2mm}
      \includegraphics[width=0.15\linewidth, height = 0.15\linewidth]{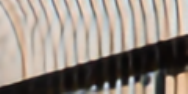} \\
      \multicolumn{3}{c}  {\hspace{-3.5mm} img024($\times4$) from Urban100 }  &\hspace{-3.5mm}(e) PAN &\hspace{-3.5mm}(f) ShuffleMixer  &\hspace{-3.5mm}(g) SAFMN  &\hspace{-3.5mm}(h) LMLT-Tiny \\
      \\

      \multicolumn{3}{c}{\multirow{5}*[53pt]{
        \includegraphics[width=0.3\linewidth, height=0.33\linewidth]{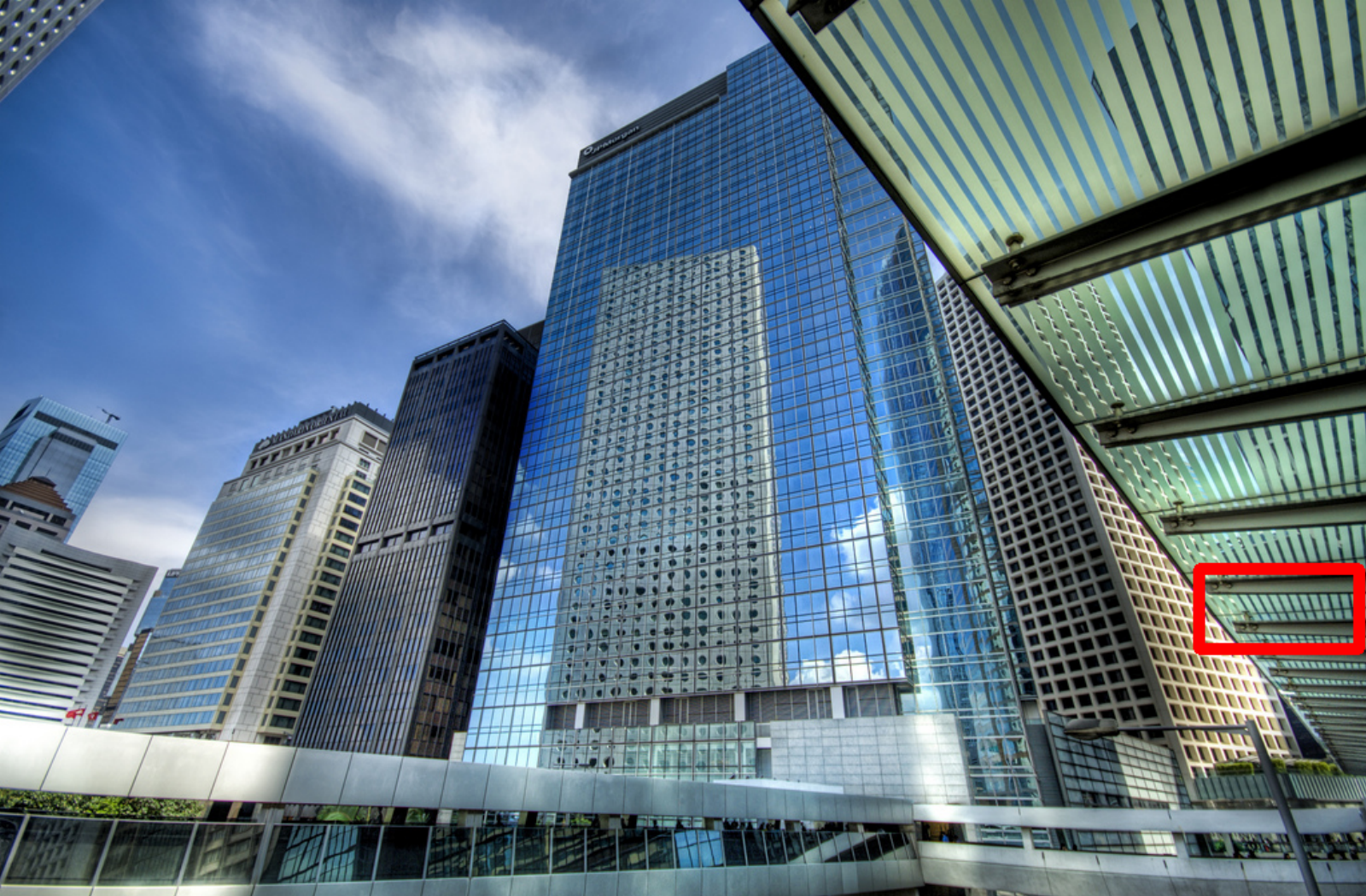}}} &\hspace{-3.5mm}
      \includegraphics[width=0.15\linewidth, height = 0.15\linewidth]{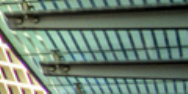} &\hspace{-2mm}
      \includegraphics[width=0.15\linewidth, height = 0.15\linewidth]{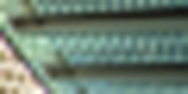} &\hspace{-2mm}
      \includegraphics[width=0.15\linewidth, height = 0.15\linewidth]{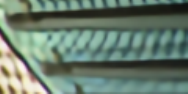} &\hspace{-2mm}
      \includegraphics[width=0.15\linewidth, height = 0.15\linewidth]{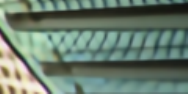} \\
      \multicolumn{3}{c}{~} &\hspace{-3.5mm}(a) GT  &\hspace{-3.5mm}(b) Bicubic &\hspace{-3.5mm}(c) CARN &\hspace{-3.5mm}(d) EDSR~ \\

      \multicolumn{3}{c}{~} & \hspace{-3.5mm}
      \includegraphics[width=0.15\linewidth, height = 0.15\linewidth]{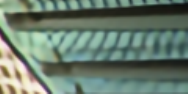} & \hspace{-2mm}
      \includegraphics[width=0.15\linewidth, height = 0.15\linewidth]{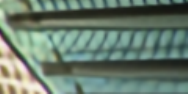} & \hspace{-2mm}
      \includegraphics[width=0.15\linewidth, height = 0.15\linewidth]{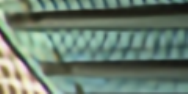} & \hspace{-2mm}
      \includegraphics[width=0.15\linewidth, height = 0.15\linewidth]{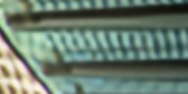} \\
      \multicolumn{3}{c}{\hspace{-3.5mm} img061($\times4$) from Urban100 } &\hspace{-3.5mm}(e) PAN &\hspace{-3.5mm}(f) ShuffleMixer  &\hspace{-3.5mm}(g) SAFMN  &\hspace{-3.5mm}(h) LMLT-Tiny \\
      \\

      \multicolumn{3}{c}{\multirow{5}*[53pt]{
        \includegraphics[width=0.3\linewidth, height=0.33\linewidth]{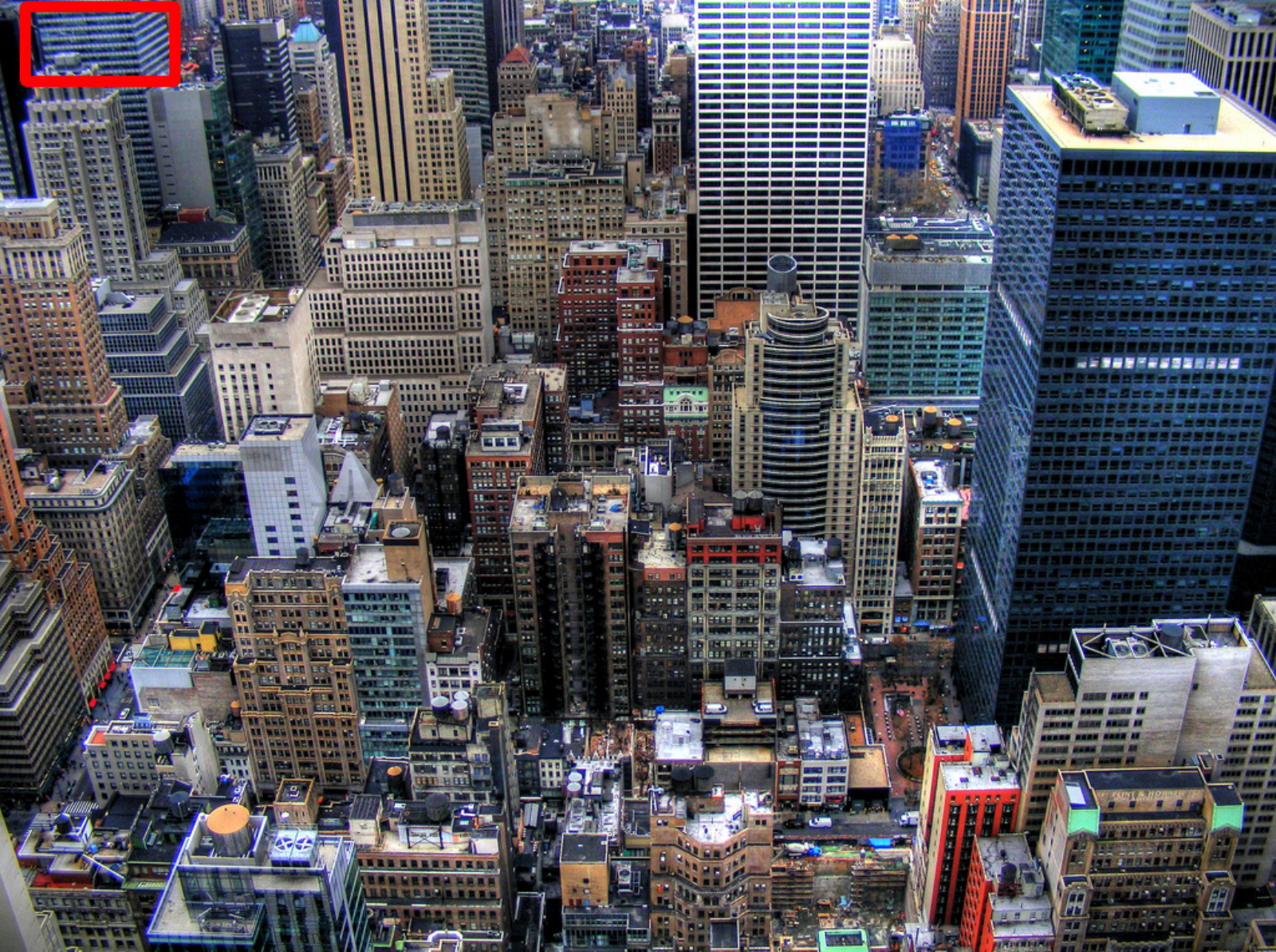}}} &\hspace{-3.5mm}
      \includegraphics[width=0.15\linewidth, height = 0.15\linewidth]{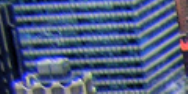} &\hspace{-2mm}
      \includegraphics[width=0.15\linewidth, height = 0.15\linewidth]{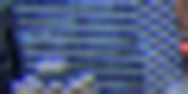} &\hspace{-2mm}
      \includegraphics[width=0.15\linewidth, height = 0.15\linewidth]{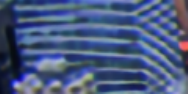} &\hspace{-2mm}
      \includegraphics[width=0.15\linewidth, height = 0.15\linewidth]{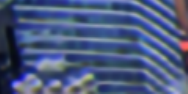} \\
      \multicolumn{3}{c}{~} &\hspace{-3.5mm}(a) GT  &\hspace{-3.5mm}(b) Bicubic &\hspace{-3.5mm}(c) CARN &\hspace{-3.5mm}(d) EDSR \\

      \multicolumn{3}{c}{~} & \hspace{-3.5mm}
      \includegraphics[width=0.15\linewidth, height = 0.15\linewidth]{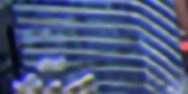} & \hspace{-2mm}
      \includegraphics[width=0.15\linewidth, height = 0.15\linewidth]{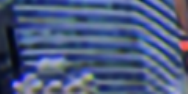} & \hspace{-2mm}
      \includegraphics[width=0.15\linewidth, height = 0.15\linewidth]{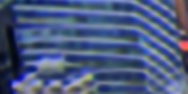} & \hspace{-2mm}
      \includegraphics[width=0.15\linewidth, height = 0.15\linewidth]{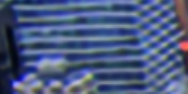} \\
      \multicolumn{3}{c}{\hspace{-3.5mm} img073($\times4$) from Urban100 } &\hspace{-3.5mm}(e) PAN &\hspace{-3.5mm}(f) ShuffleMixer &\hspace{-3.5mm}(g) SAFMN  &\hspace{-3.5mm}(h) LMLT-Tiny \\
      
\\

 \end{tabular}
 \end{center}
 \caption{Visual comparisons for $\times 4$ SR on Urban100 dataset. Compared with the results in (c) to (g), the Ours(LMLT-Tiny, (h)) restore much more accurate and clear images.}
 \label{fig:tiny_compare_appen}
\end{figure*}
\begin{figure*}[!t]\footnotesize
 \begin{center}
  \begin{tabular}{ccccccc}
      \multicolumn{3}{c}{\multirow{5}*[53pt]{
        \includegraphics[width=0.3\linewidth, height=0.33\linewidth]{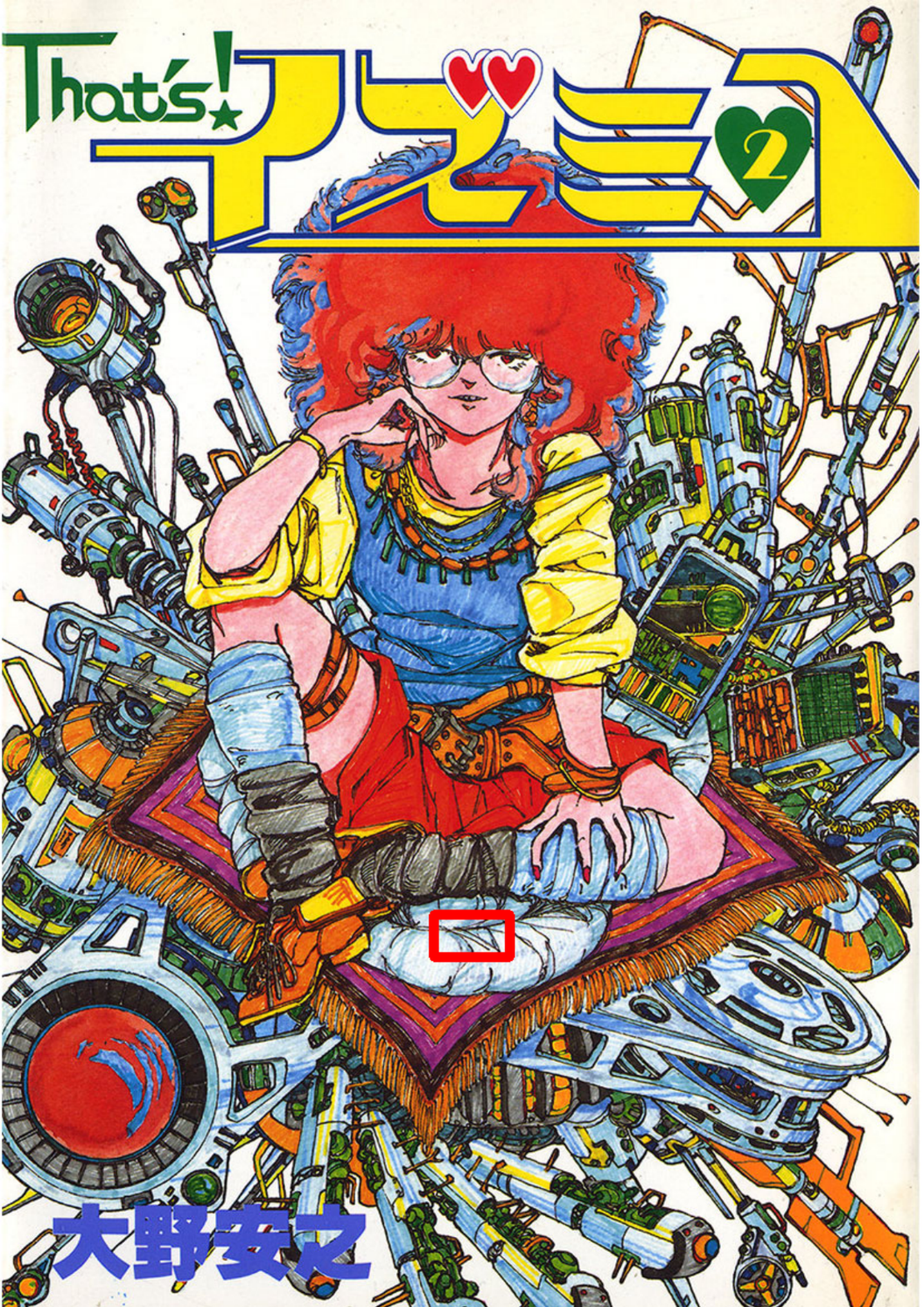}}} &\hspace{-3.5mm}
      \includegraphics[width=0.15\linewidth, height = 0.15\linewidth]{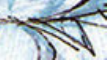} &\hspace{-2mm}
      \includegraphics[width=0.15\linewidth, height = 0.15\linewidth]{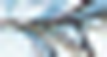} &\hspace{-2mm}
      \includegraphics[width=0.15\linewidth, height = 0.15\linewidth]{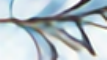} &\hspace{-2mm}
      \includegraphics[width=0.15\linewidth, height = 0.15\linewidth]{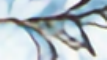} \\
      \multicolumn{3}{c}{~} &\hspace{-3.5mm}(a) GT  &\hspace{-3.5mm}(b) Bicubic &\hspace{-3.5mm}(c) IMDN &\hspace{-3.5mm}(d) NGswin \\

      \multicolumn{3}{c}{~} & \hspace{-3.5mm}
      \includegraphics[width=0.15\linewidth, height = 0.15\linewidth]{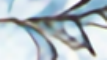} & \hspace{-2mm}
      \includegraphics[width=0.15\linewidth, height = 0.15\linewidth]{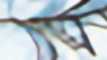} & \hspace{-2mm}
      \includegraphics[width=0.15\linewidth, height = 0.15\linewidth]{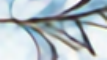} & \hspace{-2mm}
      \includegraphics[width=0.15\linewidth, height = 0.15\linewidth]{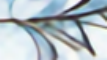} \\
      \multicolumn{3}{c}{\hspace{-3.5mm} Thatsizumiko from Manga109} &\hspace{-3.5mm}(e) SwinIR-light   &\hspace{-3.5mm}(f) SwinIR-NG  &\hspace{-3.5mm}(g) LMLT-Base  &\hspace{-3.5mm}(h) LMLT-Large \\
      \\

      \multicolumn{3}{c}{\multirow{5}*[53pt]{
        \includegraphics[width=0.3\linewidth, height=0.33\linewidth]{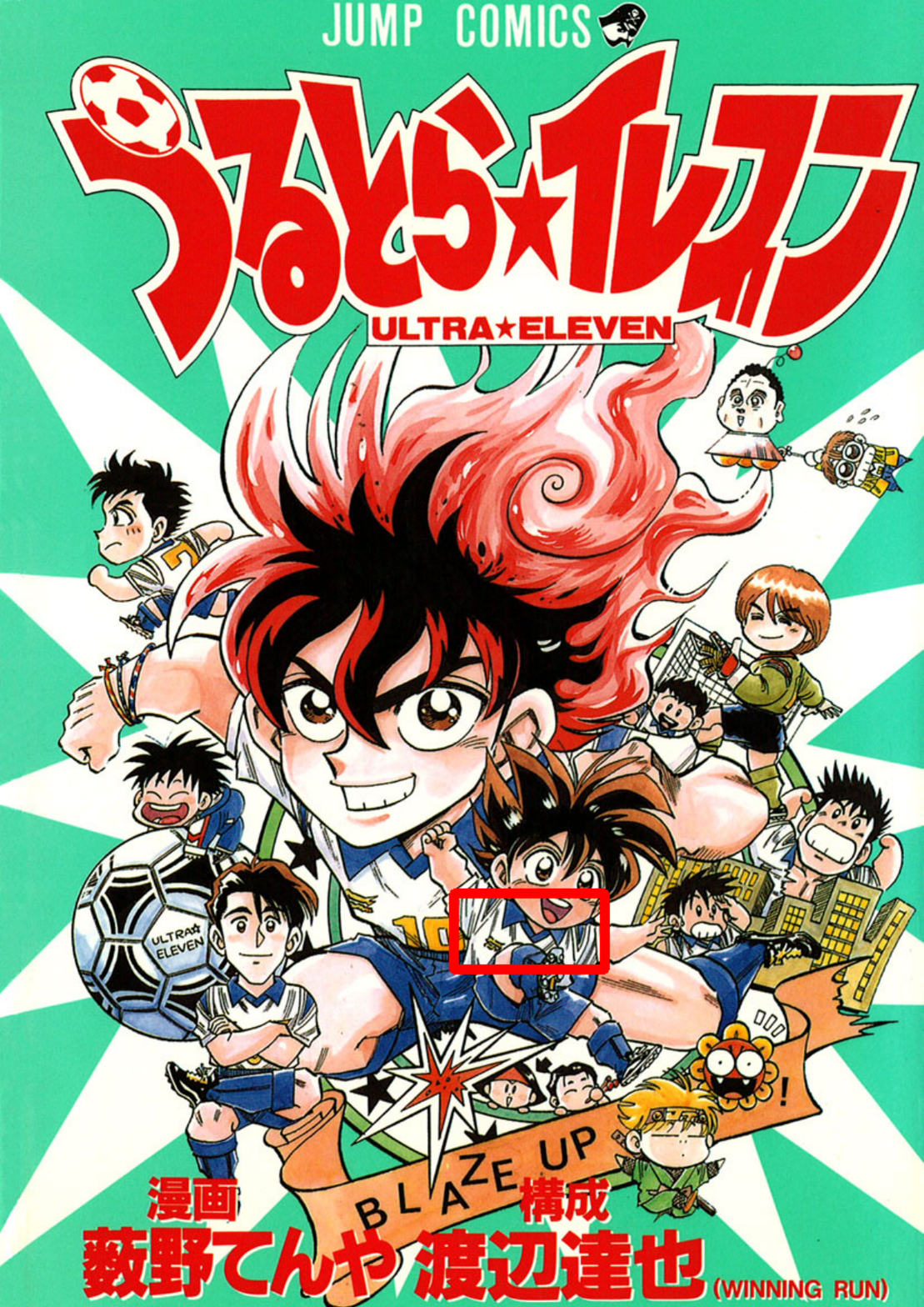}}} &\hspace{-3.5mm}
      \includegraphics[width=0.15\linewidth, height = 0.15\linewidth]{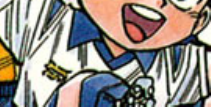} &\hspace{-2mm}
      \includegraphics[width=0.15\linewidth, height = 0.15\linewidth]{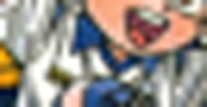} &\hspace{-2mm}
      \includegraphics[width=0.15\linewidth, height = 0.15\linewidth]{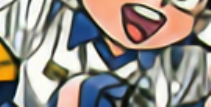} &\hspace{-2mm}
      \includegraphics[width=0.15\linewidth, height = 0.15\linewidth]{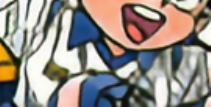} \\
      \multicolumn{3}{c}{~} &\hspace{-3.5mm}(a) GT  &\hspace{-3.5mm}(b) Bicubic &\hspace{-3.5mm}(c) IMDN &\hspace{-3.5mm}(d) NGswin \\

      \multicolumn{3}{c}{~} & \hspace{-3.5mm}
      \includegraphics[width=0.15\linewidth, height = 0.15\linewidth]{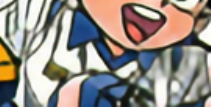} & \hspace{-2mm}
      \includegraphics[width=0.15\linewidth, height = 0.15\linewidth]{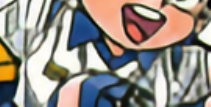} & \hspace{-2mm}
      \includegraphics[width=0.15\linewidth, height = 0.15\linewidth]{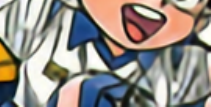} & \hspace{-2mm}
      \includegraphics[width=0.15\linewidth, height = 0.15\linewidth]{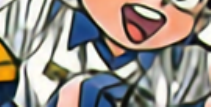} \\
      \multicolumn{3}{c}{\hspace{-3.5mm} UltraEleven from Manga109} &\hspace{-3.5mm}(e) SwinIR-light   &\hspace{-3.5mm}(f) SwinIR-NG  &\hspace{-3.5mm}(g) LMLT-Base  &\hspace{-3.5mm}(h) LMLT-Large \\
      \\

      \multicolumn{3}{c}{\multirow{5}*[53pt]{
        \includegraphics[width=0.3\linewidth, height=0.33\linewidth]{picture/mangax4/YumeiroCookingx4_1_high_GT.pdf}}} &\hspace{-3.5mm}
      \includegraphics[width=0.15\linewidth, height = 0.15\linewidth]{picture/mangax4/YumeiroCookingx4_1_GT.pdf} &\hspace{-2mm}
      \includegraphics[width=0.15\linewidth, height = 0.15\linewidth]{picture/mangax4/YumeiroCookingx4_0_bcbic.pdf} &\hspace{-2mm}
      \includegraphics[width=0.15\linewidth, height = 0.15\linewidth]{picture/mangax4/YumeiroCookingx4_2_IMDN.pdf} &\hspace{-2mm}
      \includegraphics[width=0.15\linewidth, height = 0.15\linewidth]{picture/mangax4/YumeiroCookingx4_3_NGswin.pdf} \\
      \multicolumn{3}{c}{~} &\hspace{-3.5mm}(a) GT  &\hspace{-3.5mm}(b) Bicubic &\hspace{-3.5mm}(c) IMDN &\hspace{-3.5mm}(d) NGswin \\

      \multicolumn{3}{c}{~} & \hspace{-3.5mm}
      \includegraphics[width=0.15\linewidth, height = 0.15\linewidth]{picture/mangax4/YumeiroCookingx4_4_SwinIR.pdf} & \hspace{-2mm}
      \includegraphics[width=0.15\linewidth, height = 0.15\linewidth]{picture/mangax4/YumeiroCookingx4_5_Swinirng.pdf} & \hspace{-2mm}
      \includegraphics[width=0.15\linewidth, height = 0.15\linewidth]{picture/mangax4/YumeiroCookingx4_6_lmlt_60.pdf} & \hspace{-2mm}
      \includegraphics[width=0.15\linewidth, height = 0.15\linewidth]{picture/mangax4/YumeiroCookingx4_7_lmlt_84.pdf} \\
      \multicolumn{3}{c}{\hspace{-3.5mm} YumeiroCooking from Manga109} &\hspace{-3.5mm}(e) SwinIR-light   &\hspace{-3.5mm}(f) SwinIR-NG  &\hspace{-3.5mm}(g) LMLT-Base  &\hspace{-3.5mm}(h) LMLT-Large
      \\
      
 \end{tabular}
 \end{center}
 \caption{Visual comparisons for $\times 4$ SR on Manga109 dataset. Compared with the results in (c) to (f), the Ours(LMLT-Base and LMLT-Large, (g) to (h)) restore much more accurate and clear images.}
 \label{fig:baselarge_compare}
\end{figure*}
In this paragraph, we examine the qualitative comparisons of the LMLT-Tiny model and other models on the Urban100~\cite{Urban100} $\times$4 scale. The comparison includes CARN~\cite{CARN}, EDSR~\cite{Flickr2K-EDSR}, PAN~\cite{PAN}, ShuffleMixer~\cite{ShuffleMixer}, and SAFMN~\cite{SAFMN}. The results can be seen in Figure~\ref{fig:tiny_compare_appen}. As mentioned in section~\ref{comparisons}, we observe that our model reconstructs images with continuous stripes better than other models.

Additionally, we compare our proposed models LMLT-Base and LMLT-Large with IMDN~\cite{IMDN}, NGswin~\cite{ngswin}, SwinIR-light~\cite{SwinIR}, and SwinIR-NG~\cite{ngswin} on the Manga109~\cite{Manga109} dataset at$\times4$ scale. As explained earlier in section~\ref{comparisons}, our model shows strength in areas with continuous lines compared to other models. Figure~\ref{fig:baselarge_compare} illustrates the differences between our LMLT-Base, LMLT-Large and other state-of-the-arts models.

\end{document}